\pdfoutput=1

\documentclass[11pt]{article}

\usepackage{acl}
\usepackage{acronym}

\usepackage{times}
\usepackage{latexsym}
\usepackage{pifont}
\usepackage{tcolorbox}
\usepackage{xcolor}
\usepackage{amsmath}
\usepackage{amssymb} 
\usepackage{appendix}
\usepackage[normalem]{ulem}
\usepackage{booktabs}  
\usepackage{hyperref}
\usepackage{multirow}

\newtcolorbox{alertbox}{
  colback=red!5!white,
  colframe=red!75!black,
  fonttitle=\bfseries,
  title=Takeaway
}

\definecolor{y1}{HTML}{FFFBF8} 
\definecolor{beige1}{HTML}{EDD5A4} 
\definecolor{beige2}{HTML}{F6E4B3} 
\definecolor{beige3}{HTML}{F1D9A7} 
\definecolor{beige4}{HTML}{FAF0DC} 
\definecolor{beige5}{HTML}{EED39D} 
\definecolor{apricot1}{HTML}{FFF3D6}
\definecolor{custom}{HTML}{FFD57D}
\definecolor{custom2}{HTML}{FDFFEE}

\definecolor{beige3}{HTML}{F1D9A7}

\tcbset{
  colback=y1,
  colframe=apricot1,
  fonttitle=\bfseries
}

\usepackage[T1]{fontenc}

\usepackage[utf8]{inputenc}

\usepackage{microtype}

\usepackage{inconsolata}

\usepackage{graphicx}
\usepackage{blindtext}

\usepackage{subcaption}
\usepackage{soul}
\usepackage{xcolor}
\usepackage{booktabs}
\usepackage{tabularx}
\usepackage{xcolor}
\usepackage{soul}
\usepackage{array}
\usepackage{tcolorbox}

\newcolumntype{L}[1]{>{\raggedright\arraybackslash}p{#1}}

\definecolor{palered}{rgb}{1,0.6,0.6}
\definecolor{palegreen}{rgb}{0.7,1,0.7} 

\title{Judging Quality Across Languages: A Multilingual Approach to Pretraining Data Filtering with Language Models}

\author{
   Mehdi Ali\textsuperscript{1,2}$^\dagger$  
   Manuel Brack\textsuperscript{3,5}$^\dagger$ 
   Max Lübbering\textsuperscript{1,2}$^\dagger$ 
   Elias Wendt\textsuperscript{5}$^\dagger$ 
   Abbas Goher Khan\textsuperscript{1}$^\dagger$ \\
   Richard Rutmann\textsuperscript{1,2}
   Alex Jude\textsuperscript{2}
   Maurice Kraus\textsuperscript{5}
   Alexander Arno Weber\textsuperscript{1,2} \\
   David Kaczér\textsuperscript{1}
   Florian Mai\textsuperscript{1}
   Lucie Flek\textsuperscript{1}
   Rafet Sifa\textsuperscript{1,2}
   Nicolas Flores-Herr\textsuperscript{2} \\
   Joachim Köhler\textsuperscript{1,2} 
   Patrick Schramowski\textsuperscript{3,4,5}
   Michael Fromm\textsuperscript{1,2}
   Kristian Kersting\textsuperscript{3,4,5} \\ \\
   \textsuperscript{1}Lamarr Institute,
   \textsuperscript{2}Fraunhofer IAIS,    \textsuperscript{3}DFKI SAINT, \\ 
   \textsuperscript{4}Hessian AI,
   \textsuperscript{5}TU Darmstadt \\
   \texttt{mehdi.ali@iais.fraunhofer.de, brack@cs.tu-darmstadt.de}
\Thanks{\textdagger Equal contribution.}}

\begin{document}

\acrodef{LLM}{Large Language Model}

\maketitle
\begin{abstract}

High-quality multilingual training data is essential for effectively pretraining large language models (LLMs). Yet, the availability of suitable open-source multilingual datasets remains limited. Existing state-of-the-art datasets mostly rely on heuristic filtering methods, restricting both their cross-lingual transferability and scalability. Here, we introduce JQL, a systematic approach that efficiently curates diverse and high-quality multilingual data at scale while significantly reducing computational demands. JQL distills LLMs' annotation capabilities into lightweight annotators 
based on pretrained multilingual embeddings. These models exhibit robust multilingual and cross-lingual performance, even for languages and scripts unseen during training. Evaluated empirically across 35 languages, the resulting annotation pipeline substantially outperforms current heuristic filtering methods like Fineweb2. JQL notably enhances downstream model training quality and increases data retention rates. Our research provides practical insights and valuable resources for multilingual data curation, raising the standards of multilingual dataset development. 

\end{abstract}

\section{Introduction}

The quality of pre-training data remains a crucial factor in LLM performance and represents one of the most effective factors for reducing training costs~\cite{DBLP:conf/nips/PenedoKALMRW024}. 
Even recent improvements in post-training and scaling of inference-time compute heavily depend on the quality of the pre-trained base model~\cite{guo2025deepseek}. 
Consequently, a growing number of research efforts have focused on developing data curation pipelines for large-scale web data.~\cite{DBLP:conf/nips/PenedoKALMRW024,DBLP:conf/nips/LiFSIJGBGKAGXMH24,DBLP:journals/corr/abs-2412-02595}. 

The overall goal of any data filtering set-up is to achieve the largest possible dataset of the highest quality. 
Traditionally, heuristic-based approaches rely on predefined rules to filter the raw training data~\cite{DBLP:conf/lrec/AbadjiSRS22,DBLP:conf/nips/PenedoKALMRW024}.
Recently, however, there has been a shift towards machine learning-based data curation, which tends to outperform complex rule-based systems in producing high-quality pre-training corpora. 
A particularly interesting research avenue is the use of existing LLMs to identify high-quality content.
This ``LLMs as judges to filter datasets'' approach has proven highly effective in selecting high-quality data that leads to more performant models~\cite{DBLP:conf/nips/PenedoKALMRW024,DBLP:journals/corr/abs-2412-02595}.

A significant limitation, however, is that existing research in this area largely focuses on English, making it unclear whether these methods effectively transfer to highly multilingual settings, especially those involving low-resource languages.
Specifically, in contrast to English-centric data curation, multilingual settings raise additional questions on potential gaps between high- and low-resource languages and the cross-lingual performance on unseen languages.
Moreover, much of the research in this field is led by frontier AI labs, which tend to keep state-of-the-art data procurement and curation strategies closed-source, impeding reproducibility and follow-up research.

\begin{figure*}[t]
\centering
\includegraphics[width=\textwidth]{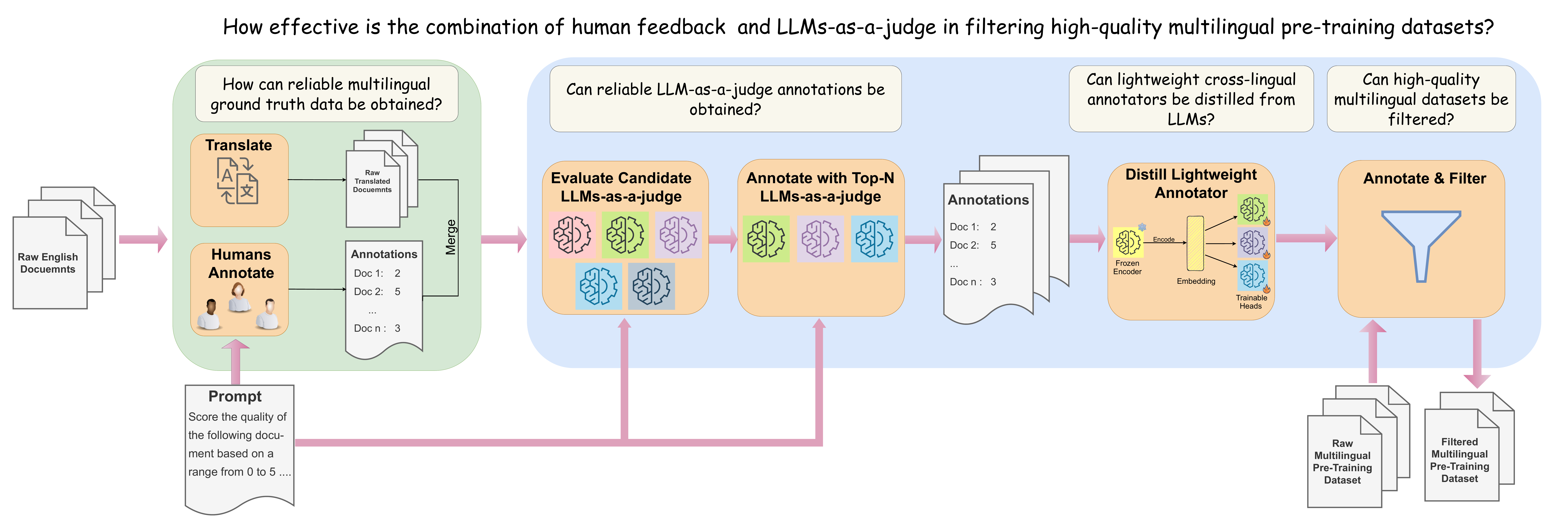}

\caption{The multilingual data filtering approach JQL: In the first stage (Sec.~\ref{sec:ground_truth}), human annotators generate ground truth (GT) annotations on monolingual documents based on an instruction set defined in a prompt. The documents are translated into all target languages to receive a multilingual GT dataset. In the second stage (Sec.~\ref{sec:llms}), based on the GT dataset, we select the top-$n$ performing LLMs-as-a-judge for annotating a multilingual dataset. In the third stage (Sec.~\ref{sec:annotator}), we use the resulting synthetic dataset to train a set of lightweight annotators. This is done at low cost by reusing shared embeddings. 
Using these annotators, we can efficiently annotate pre-training corpora and filter high-quality subsets (Sec.~\ref{sec:downtream_analysis}).}
\label{fig:pipeline}
\end{figure*}





\begin{table}[b!]
\centering
\small
\setlength{\tabcolsep}{1.5pt}
\begin{tabular}{ll}
\toprule
\textbf{Family} & \textbf{Languages} \\
\midrule
\textbf{Slavic} (9) & \textbf{Bulgarian}, Czech, Croatian, Macedonian,\\
            & \textbf{Polish}, Slovak, \\
            & Slovenian, Serbian, \textbf{Ukrainian} \\
\textbf{Germanic} (7) & Danish, \textbf{German}, Icelandic, Dutch, \\
            & \textbf{Norwegian} (\textbf{Bokmål} \& Nynorsk), Swedish \\
\textbf{Romance} (7) & Catalan, \textbf{Spanish}, \textbf{French}, Galician, \textbf{Italian}, \\
            & Portuguese, Romanian \\
\textbf{Uralic} (3) & Estonian, \textbf{Finnish}, \textbf{Hungarian} \\
\textbf{Baltic} (2) & \textbf{Lithuanian}, Latvian \\ \hline
\textbf{Singleton} &  Hellenic (\textbf{Greek}), Celtic (Irish), \\
\textbf{families} & Basque (Basque), West Semitic (Maltese),  \\
& Turkic (\textbf{Turkish}), Albanoid (Albanian), \\
& Armenian (Armenian) \\
\bottomrule
\end{tabular}
\caption{Languages and respective language families considered in this study. The richness of European language families allows for structured research into the influence of inter-language similarities for cross-lingual transfer. For better readability, we report values for languages highlighted in \textbf{bold} in the main body, with remaining values supplied in the Appendix.}
\label{tab:languages}
\vskip -.5em
\end{table}

Addressing these limitations, we propose a multilingual data filtering approach called JQL (\textbf{J}udging \textbf{Q}uality across \textbf{L}anguages)\footnote{pronounced \textit{Jackal}} comprising the four stages outlined in Fig.~\ref{fig:pipeline}. 
With minimal human supervision and small amounts of distilled annotation data, we are able to train lightweight regressors for efficient filtering of multilingual, large-scale data at low computational cost. 
JQL is language agnostic and can be extended to arbitrary filter criteria.

We provide actionable insights and release valuable artifacts from 
each pipeline step\footnote{\url{https://huggingface.co/spaces/JQL-AI/JQL}}. Overall, we  
make the following contributions:
%
%
%
(1) A human-centric approach to creating ground truth by using human annotations to build a reliable dataset for evaluating and guiding pipeline component selection. In this context, we release a novel ground truth dataset comprising 511 manually annotated documents, translated into 35 languages (Sec.~\ref{sec:ground_truth}).
(2) A study investigating LLM capabilities in assessing the quality of multilingual documents (Sec.\ref{sec:llms}). As part of this study, we release annotations from the three best-performing LLMs across 35 languages, covering over 14 million documents.
(3) A study investigating the multi- \& cross-lingual transfer capabilities of lightweight annotator models, evaluating how well judgment abilities generalize to unseen languages (Sec.~\ref{sec:annotator}).
(4) Demonstration that our approach leads to high-quality pre-training datasets that improve the downstream performance of LLMs (Sec.~\ref{sec:downtream_analysis}).

\section{Collecting Human Annotations}\label{sec:ground_truth}

The first step in the JQL pipeline is to collect human ground truth annotations. 
These annotations then serve as the cornerstone of our structured approach for building multilingual data annotators, enabling meaningful cross-validation of all design choices.

\subsection{User Study Design}\label{sec:user_study}

To construct a multilingual ground truth dataset for selecting a large language model (LLM) to serve as a judge in evaluating the educational value of documents, we conducted a human annotation study. 

As a starting point, we leveraged the English LLM-annotated dataset from Fineweb-Edu \cite{DBLP:conf/nips/PenedoKALMRW024}, which contains approximately 450,000 annotations assessing the educational value of documents.
Given the demonstrated effectiveness of their scoring scheme, we adopted the same 6-point scale, ranging from 0 (lowest educational value) to 5 (highest). 
To ensure balanced representation across the scoring spectrum, we sampled 100 documents for each score level. Since only 11 documents were available for score 5, the resulting dataset totals 511 samples. 
These documents form the basis of our human annotation study involving 15 annotators with backgrounds in computer science, English studies, physics and mathematics (details are provided in App~\ref{app:ground_truth}).

To ensure annotation quality and consistency, we employed the educational prompt defined by Fineweb-Edu as annotation guidelines, and conducted a dedicated annotator training session. 
This training proved essential since in a preliminary pilot without training, some annotators partially misunderstood the task despite having access to the written guidelines.
In the main annotation phase, each of the 511 documents received three independent annotations, thus capturing variability in human judgments.
To aggregate the three annotations for each document into a single score, we applied majority voting and averaging  when no clear majority emerged.


\subsection{Multilingual Extension}
For multilingual support, we translated the English ground truth dataset into the 35 European languages outlined in Tab.~\ref{tab:languages}. We decided to focus on these languages, since they offer a good trade-off between linguistic diversity and well-populated language families. 
Nonetheless, we demonstrate in Sec.~\ref{sec:generalization} that our annotation pipeline works equally well on typologically different languages such as Chinese, without requiring any modifications.  
%
We used DeepL for the 22 languages it supports, and GPT-4o for the remaining 13 languages. 
To improve correctness  of the GPT-translated texts, we ran a language classifier over all documents and discarded those not matching the target language.
Additionally, we removed prefatory phrases added by GPT-4o to ensure overall consistency.

\subsection{Assessing Inter-Annotator Agreement}

To verify the consistency of our annotation process, we analyzed the collected labels and annotator consensus. 
We observed a high level of agreement across annotators, as evidenced by a majority agreement for 78.5\% of documents and an overall standard deviation of 0.56.
While the annotation spread was $\leq 2$ for 86\% of the data, a few documents exhibited a spread $>$ 3. 
Upon manual inspection, we found that the educational value of these examples is indeed highly subjective, which resulted in disagreement between annotators.
Overall, our rigorous annotator training and data cleaning procedure have resulted in a reliable ground truth, suitable for robustly evaluating ML-based annotators.

\subsection{Suitable Evaluation Criteria}\label{sec:metric_lm_selection}
Choosing an appropriate evaluation metric is essential for assessing the performance of LLM-based annotators against human-annotated ground truth.

While standard classification metrics like F1 score are appropriate for discrete categories with clear semantic boundaries (e.g., spam vs. non-spam), they are less suitable for ordered categorical labels that span a semantic continuum (e.g., very low, low, medium, high, excellent). 
These metrics are order-invariant, failing to reflect the severity of misclassifications, and are sensitive to scale shifts.
For the task of identifying high-quality documents in a web-scale corpus, the relative ranking of documents is significantly more relevant than adherence to an arbitrary scoring scheme. 

To overcome these limitations, we adopt Spearman correlation as our primary evaluation metric. 
Spearman correlation captures the ordinal structure of the data and is robust to monotonic scale transformations, making it well-suited for assessing models on tasks with ordered semantic categories.

\begin{tcolorbox}[leftrule=1.5mm,top=0.8mm,bottom=0.5mm,]
\textbf{Key Insights:}
\begin{itemize}
    \item Well-trained human annotators can produce consistent, hiqh-quality groundtruth annotations.
    \item Rank-based evaluation metrics are better suited than classification metrics for model selection.
\end{itemize}
\textbf{Released Artifacts:}

17,500 documents in 35 languages with human groundtruth annotations of educational value.\footnote{\tiny \url{https://huggingface.co/datasets/JQL-AI/jql_human_edu_annotations}} 

\end{tcolorbox}

\begin{figure*}[t]
\centering
\includegraphics[width=1\textwidth]{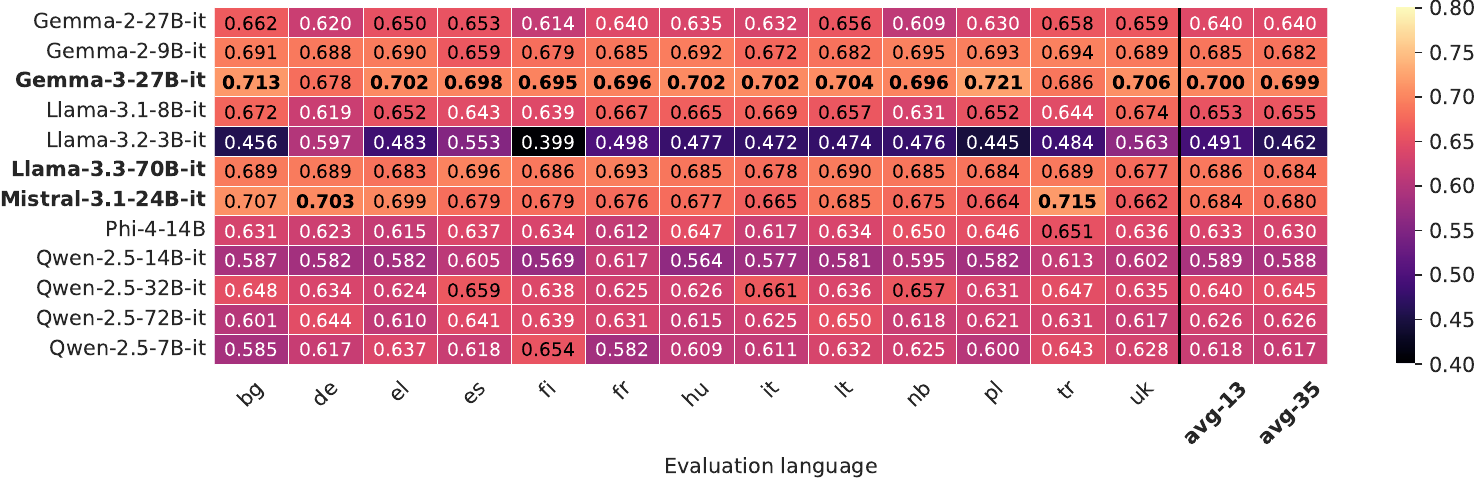}
\caption{
LLMs show varying ranking performance for educational quality. Some models exhibit strong multilingual capabilities. 
We show Spearman Correlation between model predictions and the respective human GT annotations. Scores are displayed for the 13 language subset, their average correlation (avg-13) and the average correlation across all 35 considered languages. The numbers highlighted in bold represent the largest value for each column.
}
\label{figure:correlation_gt_llms}
\end{figure*}

\section{Harnessing LLMs for Multilingual Data Annotation}\label{sec:llms}

%
%

Next, we identify LLMs that are reliable judges of the educational value of documents. 
Subsequently, we can distill these capabilities into more efficient models suitable for data processing at scale. 
We use the ground truth data obtained in the previous JQL step (Section \ref{sec:ground_truth}) to guide model selection.





\subsection{Experimental Setup}\label{sec:llm_inference_set_up}

We selected a diverse set of strong, multilingual LLMs across model sizes and families (Fig.~\ref{figure:correlation_gt_llms}).
To ensure consistency across languages and to leverage the models’ strong English capabilities, we used the original English FineWeb~\cite{DBLP:conf/nips/PenedoKALMRW024} educational prompt for all evaluations. 
We also instructed models to produce English assesments, allowing us to focus on their multilingual natural language understanding  (NLU) rather than their generation capabilities (NLG).
Thus, leveraging the fact that LLMs tend to have good "understanding" in low-resource languages for which they cannot reliably \textit{generate} cohesive outputs \cite{mahfuz-etal-2025-late, luukkonen2024poro, dargis-etal-2024-evaluating}. 
%
Similar to our human annotation setup, we sampled three scores from each model and aggregated them as described in Sec.~\ref{sec:user_study}.

\subsection{Multilingual Evaluation}\label{sec:llm_evaluation}
In ~Fig.~\ref{figure:correlation_gt_llms}, we report the LLMs' capabilities in judging educational content by measuring the correlation with our ground truth annotation. 
We observe substantial differences in performance both across and within model families.
Notably, the smallest model tested, LLaMA-3.2-3B-it, performs significantly worse than all other evaluated models. 
Consequently, effective document quality assessment may require models to exceed a certain parameter threshold, especially if they have not been explicitly trained for such tasks.
With the exception of LLaMA-3.1-8B-it, all models show limited performance variance across languages, supporting our hypothesis that modern LLMs exhibit robust multilingual NLU, even in low-resource settings.
Interestingly, we observed relatively poor classification performance (App.~\ref{app:llm_classification}) for Gemma-3-27B-it despite exhibiting the strongest ranking capabilities. 
Nonetheless, we demonstrate that the model can reliably identify high-quality documents (App.~\ref{app:classification_ablation}), again showcasing the importance of prioritizing ranking metrics and correlation-based evaluation.

Among the evaluated models, Gemma-3-27B-it, Mistral-3.1-24B-it, and LLaMA-3.3-70B-it emerged as the top performing annotators from unique model families. 
We therefore used these models to generate training data for distilling annotation capabilities into lightweight annotators.\footnote{For better readability in the subsequent sections, we refer to Gemma-3-27B-it, Mistral-3.1-24B-it, and LLaMA-3.3-70B-it as Gemma, Mistral, and Llama, respectively.}
Specifically, we randomly sampled up to 500k documents for each of the 35 languages from the unfiltered but de-duplicated Fineweb2\footnote{\tiny\url{https://huggingface.co/datasets/HuggingFaceFW/fineweb-2}} (FW2) dataset, and used each model to generate three predictions per document.\\ 

\begin{tcolorbox}[leftrule=1.5mm,top=0.8mm,bottom=0.5mm,]
\textbf{Key Insights:}
\begin{itemize}
    \item Strong LLMs can reliably assess educational value of web documents.
    \item Using English instructions and responses, LLMs can judge documents in low-resource languages. 
\end{itemize}
\textbf{Artifacts:}
14 Million documents in 35 languages annotated on their educational value by the top-three performing LLMs.\footnote{\tiny \url{https://huggingface.co/datasets/JQL-AI/jql_llms_edu_annotations}}
\end{tcolorbox}


\section{Distilling Lightweight Annotators}\label{sec:annotator}
Next, we distilled lightweight multilingual annotators suitable for curating web-scale data corpora. We use the synthetic labels generated in Sec.~\ref{sec:llms} for training and the human-annotated data obtained in Sec.~\ref{sec:ground_truth} for evaluation.

\subsection{Architecture and Backbone Selection}\label{regressor:architecture_ablation}
We focused on cross-lingual embedding models with long context windows \cite{zhang2024mgte, sturua2024jinaembeddingsv3multilingualembeddingstask, yu2024arcticembed20multilingualretrieval}. These models efficiently process long web documents and produce well-aligned representations that map semantically equivalent texts across languages to similar embeddings. Thus, enabling effective cross-lingual transfer to unseen languages when using these representations as a backbone.

\begin{figure*}
    \centering
    \includegraphics[width=\linewidth]{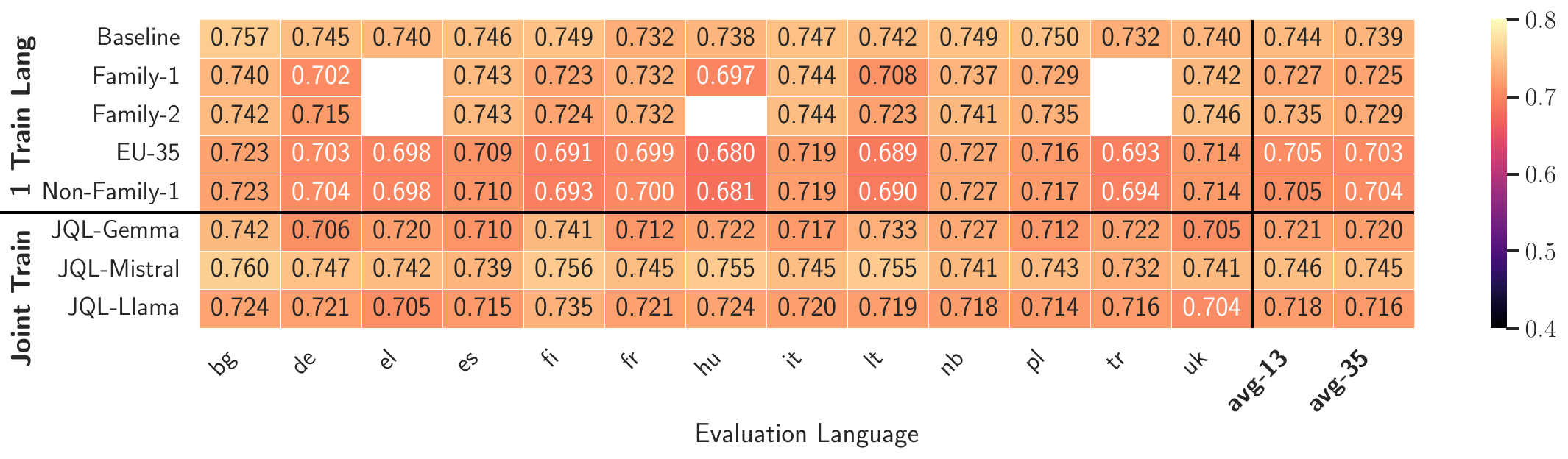}
    \caption{Lightweight JQL annotators show strong multilingual and cross-lingual performance. Training on the same language as the evaluation target serves as a baseline (row 1). We show cross-lingual capabilities by comparing against training on languages within the same language family from Tab.~\ref{tab:languages} (row 2), those within the same, lower-level family (row 3), the full set of the remaining 34 languages (row 4), and those outside the first-order family (row 5). We also show performance for joint training on all languages with the respective LLM data (last 3 rows). Empty cells occur when no related language is present in our dataset. We depict Spearman correlation with ground truth annotation. }
    \label{tab:cross_lingual}
\end{figure*}
%
In our preliminary analysis, Snowflake Arctic Embed v2.0 \cite{yu2024arcticembed20multilingualretrieval} consistently outperformed other candidates (App~\ref{app:annotators_backbone}).
We therefore selected that model as the embedding backbone for our subsequent experiments.
Our results further indicated that keeping the embedding model’s weights frozen while training a lightweight regression head (a simple multilayer perceptron (MLP) with ReLU activation applied to the embeddings) is sufficient to produce high-quality annotations.
We provide detailed results and ablations in App.~\ref{app:annotators}.

This final setup is highly efficient: the lightweight regression head accounts for less than 1\% of total parameters, with embedding computation being the main runtime cost. 
As a result, multiple annotators and tasks, e.g., adult content filtering, mathematical accuracy, or code quality can be supported in parallel by attaching different heads to a shared backbone at minimal additional cost (both training and inference).
Our custom annotation pipeline achieves a throughput of roughly 11,000 annotations per minute on a single A100 with an average of 690 tokens per document.\footnote{Implementation based on Datatrove. Using 6 JQL annotation heads with frozen Snowflake embedding model.}

\subsection{Multilingual Evaluation}\label{regressor:cross_lingual_capabilities}

We present the performance results of the regression-based annotators in Fig.~\ref{tab:cross_lingual}. We observe that baseline performance when training in individual languages remains consistently strong (first row in Fig.~\ref{tab:cross_lingual}), highlighting the robustness of our multilingual architecture. Additionally, we see only slight performance decreases for checkpoints trained on all languages (last 3 rows in Fig.~\ref{tab:cross_lingual}).  
On average, the distilled regression heads even slightly outperform the LLMs from which the training annotations were derived. While part of this improvement is attributable to the shift to continuous labels, the gains also reflect the strength of the pre-trained embedding model.
Only three linguistically isolated languages, Irish, Maltese, and Basque—show notable performance degradation, likely due to their limited representation in the Snowflake training data.

Importantly, these results also support our motivation of strong cross-lingual support through aligned embedding representations. 
We evaluate cross-lingual generalization by considering different typological groups of languages. 
This includes languages within the same language family (Tab.~\ref{tab:languages}; row 2 in Fig.~\ref{tab:cross_lingual}), those within the same family at lower typological level (row 3)\footnote{We consider the following second-order families with more than one representative language: West-, South- \& East-Slavic; North- \& West-Germanic; Italo-Western Romance; and Finnic}, the full set of the remaining 34 languages (row 4) and those outside the first-order family altogether (row 5).
Despite these outliers, cross-lingual performance remains generally robust. Annotators tend to perform slightly worse when evaluated on languages outside their respective first-order families, but models trained on languages from the same family consistently yield stronger results.

We further extend on the cross-lingual capabilities by demonstrating generalization to unseen languages in Sec.~\ref{sec:generalization}. 

\subsection{Building the Final Annotator}
To systematically explore the amount of data required to effectively train our lightweight annotation models, we conducted a controlled experiment involving all 35 languages. 
The performance converged with 500k training samples 
(App~\ref{app:annotators_data}).


Building upon the insights gained, we trained our final lightweight annotator models. We used a frozen Snowflake Arctic Embed v2 backbone, trained on 500,000 documents sampled evenly across all 35 languages.
We trained dedicated annotation heads for each LLM annotator---Gemma, Mistral, and Llama---to facilitate targeted comparisons and flexibility. Furthermore, for each lightweight annotator, we consider two distinct regression heads. The first set of heads is trained on randomly drawn samples representative of the natural distribution of labels. For the second, we strategically selected samples per language to achieve the most uniform possible label distribution, to counteract potential biases towards over-represented labels. In practice, we thus highly over-sampled documents with scores 4 and 5.


\begin{tcolorbox}[leftrule=1.5mm,top=0.8mm,bottom=0.5mm,]
\textbf{Key Insights:}
\begin{itemize}
    \item Well calibrated, multilingual embedding models serve as powerful backbones for data annotation. 
    \item Lightweight regression heads enable efficient annotation and zero-shot cross-lingual transfer.
\end{itemize}
\textbf{Artifacts:}
Three lightweight annotators for educational quality\footnote{\url{https://huggingface.co/JQL-AI/JQL-Edu-Heads}} for use in our custom data-annotation pipeline.\footnote{\url{https://github.com/JQL-AI/JQL-Annotation-Pipeline/}}
\end{tcolorbox}

\section{Assessing Training Data Quality}\label{sec:downtream_analysis}
Next, we assess the effectiveness of the JQL lightweight annotators in identifying high-quality pre-training data.

\subsection{Experimental Setup}

To that end, we conducted extensive ablation studies using the raw, unfiltered FW2 dataset \cite{penedo2024fineweb-2}. 
This dataset originates from Common Crawl WARC files and includes standard preprocessing such as HTML extraction, language identification, and deduplication. 
Using the unfiltered raw data ensures that our comparisons directly reflect differences introduced by our annotator-driven filtering methods, rather than preprocessing variations. 
We benchmark our annotation-based filters against the original heuristic filtering approach used by FW2. 
For these experiments, we selected 13 languages that collectively represent major European language families, ensuring diverse linguistic coverage (see \textbf{bold} languages in Tab.~\ref{tab:languages}).



\begin{figure}[b!]
    \centering
    \includegraphics[width=.95\linewidth]{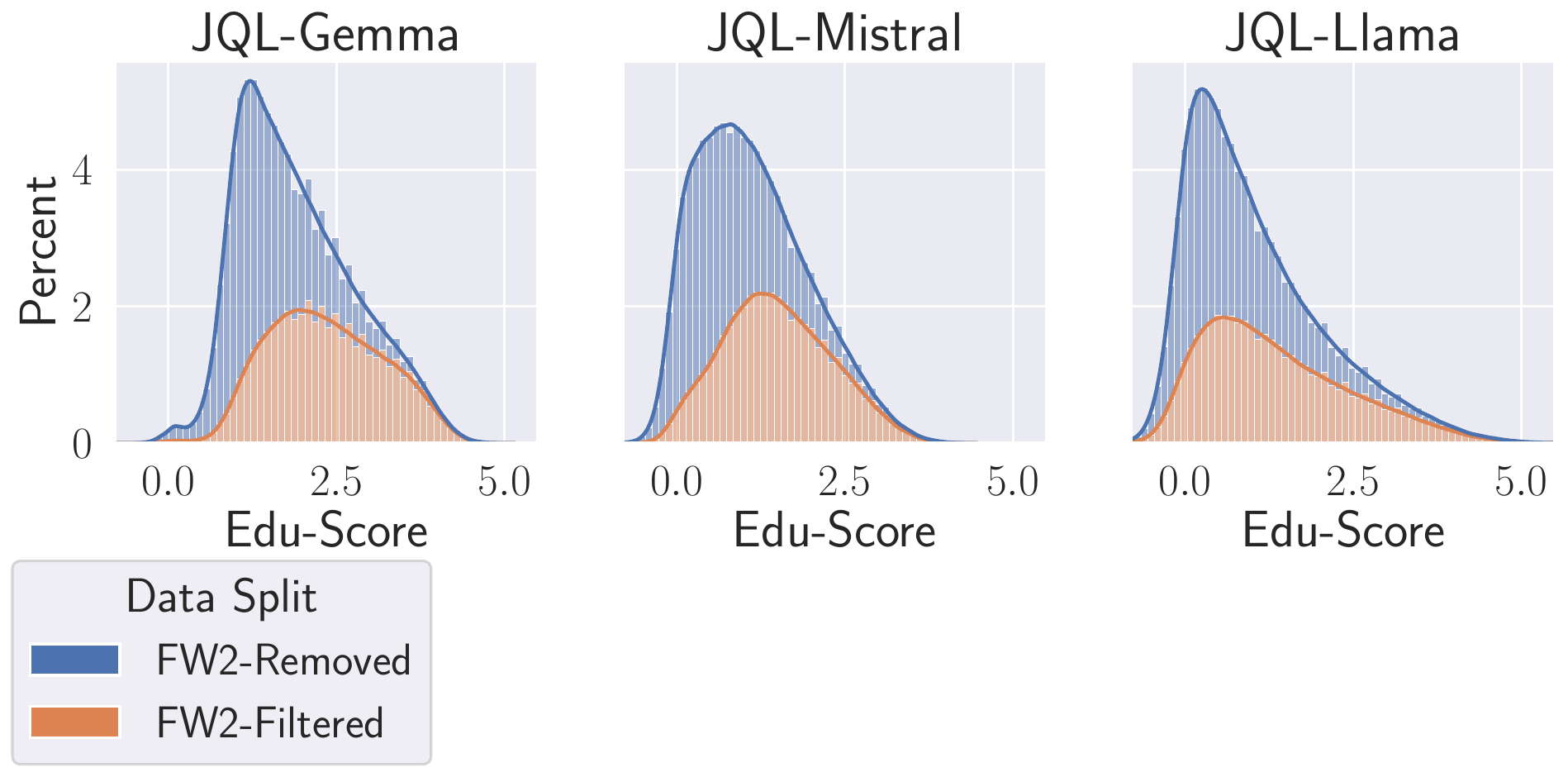}
    \caption{Lightweight annotators trained on different synthetic labels produce different educational score distributions. On average, Gemma assigns higher values than Mistral or Llama. Consequently, thresholding needs to be dynamic and account for the annotators' distribution.  Example plotted for CC release 2024-14 over 13 languages.}
    \label{fig:annotation_distribution}
\end{figure}

For all training ablations, we used dense decoder-only models with 2 billion parameters, following the LLaMA architecture \cite{touvron2023llama2}. 
The training datasets comprised 27 billion and 14 billion monolingual tokens, with 14 billion tokens used for the languages with limited training data.
A detailed description of the training hyper-parameters is provided in App.~\ref{app:downstream_analysis_setup}.


To compare model quality across training runs and respective datasets, we used multilingual versions of MMLU \cite{hendryckstest2021}, HellaSwag \cite{zellers2019hellaswag}, and ARC \cite{allenai:arc}. 
Instead of accuracy, we relied on the token-normalized probability of the correct answer as our main metric, as it yields smoother and more interpretable learning curves. 



Experiments at this parameter and token count reliably predict which datasets perform better when scaling to larger models and more data \cite{magnusson2025datadecidep}. 
However, the absolute benchmark are not indicative of final downstream performance, as our ablation models remain heavily under-trained. 
The relationship between performance at this scale and that of large-scale pre-training is governed by more complex scaling laws.


\begin{figure}
    \centering
    \includegraphics[width=\linewidth]{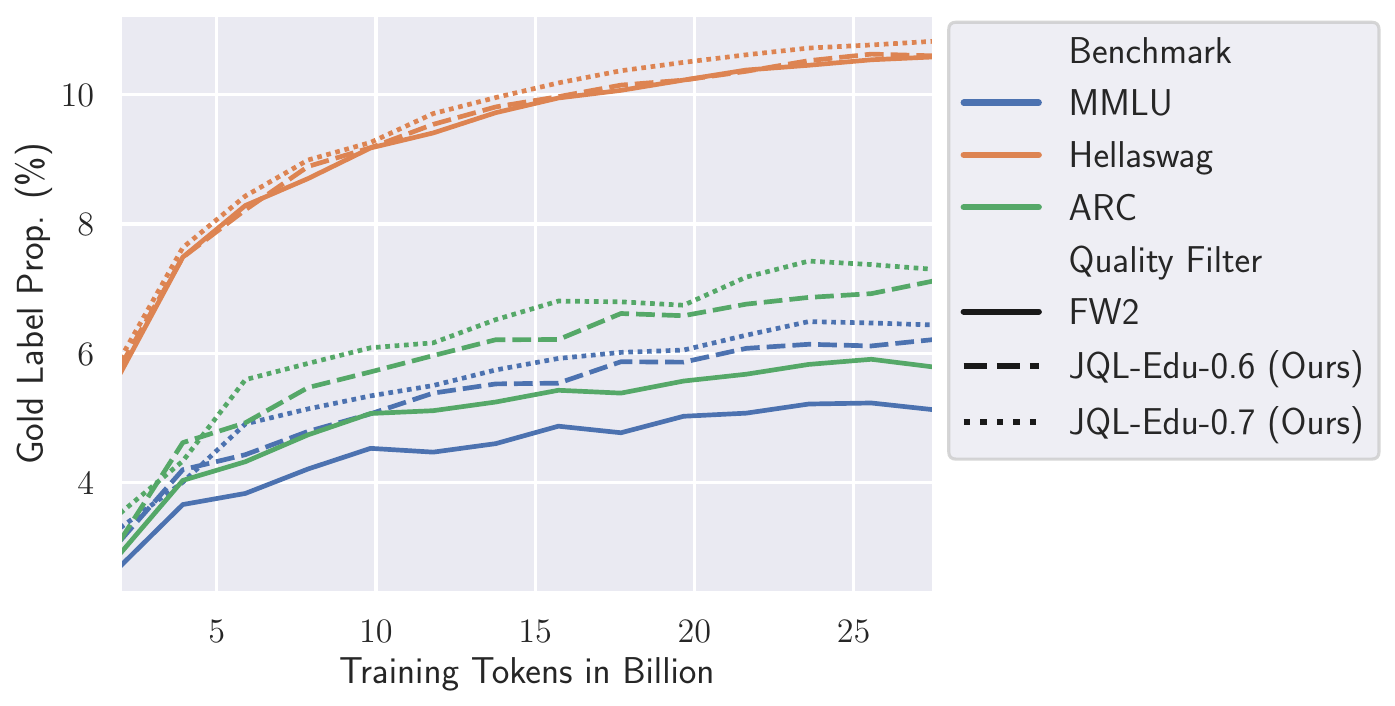}
    \caption{Our JQL annotators improve pre-training data quality over heuristic baselines (FW2). 
    The exemplary plot depicts results for the Spanish dataset.}
    \label{fig:es_progress_example} 
\end{figure}

\subsection{Annotation Analysis}\label{sec:annotation_distribution}
Following the annotation phase, we conducted a detailed statistical analysis of the score distributions produced by different lightweight annotators, as shown in Fig.~\ref{fig:annotation_distribution}.
First, we observe that the heuristically filtered subset of FW2 (orange) exhibits notably higher average educational quality scores compared to the removed data (blue). This serves as a sanity check, indicating that FW2’s heuristic filters capture a meaningful baseline signal.
Additionally, the regression heads trained on synthetic labels generated by different LLMs, i.e., Gemma, Mistral, and Llama, exhibit significantly different score distributions.
In particular, JQL-annotators based on Gemma consistently assign higher educational quality scores than those based on Mistral, which in turn rate samples higher than Llama on average. Notably, this property is inherited from the LLM-based annotators which have different but order-preserving scales of educational content (App.~Fig.\,\ref{fig:llm_annotator_cumulative_score_dist}).
We also found regression heads trained on datasets with more balanced label distributions to produce less skewed annotation outputs, which may facilitate more stable and interpretable threshold selection (App.~\ref{app:downstream_analysis_distribution}).

\begin{table}[]
    \centering
    \small
    \begin{tabular}{l c c c }
    \toprule
    & \multicolumn{3}{c}{\textbf{Change over FW2 baselines (\%)}} \\
    \midrule
    \multirow{ 2}{*}{\textbf{Quantile}} & \multirow{ 2}{*}{Tokens (\%)} & \multicolumn{2}{c}{Benchmark} \\
     &  & Avg. & Final \\
     0.6    & $+$\phantom{0}4.8	& $+$4.27 & $+$4.6\\
     0.7    & $-$15.8	& $+$6.70 & $+$7.2	\\
     
     \bottomrule
    \end{tabular}
    \caption{Percentile-based filtering on JQL annotations provides reliable trade-offs in performance improvements and achieves higher data quality and document retention. Retained tokens and benchmark performance are reported relative to the FW2 baseline and aggregated over 13 languages. Benchmark "Avg." and "Final" depict the relative difference in the mean and final checkpoint performances, respectively (see Fig.~\ref{fig:es_progress_example}). }
    \label{tab:progress_agg}
\end{table}

Despite differences in absolute score distributions, the annotations showed very high correlation (Spearman’s $r$ > 0.87), indicating strong agreement in the relative ranking of document quality across annotators.
This observation aligns with our discussion (Sec.~\ref{sec:llms}) that all models are similarly effective at ranking document quality, even if their classification accuracy  varies.
This finding highlights that absolute thresholds (e.g., scores $\geq3$) lack general validity unless supported by extensive ablation.
We adopt percentile-based (relative) thresholds computed per regression head to address this oversight, enabling more robust comparisons and filtering.
This approach allows to directly control the trade-off between document quality and corpus size.

\subsection{Evaluating Pre-training Data Quality}

We evaluated the impact of JQL on downstream model performance by filtering the pre-training data based on two relative threshold values: the 0.6 and 0.7 percentiles per lightweight annotator head.
To include a document in the final training dataset, we required agreement across an ensemble of three distinct lightweight annotators (Gemma, Mistral, and Llama)\footnote{These heads were trained once on balanced labels and remained fixed throughout.}. Each had to rate the document above its respective percentile threshold.
This ensemble-based filtering approach enhances robustness by reducing the influence of individual annotator biases and minimizing the noise present in single-model annotations. The original FW2 heuristic filtering method serves as our baseline, providing reference points for both the volume of retained tokens and downstream model performance.


Figure \ref{fig:es_progress_example} exemplarily demonstrates the effectiveness of our approach for Spanish, with aggregated cross-lingual results shown in Table \ref{tab:progress_agg}. The results clearly demonstrate that JQL-based filtering consistently outperforms FW2’s heuristic baseline in terms of data quality. We also observe a correlation between threshold strictness and quality gains, with the higher percentile threshold (0.7) consistently yielding better results than 0.6. Overall, JQL offers a scalable and reliable signal for data quality, enabling systematic control of the quality–quantity trade-off, which is particularly useful for scenarios like curriculum learning.


Importantly, our annotation-driven filtering achieves higher-quality training outcomes without excessively aggressive data reduction. For example, in the Spanish language case, applying the 0.6 threshold retains over 9\% more tokens than FW2 while still surpassing its quality. This advantageous trend holds consistently across languages, as confirmed by our aggregated results. Thus, demonstrating that our approach effectively improves training performance even when preserving more documents compared to heuristic baselines. Eliminating overly aggressive filtering is especially relevant in multilingual scenarios, where limited data is available for many languages. 

\begin{tcolorbox}[leftrule=1.5mm,top=0.8mm,bottom=0.5mm,]
\textbf{Key Insights:}
\begin{itemize}
    \item JQL outperforms multilingual heuristic filtering.
    \item Percentile-based filtering is better suited than threshold-based filtering
    \item Higher percentile thresholds trade-off better data quality for reduced number of tokens.
\end{itemize}
\end{tcolorbox}

\section{Generalization to Unseen Languages}\label{sec:generalization}

To validate the versatile and robust cross-lingual capabilities of our JQL approach beyond European languages, we conducted additional experiments on three linguistically and typologically distinct languages,  specifically Arabic, Thai, and Mandarin Chinese, which represent language families completely unseen during training. 
We first validated the capabilities of the existing lightweight annotators on those languages. When measuring their correlation on respective translations of the ground truth data, we observed similar performance as for the European languages (App.~\ref{app:generalization_annotator}). Consequently, we can simply use the existing lightweight annotators with no further training required.
We applied the same dynamic percentile-based filtering approach (specifically, the 0.7 quantile threshold) that had previously proven effective across our European language annotations.

\begin{figure}
    \centering
    \small
    \includegraphics[width=0.9\linewidth]{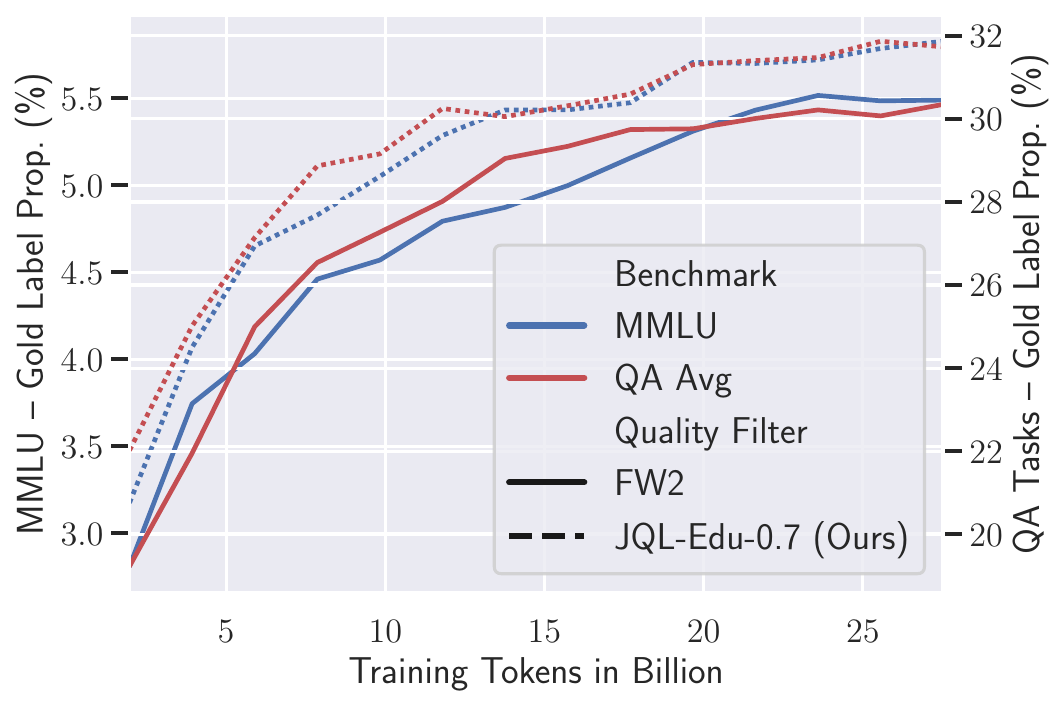}
    \caption{Our JQL  lightweight annotators generalize to unseen, topologically different languages. The figure shows aggregated performance on Arabic, Thai and Chinese. With limited available of standard benchmarks, we relied on language-specific benchmarks selected by Fineweb2 \cite{penedo2024fineweb-2}.}
\label{fig:lingual_generalization}
\end{figure}

The results in Fig.~\ref{fig:lingual_generalization} demonstrate that even for these entirely unseen languages, the JQL pipeline maintains strong zero-shot performance, confirming their capability to effectively generalize across diverse linguistic contexts. These findings highlight the broad applicability and practical scalability of our approach.  Consequently, JQL is suitable for extending robust data curation practices into low-resource and underrepresented languages with minimal additional overhead.

\section{Related work}
\textbf{Heuristic Based Data Curation Pipelines.} The vast majority of training data for large language models is sourced from the web, with Common Crawl (CC) being the most important corpus. 
Traditionally, many works have relied heavily, and in some cases exclusively, on heuristic-based filtering methods to clean and select web data~\cite{raffel2020exploring, gao2020pile, weber2024maurice, penedo2023refined}.
These heuristics typically focus on document-level syntax, such as removing ill-formed or overly short texts, as well as filtering out documents containing blocklisted keywords.
Web-based corpora are often further enriched with high-quality sources such as code, academic literature, or Wikipedia articles~\cite{gao2020pile}.

\textbf{Neural Data Curation Pipelines.}
A major drawback of heuristic filters is their inability to assess the \textit{semantic} quality of documents. Consequently, more recent dataset curation incorporates neural networks into the process \cite{DBLP:conf/icml/WettigGM024,DBLP:journals/corr/abs-2412-02595,DBLP:conf/nips/PenedoKALMRW024,DBLP:conf/emnlp/ZhaoTZHZBC0024,DBLP:conf/nips/LiFSIJGBGKAGXMH24,DBLP:conf/emnlp/ZhaoTZHZBC0024,sachdeva2024train,DBLP:conf/icml/KorbakSCBBPBP23}. To scale these approaches to billions of documents, small and task-specific FastText classifiers \cite{joulin2016bag} are the most common choice. 

These quality annotators are increasingly trained on synthetic labels derived from strong, general-purpose LLMs. Specifically, annotations and filters judging the \textit{educational quality} of a document have produced hiqh-quality datasets \cite{DBLP:journals/corr/abs-2412-02595, DBLP:conf/nips/PenedoKALMRW024, DBLP:conf/icml/WettigGM024}. 

\textbf{Multilingual Data Curation Pipelines.}
Despite these advances in dataset curation, they remain largely English-centric (with a growing body of research dedicated to Chinese). While large multilingual datasets exist, the respective filtering pipelines and dataset sizes are not on par with the high-quality ones for English data \cite{kudugunta2023Madlad, nguyen2024culturax,brack2024communityoscar,xue2021mc4,burchell2025expanded}

The best-performing large-scale multilingual dataset is FineWeb2 \cite{penedo2024fineweb-2}, which solely relies on heuristic filtering. 
In this paper, we developed a data curation pipeline that provides advanced quality filtering in the multilingual setting and seamlessly transfers to unseen languages. 

\section{Conclusion \& Future Directions}

In this work, we proposed JQL, a multilingual pre-training data filtering approach that requires minimal human supervision and leverages language models as judges.
We systematically evaluate JQL across 35 languages for filtering educationally valuable content. Our experiments provide extensive evidence that JQL effectively selects high-quality multilingual pre-training data, significantly outperforming heuristic-based filtering methods.
Further, our approach is scalable to large datasets, generalizes to unseen languages, and is easily extendable.


JQL opens several promising avenues for future research.
First, it is readily applicable to arbitrarily filtering criteria, including code quality, mathematical correctness, and adult content moderation.
Second, it can be used not only for curating pre-training datasets but also for selecting relevant data in various post-training stages, such as instruction tuning and alignment.
Ultimately, our contributions lay a rigorous foundation for improved multilingual data curation and set a new standard for leveraging language and embedding models effectively in multilingual contexts.

\section{Limitations}

Despite the breadth and generalizability of our work, we acknowledge the following limitations.

First, due to the infeasibility of manual annotation at scale, we machine-translated our human-annotated English ground truth dataset into the 35 target languages rather than manually annotating ground truth data in each language.

Second, while we demonstrated the effectiveness of JQL in filtering high-quality multilingual documents solely based on their educational value, our approach is not limited to this specific criterion. JQL is designed to support arbitrary filtering objectives. We chose educational value as our primary focus because it has been shown to be a strong indicator for identifying high-quality multilingual pre-training data~\cite{DBLP:conf/icml/WettigGM024}.

Finally, due to the high computational cost, we conducted our ablation studies at a single model scale (2 billion parameters). 
Despite this limitation, we observed consistent improvements in downstream performance, indicating the effectiveness of JQL-filtered datasets.
Given the evidence of current scaling laws \cite{magnusson2025datadecidep}, we expect equally strong performance gains at larger model scales, which we leave for future work.

\section{Acknowledgment}

This work was funded by the Federal Ministry of Research, Technology \& Space Germany (BMFTR) and the state of North Rhine-Westphalia as part of the Lamarr Institute for Machine Learning and Artificial Intelligence (LAMARR22B), as well as by the European Union’s Horizon 2020 research and innovation program under grant agreement No. 101135671 (TrustLLM).

The authors gratefully acknowledge EuroHPC (\url{https://eurohpc-ju.europa.eu/index_en}) and the Barcelona Supercomputing Center (\url{https://www.bsc.es/}) for providing computational resources on MareNostrum 5. 
Furthermore, we thank hessian.AI for providing easy access to their 42 supercomputers, and acknowledge the support of the hessian.AI Innovation Lab (funded by the Hessian Ministry for Digital Strategy and Innovation), the hessian.AISC Service Center (funded by the BMFTR, grant No 01IS22091), and the Center for European Research in Trusted AI (CERTAIN). Further, this work benefited from the National High Performance Computing Center for Computational Engineering Science (NHR4CES) and project “XEI” (FKZ 01IS24079B) funded by the BMFTR.
Finally, we thank Felix Friedrich and Pedro Ortiz Suarez for their feedback.  

\bibliography{references}

\begin{thebibliography}{38}
\providecommand{\natexlab}[1]{#1}

\bibitem[{Abadji et~al.(2022)Abadji, Su{\'{a}}rez, Romary, and
  Sagot}]{DBLP:conf/lrec/AbadjiSRS22}
Julien Abadji, Pedro Javier~Ortiz Su{\'{a}}rez, Laurent Romary, and
  Beno{\^{\i}}t Sagot. 2022.
\newblock \href {https://aclanthology.org/2022.lrec-1.463} {Towards a cleaner
  document-oriented multilingual crawled corpus}.
\newblock In \emph{Proceedings of the Thirteenth Language Resources and
  Evaluation Conference, {LREC}}. European Language Resources Association.

\bibitem[{Allal et~al.(2025)Allal, Lozhkov, Bakouch, Blázquez, Penedo,
  Tunstall, Marafioti, Kydlíček, Lajarín, Srivastav, Lochner, Fahlgren,
  Nguyen, Fourrier, Burtenshaw, Larcher, Zhao, Zakka, Morlon, Raffel, von
  Werra, and Wolf}]{allal2025smollm2smolgoesbig}
Loubna~Ben Allal, Anton Lozhkov, Elie Bakouch, Gabriel~Martín Blázquez,
  Guilherme Penedo, Lewis Tunstall, Andrés Marafioti, Hynek Kydlíček,
  Agustín~Piqueres Lajarín, Vaibhav Srivastav, Joshua Lochner, Caleb
  Fahlgren, Xuan-Son Nguyen, Clémentine Fourrier, Ben Burtenshaw, Hugo
  Larcher, Haojun Zhao, Cyril Zakka, Mathieu Morlon, Colin Raffel, Leandro von
  Werra, and Thomas Wolf. 2025.
\newblock \href {https://arxiv.org/abs/2502.02737} {Smollm2: When smol goes big
  -- data-centric training of a small language model}.
\newblock \emph{arXiv preprint arXiv:2502.02737}.

\bibitem[{Artetxe et~al.(2019)Artetxe, Ruder, and Yogatama}]{Artetxe:etal:2019}
Mikel Artetxe, Sebastian Ruder, and Dani Yogatama. 2019.
\newblock \href {https://arxiv.org/abs/1910.11856} {On the cross-lingual
  transferability of monolingual representations}.
\newblock \emph{arXiv preprint arXiv1910.11856:}.

\bibitem[{Brack et~al.(2024)Brack, Ostendorff, Ortiz~Suarez, Saiz, Castilla,
  Palomar-Giner, Shvets, Schramowski, Rehm, Villegas, and
  Kersting}]{brack2024communityoscar}
Manuel Brack, Malte Ostendorff, Pedro Ortiz~Suarez, Jos{\'e}~Javier Saiz,
  I{\~n}aki~Lacunza Castilla, Jorge Palomar-Giner, Alexander Shvets, Patrick
  Schramowski, Georg Rehm, Marta Villegas, and Kristian Kersting. 2024.
\newblock \href {https://aclanthology.org/2024.mrl-1.19/} {Community oscar: A
  community effort for multilingual web data}.
\newblock In \emph{Proceedings of the Fourth Workshop on Multilingual
  Representation Learning (MRL)}.

\bibitem[{Burchell et~al.(2025)Burchell, de~Gibert, Arefyev, Aulamo, Bañón,
  Chen, Fedorova, Guillou, Haddow, Hajič, Helcl, Henriksson, Klimaszewski,
  Komulainen, Kutuzov, Kytöniemi, Laippala, Mæhlum, Malik, Mehryary,
  Mikhailov, Moghe, Myntti, O'Brien, Oepen, Pal, Piha, Pyysalo,
  Ramírez-Sánchez, Samuel, Stepachev, Tiedemann, Variš, Vojtěchová, and
  Zaragoza-Bernabeu}]{burchell2025expanded}
Laurie Burchell, Ona de~Gibert, Nikolay Arefyev, Mikko Aulamo, Marta Bañón,
  Pinzhen Chen, Mariia Fedorova, Liane Guillou, Barry Haddow, Jan Hajič,
  Jindřich Helcl, Erik Henriksson, Mateusz Klimaszewski, Ville Komulainen,
  Andrey Kutuzov, Joona Kytöniemi, Veronika Laippala, Petter Mæhlum,
  Bhavitvya Malik, Farrokh Mehryary, Vladislav Mikhailov, Nikita Moghe, Amanda
  Myntti, Dayyán O'Brien, Stephan Oepen, Proyag Pal, Jousia Piha, Sampo
  Pyysalo, Gema Ramírez-Sánchez, David Samuel, Pavel Stepachev, Jörg
  Tiedemann, Dušan Variš, Tereza Vojtěchová, and Jaume Zaragoza-Bernabeu.
  2025.
\newblock \href {https://arxiv.org/abs/2503.10267} {An expanded massive
  multilingual dataset for high-performance language technologies}.
\newblock \emph{arXiv preprint arXiv:2503.10267}.

\bibitem[{Clark et~al.(2020)Clark, Choi, Collins, Garrette, Kwiatkowski,
  Nikolaev, and Palomaki}]{tydiqa}
Jonathan~H. Clark, Eunsol Choi, Michael Collins, Dan Garrette, Tom Kwiatkowski,
  Vitaly Nikolaev, and Jennimaria Palomaki. 2020.
\newblock \href {https://aclanthology.org/2020.tacl-1.30/} {Tydi qa: A
  benchmark for information-seeking question answering in typologically diverse
  languages}.
\newblock \emph{Transactions of the Association for Computational Linguistics}.

\bibitem[{Clark et~al.(2018)Clark, Cowhey, Etzioni, Khot, Sabharwal, Schoenick,
  and Tafjord}]{allenai:arc}
Peter Clark, Isaac Cowhey, Oren Etzioni, Tushar Khot, Ashish Sabharwal, Carissa
  Schoenick, and Oyvind Tafjord. 2018.
\newblock \href {https://arxiv.org/abs/1803.05457} {Think you have solved
  question answering? try arc, the ai2 reasoning challenge}.
\newblock \emph{arXiv preprint arXiv:1803.05457}.

\bibitem[{Cui et~al.(2019)Cui, Liu, Che, Xiao, Chen, Ma, Wang, and
  Hu}]{cui-etal-2019-span}
Yiming Cui, Ting Liu, Wanxiang Che, Li~Xiao, Zhipeng Chen, Wentao Ma, Shijin
  Wang, and Guoping Hu. 2019.
\newblock \href {https://aclanthology.org/D19-1600/} {A span-extraction dataset
  for {C}hinese machine reading comprehension}.
\newblock In \emph{Proceedings of the Conference on Empirical Methods in
  Natural Language Processing and the International Joint Conference on Natural
  Language Processing ({EMNLP-IJCNLP})}.

\bibitem[{Dar{\c{g}}is et~al.(2024)Dar{\c{g}}is, B{\={a}}rzdi{\c{n}}{\v{s}},
  Skadi{\c{n}}a, Gr{\={u}}z{\={i}}tis, and
  Saul{\={i}}te}]{dargis-etal-2024-evaluating}
Roberts Dar{\c{g}}is, Guntis B{\={a}}rzdi{\c{n}}{\v{s}}, Inguna Skadi{\c{n}}a,
  Normunds Gr{\={u}}z{\={i}}tis, and Baiba Saul{\={i}}te. 2024.
\newblock Evaluating open-source {LLM}s in low-resource languages: Insights
  from {L}atvian high school exams.
\newblock In \emph{Proceedings of the 4th International Conference on Natural
  Language Processing for Digital Humanities}.

\bibitem[{Gao et~al.(2020)Gao, Biderman, Black, Golding, Hoppe, Foster, Phang,
  He, Thite, Nabeshima, Presser, and Leahy}]{gao2020pile}
Leo Gao, Stella Biderman, Sid Black, Laurence Golding, Travis Hoppe, Charles
  Foster, Jason Phang, Horace He, Anish Thite, Noa Nabeshima, Shawn Presser,
  and Connor Leahy. 2020.
\newblock \href {https://arxiv.org/abs/2101.00027} {The pile: An 800gb dataset
  of diverse text for language modeling}.
\newblock \emph{arXiv preprint arXiv:2101.00027}.

\bibitem[{Guo et~al.(2025)Guo, Yang, Zhang, Song, Zhang, Xu, Zhu, Ma, Wang, Bi
  et~al.}]{guo2025deepseek}
Daya Guo, Dejian Yang, Haowei Zhang, Junxiao Song, Ruoyu Zhang, Runxin Xu,
  Qihao Zhu, Shirong Ma, Peiyi Wang, Xiao Bi, et~al. 2025.
\newblock \href {https://arxiv.org/abs/2501.12948} {Deepseek-r1: Incentivizing
  reasoning capability in llms via reinforcement learning}.
\newblock \emph{arXiv preprint arXiv:2501.12948}.

\bibitem[{Hendrycks et~al.(2021)Hendrycks, Burns, Basart, Zou, Mazeika, Song,
  and Steinhardt}]{hendryckstest2021}
Dan Hendrycks, Collin Burns, Steven Basart, Andy Zou, Mantas Mazeika, Dawn
  Song, and Jacob Steinhardt. 2021.
\newblock \href {https://openreview.net/forum?id=d7KBjmI3GmQ} {Measuring
  massive multitask language understanding}.
\newblock In \emph{Proceedings of the International Conference on Learning
  Representations ({ICLR})}.

\bibitem[{Joulin et~al.(2016)Joulin, Grave, Bojanowski, and
  Mikolov}]{joulin2016bag}
Armand Joulin, Edouard Grave, Piotr Bojanowski, and Tomas Mikolov. 2016.
\newblock \href {https://arxiv.org/abs/1607.01759} {Bag of tricks for efficient
  text classification}.
\newblock \emph{arXiv preprint arXiv:1607.01759}.

\bibitem[{Korbak et~al.(2023)Korbak, Shi, Chen, Bhalerao, Buckley, Phang,
  Bowman, and Perez}]{DBLP:conf/icml/KorbakSCBBPBP23}
Tomasz Korbak, Kejian Shi, Angelica Chen, Rasika~Vinayak Bhalerao,
  Christopher~L. Buckley, Jason Phang, Samuel~R. Bowman, and Ethan Perez. 2023.
\newblock \href {https://proceedings.mlr.press/v202/korbak23a/korbak23a.pdf}
  {Pretraining language models with human preferences}.
\newblock In \emph{Proceedings of the International Conference on Machine
  Learning ({ICML})}, Proceedings of Machine Learning Research.

\bibitem[{Kudugunta et~al.(2023)Kudugunta, Caswell, Zhang, Garcia, Xin,
  Kusupati, Stella, Bapna, and Firat}]{kudugunta2023Madlad}
Sneha Kudugunta, Isaac Caswell, Biao Zhang, Xavier Garcia, Derrick Xin, Aditya
  Kusupati, Romi Stella, Ankur Bapna, and Orhan Firat. 2023.
\newblock \href
  {https://papers.nips.cc/paper_files/paper/2023/hash/d49042a5d49818711c401d34172f9900-Abstract-Datasets_and_Benchmarks.html}
  {{MADLAD-400:} {A} multilingual and document-level large audited dataset}.
\newblock In \emph{Proceedings of the Advances in Neural Information Processing
  Systems: Annual Conference on Neural Information Processing Systems
  ({NeurIPS})}.

\bibitem[{Lewis et~al.(2019)Lewis, Oguz, Rinott, Riedel, and
  Schwenk}]{lewis2019mlqa}
Patrick Lewis, Barlas Oguz, Ruty Rinott, Sebastian Riedel, and Holger Schwenk.
  2019.
\newblock \href {https://arxiv.org/abs/1910.07475} {Mlqa: Evaluating
  cross-lingual extractive question answering}.
\newblock \emph{arXiv preprint arXiv:1910.07475}.

\bibitem[{Li et~al.(2024)Li, Fang, Smyrnis, Ivgi, Jordan, Gadre, Bansal, Guha,
  Keh, Arora, Garg, Xin, Muennighoff, Heckel, Mercat, Chen, Gururangan,
  Wortsman, Albalak, Bitton, Nezhurina, Abbas, Hsieh, Ghosh, Gardner, Kilian,
  Zhang, Shao, Pratt, Sanyal, Ilharco, Daras, Marathe, Gokaslan, Zhang, Chandu,
  Nguyen, Vasiljevic, Kakade, Song, Sanghavi, Faghri, Oh, Zettlemoyer, Lo,
  El{-}Nouby, Pouransari, Toshev, Wang, Groeneveld, Soldaini, Koh, Jitsev,
  Kollar, Dimakis, Carmon, Dave, Schmidt, and
  Shankar}]{DBLP:conf/nips/LiFSIJGBGKAGXMH24}
Jeffrey Li, Alex Fang, Georgios Smyrnis, Maor Ivgi, Matt Jordan, Samir~Yitzhak
  Gadre, Hritik Bansal, Etash Guha, Sedrick~Scott Keh, Kushal Arora, Saurabh
  Garg, Rui Xin, Niklas Muennighoff, Reinhard Heckel, Jean Mercat, Mayee~F.
  Chen, Suchin Gururangan, Mitchell Wortsman, Alon Albalak, Yonatan Bitton,
  Marianna Nezhurina, Amro Abbas, Cheng{-}Yu Hsieh, Dhruba Ghosh, Josh Gardner,
  Maciej Kilian, Hanlin Zhang, Rulin Shao, Sarah~M. Pratt, Sunny Sanyal,
  Gabriel Ilharco, Giannis Daras, Kalyani Marathe, Aaron Gokaslan, Jieyu Zhang,
  Khyathi~Raghavi Chandu, Thao Nguyen, Igor Vasiljevic, Sham~M. Kakade, Shuran
  Song, Sujay Sanghavi, Fartash Faghri, Sewoong Oh, Luke Zettlemoyer, Kyle Lo,
  Alaaeldin El{-}Nouby, Hadi Pouransari, Alexander Toshev, Stephanie Wang, Dirk
  Groeneveld, Luca Soldaini, Pang~Wei Koh, Jenia Jitsev, Thomas Kollar, Alex
  Dimakis, Yair Carmon, Achal Dave, Ludwig Schmidt, and Vaishaal Shankar. 2024.
\newblock \href
  {https://proceedings.neurips.cc/paper_files/paper/2024/hash/19e4ea30dded58259665db375885e412-Abstract-Datasets_and_Benchmarks_Track.html}
  {Datacomp-lm: In search of the next generation of training sets for language
  models}.
\newblock In \emph{Proceedings of the Advances in Neural Information Processing
  Systems: Annual Conference on Neural Information Processing Systems
  ({NeurIPS})}.

\bibitem[{Luukkonen et~al.(2024)Luukkonen, Burdge, Zosa, Talman, Komulainen,
  Hatanp{\"a}{\"a}, Sarlin, and Pyysalo}]{luukkonen2024poro}
Risto Luukkonen, Jonathan Burdge, Elaine Zosa, Aarne Talman, Ville Komulainen,
  V{\"a}in{\"o} Hatanp{\"a}{\"a}, Peter Sarlin, and Sampo Pyysalo. 2024.
\newblock \href {https://arxiv.org/abs/2404.01856} {Poro 34b and the blessing
  of multilinguality}.
\newblock \emph{arXiv preprint arXiv:2404.01856}.

\bibitem[{Magnusson et~al.(2025)Magnusson, Tai, Bogin, Heineman, Hwang,
  Soldaini, Bhagia, Liu, Groeneveld, Tafjord, Smith, Koh, and
  Dodge}]{magnusson2025datadecidep}
Ian Magnusson, Nguyen Tai, Ben Bogin, David Heineman, Jena~D. Hwang, Luca
  Soldaini, Akshita Bhagia, Jiacheng Liu, Dirk Groeneveld, Oyvind Tafjord,
  Noah~A. Smith, Pang~Wei Koh, and Jesse Dodge. 2025.
\newblock \href {https://arxiv.org/abs/2504.11393} {Datadecide: How to predict
  best pretraining data with small experiments}.
\newblock \emph{arXiv preprint arXiv:2504.11393}.

\bibitem[{Mahfuz et~al.(2025)Mahfuz, Dey, Naswan, Adil, Sayeed, and
  Shahgir}]{mahfuz-etal-2025-late}
Tamzeed Mahfuz, Satak~Kumar Dey, Ruwad Naswan, Hasnaen Adil, Khondker~Salman
  Sayeed, and Haz~Sameen Shahgir. 2025.
\newblock \href {https://aclanthology.org/2025.coling-main.79.pdf} {Too late to
  train, too early to use? a study on necessity and viability of low-resource
  {B}engali {LLM}s}.
\newblock In \emph{Proceedings of the International Conference on Computational
  Linguistics ({COLING})}.

\bibitem[{Mozannar et~al.(2019)Mozannar, Maamary, El~Hajal, and
  Hajj}]{mozannar-etal-2019-neural}
Hussein Mozannar, Elie Maamary, Karl El~Hajal, and Hazem Hajj. 2019.
\newblock \href {https://www.aclweb.org/anthology/W19-4612} {Neural {A}rabic
  question answering}.
\newblock In \emph{Proceedings of the Fourth Arabic Natural Language Processing
  Workshop}. Association for Computational Linguistics.

\bibitem[{Nguyen et~al.(2024)Nguyen, Nguyen, Lai, Man, Ngo, Dernoncourt, Rossi,
  and Nguyen}]{nguyen2024culturax}
Thuat Nguyen, Chien~Van Nguyen, Viet~Dac Lai, Hieu Man, Nghia~Trung Ngo, Franck
  Dernoncourt, Ryan~A. Rossi, and Thien~Huu Nguyen. 2024.
\newblock \href
  {http://www.lrec-conf.org/proceedings/lrec-coling-2024/pdf/2024.main-1.377.pdf}
  {{C}ultura{X}: A cleaned, enormous, and multilingual dataset for large
  language models in 167 languages}.
\newblock In \emph{Proceedings of the Joint International Conference on
  Computational Linguistics, Language Resources and Evaluation (LREC-COLING)}.

\bibitem[{Penedo et~al.(2024{\natexlab{a}})Penedo, Kydl{\'{\i}}cek, Allal,
  Lozhkov, Mitchell, Raffel, von Werra, and
  Wolf}]{DBLP:conf/nips/PenedoKALMRW024}
Guilherme Penedo, Hynek Kydl{\'{\i}}cek, Loubna~Ben Allal, Anton Lozhkov,
  Margaret Mitchell, Colin~A. Raffel, Leandro von Werra, and Thomas Wolf.
  2024{\natexlab{a}}.
\newblock \href
  {https://proceedings.neurips.cc/paper_files/paper/2024/hash/370df50ccfdf8bde18f8f9c2d9151bda-Abstract-Datasets_and_Benchmarks_Track.html}
  {The fineweb datasets: Decanting the web for the finest text data at scale}.
\newblock In \emph{Proceedings of the Advances in Neural Information Processing
  Systems: Annual Conference on Neural Information Processing Systems
  ({NeurIPS})}.

\bibitem[{Penedo et~al.(2024{\natexlab{b}})Penedo, Kydlíček, Sabolčec,
  Messmer, Foroutan, Jaggi, von Werra, and Wolf}]{penedo2024fineweb-2}
Guilherme Penedo, Hynek Kydlíček, Vinko Sabolčec, Bettina Messmer, Negar
  Foroutan, Martin Jaggi, Leandro von Werra, and Thomas Wolf.
  2024{\natexlab{b}}.
\newblock \href {https://doi.org/10.57967/hf/3744} {Fineweb2: A sparkling
  update with 1000s of languages}.

\bibitem[{Penedo et~al.(2023)Penedo, Malartic, Hesslow, Cojocaru, Alobeidli,
  Cappelli, Pannier, Almazrouei, and Launay}]{penedo2023refined}
Guilherme Penedo, Quentin Malartic, Daniel Hesslow, Ruxandra Cojocaru, Hamza
  Alobeidli, Alessandro Cappelli, Baptiste Pannier, Ebtesam Almazrouei, and
  Julien Launay. 2023.
\newblock \href
  {https://proceedings.neurips.cc/paper_files/paper/2023/hash/fa3ed726cc5073b9c31e3e49a807789c-Abstract-Datasets_and_Benchmarks.html}
  {The refinedweb dataset for falcon {LLM:} outperforming curated corpora with
  web data only}.
\newblock In \emph{Proceedings of the Advances in Neural Information Processing
  Systems: Annual Conference on Neural Information Processing Systems
  ({NeurIPS})}.

\bibitem[{Raffel et~al.(2020)Raffel, Shazeer, Roberts, Lee, Narang, Matena,
  Zhou, Li, and Liu}]{raffel2020exploring}
Colin Raffel, Noam Shazeer, Adam Roberts, Katherine Lee, Sharan Narang, Michael
  Matena, Yanqi Zhou, Wei Li, and Peter~J. Liu. 2020.
\newblock \href {https://jmlr.org/papers/volume21/20-074/20-074.pdf} {Exploring
  the limits of transfer learning with a unified text-to-text transformer}.
\newblock \emph{Journal of Machine Learning Research (JMLR)}, 21.

\bibitem[{Sachdeva et~al.(2024)Sachdeva, Coleman, Kang, Ni, Hong, Chi,
  Caverlee, McAuley, and Cheng}]{sachdeva2024train}
Noveen Sachdeva, Benjamin Coleman, Wang-Cheng Kang, Jianmo Ni, Lichan Hong,
  Ed~H Chi, James Caverlee, Julian McAuley, and Derek~Zhiyuan Cheng. 2024.
\newblock \href {https://arxiv.org/abs/2402.09668} {How to train data-efficient
  llms}.
\newblock \emph{arXiv preprint arXiv:2402.09668}.

\bibitem[{Sturua et~al.(2024)Sturua, Mohr, Akram, Günther, Wang, Krimmel,
  Wang, Mastrapas, Koukounas, Koukounas, Wang, and
  Xiao}]{sturua2024jinaembeddingsv3multilingualembeddingstask}
Saba Sturua, Isabelle Mohr, Mohammad~Kalim Akram, Michael Günther, Bo~Wang,
  Markus Krimmel, Feng Wang, Georgios Mastrapas, Andreas Koukounas, Andreas
  Koukounas, Nan Wang, and Han Xiao. 2024.
\newblock \href {https://arxiv.org/abs/2409.10173} {jina-embeddings-v3:
  Multilingual embeddings with task lora}.
\newblock \emph{arXiv preprint arXiv:2409.10173}.

\bibitem[{Su et~al.(2024)Su, Kong, Lin, Jennings, Norick, Kliegl, Patwary,
  Shoeybi, and Catanzaro}]{DBLP:journals/corr/abs-2412-02595}
Dan Su, Kezhi Kong, Ying Lin, Joseph Jennings, Brandon Norick, Markus Kliegl,
  Mostofa Patwary, Mohammad Shoeybi, and Bryan Catanzaro. 2024.
\newblock \href {https://arxiv.org/abs/2412.02595} {Nemotron-cc: Transforming
  common crawl into a refined long-horizon pretraining dataset}.
\newblock \emph{arXiv preprint arXiv:2412.02595}.

\bibitem[{Team et~al.(2025)Team, Kamath, Ferret, Pathak, Vieillard, Merhej,
  Perrin, Matejovicova, Ramé, Rivière, Rouillard, Mesnard, Cideron, bastien
  Grill, Ramos, Yvinec, Casbon, Pot, Penchev, Liu, Visin, Kenealy, Beyer, Zhai,
  Tsitsulin, Busa-Fekete, Feng, Sachdeva, Coleman, Gao, Mustafa, Barr,
  Parisotto, Tian, Eyal, Cherry, Peter, Sinopalnikov, Bhupatiraju, Agarwal,
  Kazemi, Malkin, Kumar, Vilar, Brusilovsky, Luo, Steiner, Friesen, Sharma,
  Sharma, Gilady, Goedeckemeyer, Saade, Feng, Kolesnikov, Bendebury, Abdagic,
  Vadi, György, Pinto, Das, Bapna, Miech, Yang, Paterson, Shenoy, Chakrabarti,
  Piot, Wu, Shahriari, Petrini, Chen, Lan, Choquette-Choo, Carey, Brick,
  Deutsch, Eisenbud, Cattle, Cheng, Paparas, Sreepathihalli, Reid, Tran, Zelle,
  Noland, Huizenga, Kharitonov, Liu, Amirkhanyan, Cameron, Hashemi,
  Klimczak-Plucińska, Singh, Mehta, Lehri, Hazimeh, Ballantyne, Szpektor,
  Nardini, Pouget-Abadie, Chan, Stanton, Wieting, Lai, Orbay, Fernandez,
  Newlan, yeong Ji, Singh, Black, Yu, Hui, Vodrahalli, Greff, Qiu, Valentine,
  Coelho, Ritter, Hoffman, Watson, Chaturvedi, Moynihan, Ma, Babar, Noy, Byrd,
  Roy, Momchev, Chauhan, Sachdeva, Bunyan, Botarda, Caron, Rubenstein,
  Culliton, Schmid, Sessa, Xu, Stanczyk, Tafti, Shivanna, Wu, Pan, Rokni,
  Willoughby, Vallu, Mullins, Jerome, Smoot, Girgin, Iqbal, Reddy, Sheth,
  Põder, Bhatnagar, Panyam, Eiger, Zhang, Liu, Yacovone, Liechty, Kalra, Evci,
  Misra, Roseberry, Feinberg, Kolesnikov, Han, Kwon, Chen, Chow, Zhu, Wei,
  Egyed, Cotruta, Giang, Kirk, Rao, Black, Babar, Lo, Moreira, Martins,
  Sanseviero, Gonzalez, Gleicher, Warkentin, Mirrokni, Senter, Collins, Barral,
  Ghahramani, Hadsell, Matias, Sculley, Petrov, Fiedel, Shazeer, Vinyals, Dean,
  Hassabis, Kavukcuoglu, Farabet, Buchatskaya, Alayrac, Anil, Dmitry, Lepikhin,
  Borgeaud, Bachem, Joulin, Andreev, Hardin, Dadashi, and
  Hussenot}]{gemmateam2025gemma3technicalreport}
Gemma Team, Aishwarya Kamath, Johan Ferret, Shreya Pathak, Nino Vieillard,
  Ramona Merhej, Sarah Perrin, Tatiana Matejovicova, Alexandre Ramé, Morgane
  Rivière, Louis Rouillard, Thomas Mesnard, Geoffrey Cideron, Jean bastien
  Grill, Sabela Ramos, Edouard Yvinec, Michelle Casbon, Etienne Pot, Ivo
  Penchev, Gaël Liu, Francesco Visin, Kathleen Kenealy, Lucas Beyer, Xiaohai
  Zhai, Anton Tsitsulin, Robert Busa-Fekete, Alex Feng, Noveen Sachdeva,
  Benjamin Coleman, Yi~Gao, Basil Mustafa, Iain Barr, Emilio Parisotto, David
  Tian, Matan Eyal, Colin Cherry, Jan-Thorsten Peter, Danila Sinopalnikov,
  Surya Bhupatiraju, Rishabh Agarwal, Mehran Kazemi, Dan Malkin, Ravin Kumar,
  David Vilar, Idan Brusilovsky, Jiaming Luo, Andreas Steiner, Abe Friesen,
  Abhanshu Sharma, Abheesht Sharma, Adi~Mayrav Gilady, Adrian Goedeckemeyer,
  Alaa Saade, Alex Feng, Alexander Kolesnikov, Alexei Bendebury, Alvin Abdagic,
  Amit Vadi, András György, André~Susano Pinto, Anil Das, Ankur Bapna,
  Antoine Miech, Antoine Yang, Antonia Paterson, Ashish Shenoy, Ayan
  Chakrabarti, Bilal Piot, Bo~Wu, Bobak Shahriari, Bryce Petrini, Charlie Chen,
  Charline~Le Lan, Christopher~A. Choquette-Choo, CJ~Carey, Cormac Brick,
  Daniel Deutsch, Danielle Eisenbud, Dee Cattle, Derek Cheng, Dimitris Paparas,
  Divyashree~Shivakumar Sreepathihalli, Doug Reid, Dustin Tran, Dustin Zelle,
  Eric Noland, Erwin Huizenga, Eugene Kharitonov, Frederick Liu, Gagik
  Amirkhanyan, Glenn Cameron, Hadi Hashemi, Hanna Klimczak-Plucińska, Harman
  Singh, Harsh Mehta, Harshal~Tushar Lehri, Hussein Hazimeh, Ian Ballantyne,
  Idan Szpektor, Ivan Nardini, Jean Pouget-Abadie, Jetha Chan, Joe Stanton,
  John Wieting, Jonathan Lai, Jordi Orbay, Joseph Fernandez, Josh Newlan,
  Ju~yeong Ji, Jyotinder Singh, Kat Black, Kathy Yu, Kevin Hui, Kiran
  Vodrahalli, Klaus Greff, Linhai Qiu, Marcella Valentine, Marina Coelho,
  Marvin Ritter, Matt Hoffman, Matthew Watson, Mayank Chaturvedi, Michael
  Moynihan, Min Ma, Nabila Babar, Natasha Noy, Nathan Byrd, Nick Roy, Nikola
  Momchev, Nilay Chauhan, Noveen Sachdeva, Oskar Bunyan, Pankil Botarda, Paul
  Caron, Paul~Kishan Rubenstein, Phil Culliton, Philipp Schmid, Pier~Giuseppe
  Sessa, Pingmei Xu, Piotr Stanczyk, Pouya Tafti, Rakesh Shivanna, Renjie Wu,
  Renke Pan, Reza Rokni, Rob Willoughby, Rohith Vallu, Ryan Mullins, Sammy
  Jerome, Sara Smoot, Sertan Girgin, Shariq Iqbal, Shashir Reddy, Shruti Sheth,
  Siim Põder, Sijal Bhatnagar, Sindhu~Raghuram Panyam, Sivan Eiger, Susan
  Zhang, Tianqi Liu, Trevor Yacovone, Tyler Liechty, Uday Kalra, Utku Evci,
  Vedant Misra, Vincent Roseberry, Vlad Feinberg, Vlad Kolesnikov, Woohyun Han,
  Woosuk Kwon, Xi~Chen, Yinlam Chow, Yuvein Zhu, Zichuan Wei, Zoltan Egyed,
  Victor Cotruta, Minh Giang, Phoebe Kirk, Anand Rao, Kat Black, Nabila Babar,
  Jessica Lo, Erica Moreira, Luiz~Gustavo Martins, Omar Sanseviero, Lucas
  Gonzalez, Zach Gleicher, Tris Warkentin, Vahab Mirrokni, Evan Senter, Eli
  Collins, Joelle Barral, Zoubin Ghahramani, Raia Hadsell, Yossi Matias,
  D.~Sculley, Slav Petrov, Noah Fiedel, Noam Shazeer, Oriol Vinyals, Jeff Dean,
  Demis Hassabis, Koray Kavukcuoglu, Clement Farabet, Elena Buchatskaya,
  Jean-Baptiste Alayrac, Rohan Anil, Dmitry, Lepikhin, Sebastian Borgeaud,
  Olivier Bachem, Armand Joulin, Alek Andreev, Cassidy Hardin, Robert Dadashi,
  and Léonard Hussenot. 2025.
\newblock \href {https://arxiv.org/abs/2503.19786} {Gemma 3 technical report}.
\newblock \emph{arXiv preprint arXiv:2503.19786}.

\bibitem[{Touvron et~al.(2023)Touvron, Martin, Stone, Albert, Almahairi,
  Babaei, Bashlykov, Batra, Bhargava, Bhosale, Bikel, Blecher, Ferrer, Chen,
  Cucurull, Esiobu, Fernandes, Fu, Fu, Fuller, Gao, Goswami, Goyal, Hartshorn,
  Hosseini, Hou, Inan, Kardas, Kerkez, Khabsa, Kloumann, Korenev, Koura,
  Lachaux, Lavril, Lee, Liskovich, Lu, Mao, Martinet, Mihaylov, Mishra,
  Molybog, Nie, Poulton, Reizenstein, Rungta, Saladi, Schelten, Silva, Smith,
  Subramanian, Tan, Tang, Taylor, Williams, Kuan, Xu, Yan, Zarov, Zhang, Fan,
  Kambadur, Narang, Rodriguez, Stojnic, Edunov, and
  Scialom}]{touvron2023llama2}
Hugo Touvron, Louis Martin, Kevin Stone, Peter Albert, Amjad Almahairi, Yasmine
  Babaei, Nikolay Bashlykov, Soumya Batra, Prajjwal Bhargava, Shruti Bhosale,
  Dan Bikel, Lukas Blecher, Cristian~Canton Ferrer, Moya Chen, Guillem
  Cucurull, David Esiobu, Jude Fernandes, Jeremy Fu, Wenyin Fu, Brian Fuller,
  Cynthia Gao, Vedanuj Goswami, Naman Goyal, Anthony Hartshorn, Saghar
  Hosseini, Rui Hou, Hakan Inan, Marcin Kardas, Viktor Kerkez, Madian Khabsa,
  Isabel Kloumann, Artem Korenev, Punit~Singh Koura, Marie-Anne Lachaux,
  Thibaut Lavril, Jenya Lee, Diana Liskovich, Yinghai Lu, Yuning Mao, Xavier
  Martinet, Todor Mihaylov, Pushkar Mishra, Igor Molybog, Yixin Nie, Andrew
  Poulton, Jeremy Reizenstein, Rashi Rungta, Kalyan Saladi, Alan Schelten, Ruan
  Silva, Eric~Michael Smith, Ranjan Subramanian, Xiaoqing~Ellen Tan, Binh Tang,
  Ross Taylor, Adina Williams, Jian~Xiang Kuan, Puxin Xu, Zheng Yan, Iliyan
  Zarov, Yuchen Zhang, Angela Fan, Melanie Kambadur, Sharan Narang, Aurelien
  Rodriguez, Robert Stojnic, Sergey Edunov, and Thomas Scialom. 2023.
\newblock Llama 2: Open foundation and fine-tuned chat models.
\newblock \emph{arXiv preprint arXiv:2307.09288}.

\bibitem[{Weber et~al.(2024)Weber, Fu, Anthony, Oren, Adams, Alexandrov, Lyu,
  Nguyen, Yao, Adams, Athiwaratkun, Chalamala, Chen, Ryabinin, Dao, Liang,
  R{\'{e}}, Rish, and Zhang}]{weber2024maurice}
Maurice Weber, Daniel~Y. Fu, Quentin Anthony, Yonatan Oren, Shane Adams, Anton
  Alexandrov, Xiaozhong Lyu, Huu Nguyen, Xiaozhe Yao, Virginia Adams, Ben
  Athiwaratkun, Rahul Chalamala, Kezhen Chen, Max Ryabinin, Tri Dao, Percy
  Liang, Christopher R{\'{e}}, Irina Rish, and Ce~Zhang. 2024.
\newblock Redpajama: an open dataset for training large language models.
\newblock In \emph{Proceedings of the Advances in Neural Information Processing
  Systems: Annual Conference on Neural Information Processing Systems
  ({NeurIPS})}.

\bibitem[{Wettig et~al.(2024)Wettig, Gupta, Malik, and
  Chen}]{DBLP:conf/icml/WettigGM024}
Alexander Wettig, Aatmik Gupta, Saumya Malik, and Danqi Chen. 2024.
\newblock \href {https://openreview.net/forum?id=GLGYYqPwjy} {Qurating:
  Selecting high-quality data for training language models}.
\newblock In \emph{Forty-first International Conference on Machine Learning,
  {ICML} 2024, Vienna, Austria, July 21-27, 2024}. OpenReview.net.

\bibitem[{Xue et~al.(2021)Xue, Constant, Roberts, Kale, Al{-}Rfou, Siddhant,
  Barua, and Raffel}]{xue2021mc4}
Linting Xue, Noah Constant, Adam Roberts, Mihir Kale, Rami Al{-}Rfou, Aditya
  Siddhant, Aditya Barua, and Colin Raffel. 2021.
\newblock mt5: {A} massively multilingual pre-trained text-to-text transformer.
\newblock In \emph{Proceedings of the Conference of the North American Chapter
  of the Association for Computational Linguistics: Human Language Technologies
  ({NAACL-HLT})}.

\bibitem[{Yu et~al.(2024)Yu, Merrick, Nuti, and
  Campos}]{yu2024arcticembed20multilingualretrieval}
Puxuan Yu, Luke Merrick, Gaurav Nuti, and Daniel Campos. 2024.
\newblock \href {https://arxiv.org/abs/2412.04506} {Arctic-embed 2.0:
  Multilingual retrieval without compromise}.
\newblock \emph{arXiv preprint arXiv:2412.04506}.

\bibitem[{Zellers et~al.(2019)Zellers, Holtzman, Bisk, Farhadi, and
  Choi}]{zellers2019hellaswag}
Rowan Zellers, Ari Holtzman, Yonatan Bisk, Ali Farhadi, and Yejin Choi. 2019.
\newblock Hellaswag: Can a machine really finish your sentence?
\newblock In \emph{Proceedings of the Annual Meeting of the Association for
  Computational Linguistics {(ACL)}}.

\bibitem[{Zhang et~al.(2024)Zhang, Zhang, Long, Xie, Dai, Tang, Lin, Yang, Xie,
  Huang et~al.}]{zhang2024mgte}
Xin Zhang, Yanzhao Zhang, Dingkun Long, Wen Xie, Ziqi Dai, Jialong Tang, Huan
  Lin, Baosong Yang, Pengjun Xie, Fei Huang, et~al. 2024.
\newblock mgte: Generalized long-context text representation and reranking
  models for multilingual text retrieval.
\newblock In \emph{Proceedings of the 2024 Conference on Empirical Methods in
  Natural Language Processing: Industry Track}, pages 1393--1412.

\bibitem[{Zhao et~al.(2024)Zhao, Thai, Zhang, Hu, Zhou, Ba, Cai, Liu, and
  Sun}]{DBLP:conf/emnlp/ZhaoTZHZBC0024}
Ranchi Zhao, Zhen~Leng Thai, Yifan Zhang, Shengding Hu, Jie Zhou, Yunqi Ba, Jie
  Cai, Zhiyuan Liu, and Maosong Sun. 2024.
\newblock Decoratelm: Data engineering through corpus rating, tagging, and
  editing with language models.
\newblock In \emph{{EMNLP}}, pages 1401--1418. Association for Computational
  Linguistics.

\end{thebibliography}

\appendix

\onecolumn

\section{Human Annotation Study}

\subsection{Annotator Background and Study Protocol}
For our human annotation study, we used the prompt introduced by~\citet{penedo2024fineweb-2}, which was reviewed and discussed with all annotators during a dedicated training session.
Annotations were conducted using a web interface built with Argilla\footnote{https://argilla.io/}, which displayed the document text, annotation guidelines, and the 0–5 rating scale.

Our annotators are colleagues from our lab, and there is an overlap between the authors of this work and the annotation team. The majority of annotators have a technical background. 
Additional information on annotators is provided in Table~\ref{tab:annotator_backgrounds}.
Prior to the study, we informed participants about the purpose of the annotation task and obtained their consent to use the resulting annotations, along with anonymized information about the annotators, for subsequent analysis and anonymized public release.
No ethics review board approval was sought, as the study did not fall under institutional requirements for ethical review.


\begin{table}[h]
    \centering
    \begin{tabular}{lll}
        \toprule
        \textbf{Annotator (Anonymized)} & \textbf{Background} & \ \textbf{Age Group} \\
        \midrule
        Annotator 1 & MSc. in Computer Science &  20-30\\
        Annotator 2 & MSc. in Data and Knowledge Engineering & 30-40 \\
        Annotator 3 & PhD in Computer Science & 30-40 \\
        Annotator 4 & M.A. English/American Studies and German Studies & 30-40 \\
        Annotator 5 & M.Sc. in Mathematics & 30-40\\
        Annotator 6 & PhD in Computer Science & 30-40\\
        Annotator 7 & M.Sc. in Artificial Intelligence & 20-30\\
        Annotator 8 & PhD in Computer Science & 30-40 \\
        Annotator 9 & MSc. in Computer Science & 30-40\\
        Annotator 10 & MSc. in Computer Science & 30-30\\
        Annotator 11 & PhD in Theoretical Physics & 30-40 \\
        Annotator 12 & MSc. in Autonomous Systems & 30.40\\
        Annotator 13 & PhD in Computer Science & 30-40\\
        Annotator 14 & MSc. in Autonomous Systems  & 30-40\\
        Annotator 15 & MSc. in Computer Science & 30-40\\
        \bottomrule
    \end{tabular}
    \caption{Backgrounds of the human annotators (anonymized).}
    \label{tab:annotator_backgrounds}
\end{table}

\subsection{Human Annotations Evaluation}\label{app:ground_truth}

In this section, we provide additional details about the human-annotated ground truth dataset introduced in Section~\ref{sec:ground_truth}.

\paragraph{Score Distribution of Annotations}

\begin{figure}[h]
\centering
\includegraphics[width=.5\textwidth]{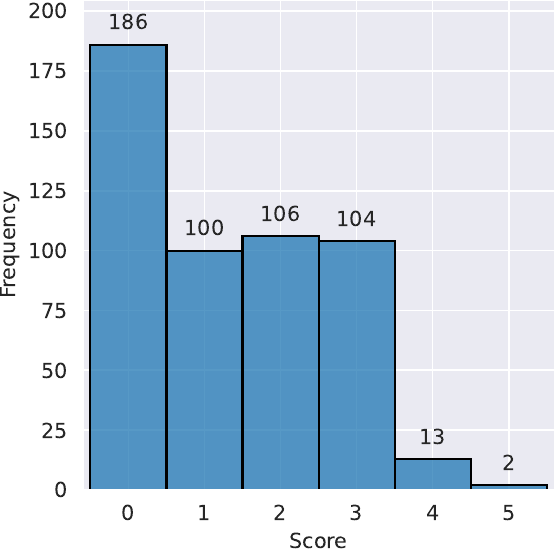}
\caption{Histogram on the distribution of the document scores judged by the human annotators.}
\label{fig:human_annotations_histogram}
\end{figure}

\begin{figure}[h]
\centering
\includegraphics[width=.7\textwidth]{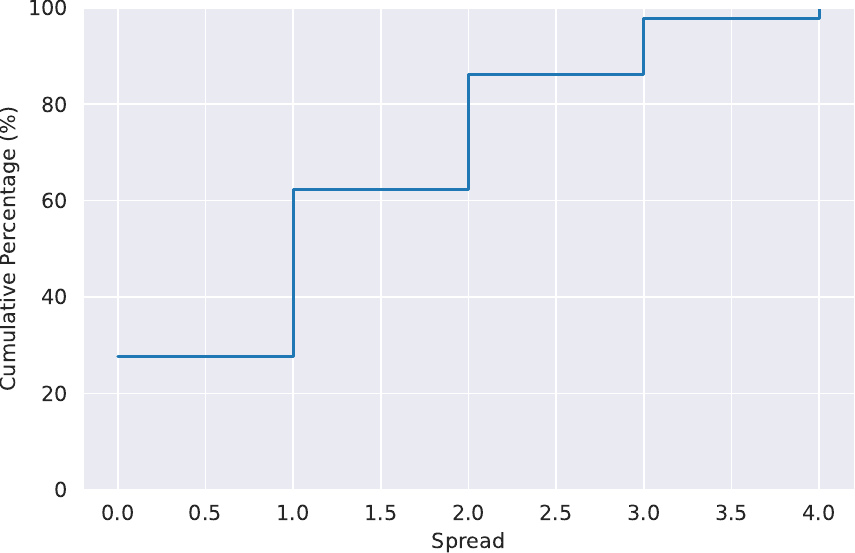}
\caption{Cumulative distribution of spread within annotations. Aligned with the majority agreement of 78.5\% and an interrating standard deviation of 0.56, (see Sec.\,\ref{sec:ground_truth}), also the spread analysis reveals high interrater consistency with a spread of $\leq$ 2 for 86\% of the documents.}
\label{fig:cumulative_distribution_spread}
\end{figure}

\paragraph{Annotator Agreement and Annotation Spread.}
To further analyze the variation in human annotations, we present the cumulative distribution of annotation spread in Figure~\ref{fig:cumulative_distribution_spread}. The plot shows that over 60\% of the samples have a maximum spread of 1, and more than 85\% have a maximum spread of 2, indicating strong agreement among annotators.


\clearpage

\section{LLM Based Annotator Evaluation}\label{app:llms}

In this Section we provide further details and ablations on our LLM based annotators discussed in Section~\ref{sec:llms}.

\subsection{Invalid Predictions}

Similar to the human annotators, we prompted the LLM-based annotators to assess the educational value of documents on a scale from 0 to 5, where 0 indicates the lowest quality and 5 the highest. For each model and document, we collected three predictions.
A prediction is considered invalid if it does not fall within the specified integer range. If all three predictions for a document are invalid, the entire annotation is marked as invalid.
When evaluating LLM performance, it is crucial to analyze the distribution of valid and invalid predictions to not obtain distorted conclusions.

\begin{figure*}[t]
\centering
\includegraphics[width=1\textwidth]{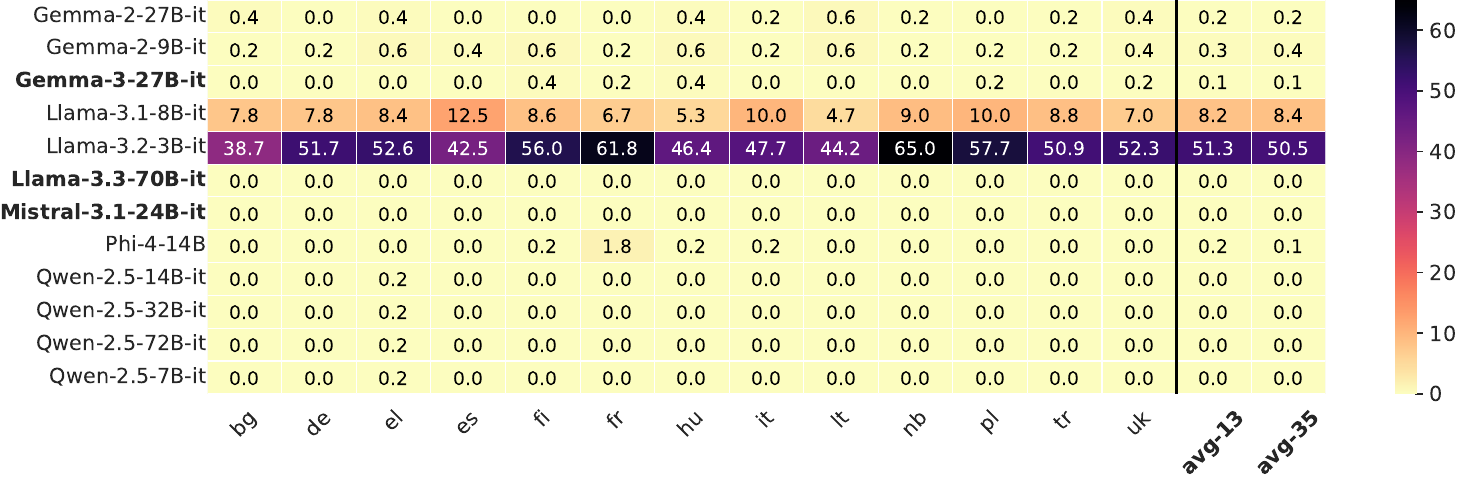}
\caption{Invalid scores predictions (in percent)}
\label{fig:llm_annotator_invalid_scores_predictions}
\end{figure*}

Figure~\ref{fig:llm_annotator_invalid_scores_predictions} shows the proportion of invalid predictions across different languages. 
While our selected models, LLaMA-3-70B-IT, Mistral-3.1-24B-IT, and Gemma-3-27B-IT, exhibit few or no invalid predictions, LLaMA-3-8B-IT produces a noticeably higher rate of invalid outputs, and LLaMA-3-3B-IT shows a substantial fraction of invalid predictions.

Based on these observations, we suggest that a consistently low rate of invalid predictions should be considered a necessary condition for further use as LLM-based annotator.
Otherwise, annotating data at scale will result in a large number of invalid predictions, leading to wasted computational resources.

\begin{figure*}[t]
\centering
\includegraphics[width=0.7\textwidth]{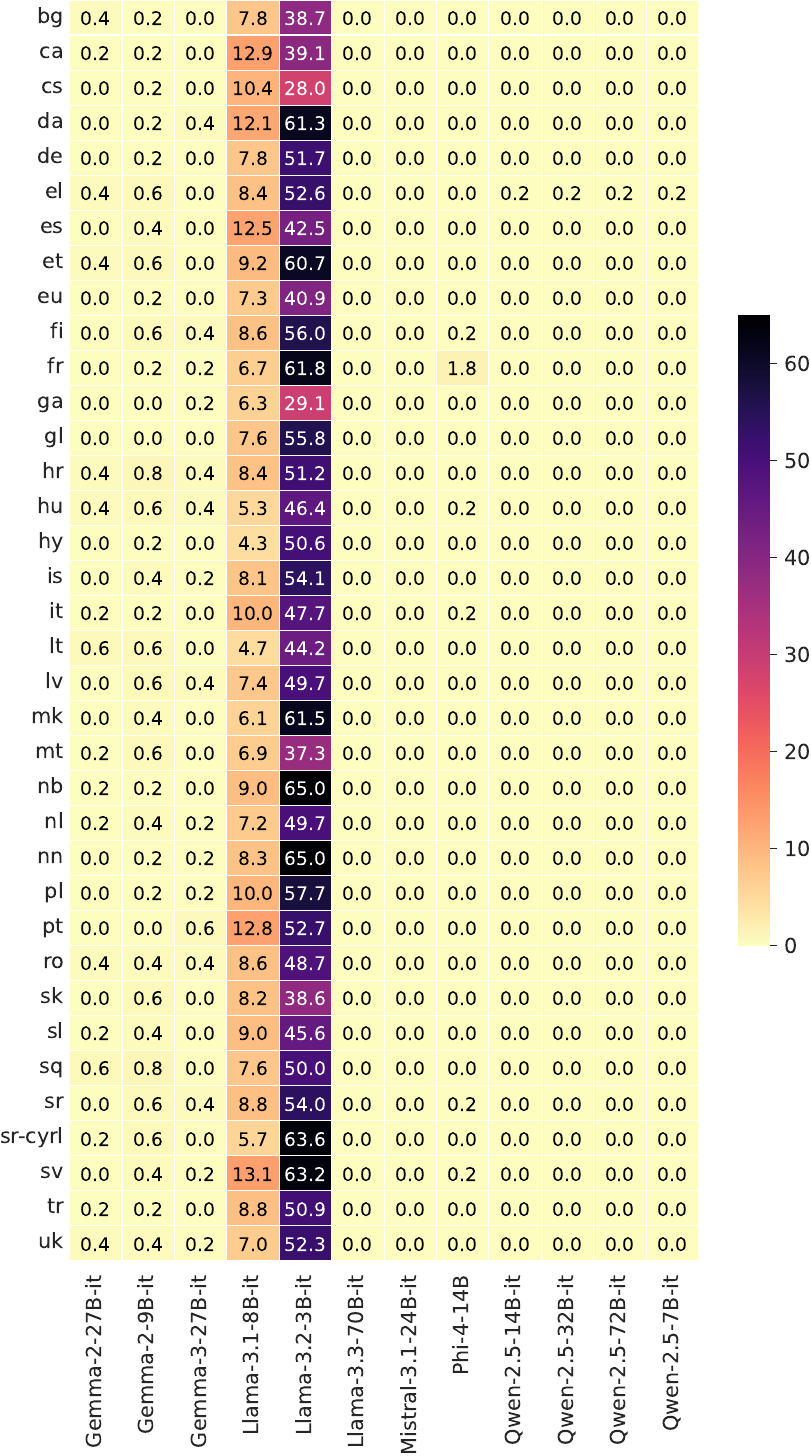}
\caption{Percentages of invalid scores (aggregated) for each model across all languages. An aggregated score (majority voted) is counted as invalid, if all three predictions for a document are invalid.}
\label{fig:}
\end{figure*}

\subsection{Statistical Significance of Correlations Between Human Annotations and LLM Predictions.}

To assess the statistical significance of the correlations presented in Fig.\ref{fig:classification_performance_llm_annotators}, we perform two-sided Student’s t-tests and compute the corresponding p-values separately for each model and language. 
Summary statistics, i.e., average, minimum, and maximum p-values, across the 35 languages are shown in Fig.\ref{fig:p_values}.

Notably, the highest p-value observed across all models and languages is 4.49e-07, indicating a consistently high level of statistical significance throughout our analysis.

\begin{table}[t]
\centering
\begin{tabular}{lrrr}
\toprule
 \textbf{LLM} & \textbf{avg} & \textbf{min} & \textbf{max} \\
\midrule
\textbf{Gemma-2-27B-it} & 1.51e-52 & 5.76e-68 & 5.38e-51 \\
\textbf{Gemma-2-9B-it} & 1.43e-61 & 2.77e-76 & 5.16e-60 \\
\textbf{Gemma-3-27B-it} & 8.38e-65 & 1.09e-85 & 3.02e-63 \\
\textbf{Llama-3.1-8B-it} & 8.90e-51 & 3.22e-73 & 3.01e-49 \\
\textbf{Llama-3.2-3B-it} & 2.04e-08 & 6.42e-27 & 4.49e-07 \\
\textbf{Llama-3.3-70B-it} & 4.06e-66 & 3.54e-76 & 1.07e-64 \\
\textbf{Mistral-3.1-24B-it} & 4.59e-62 & 2.89e-81 & 1.61e-60 \\
\textbf{Phi-4-14B} & 4.26e-46 & 2.02e-65 & 1.53e-44 \\
\textbf{Qwen-2.5-14B-it} & 1.73e-37 & 1.88e-56 & 6.22e-36 \\
\textbf{Qwen-2.5-32B-it} & 4.12e-54 & 1.68e-68 & 1.48e-52 \\
\textbf{Qwen-2.5-72B-it} & 4.18e-53 & 7.90e-64 & 1.39e-51 \\
\textbf{Qwen-2.5-7B-it} & 1.24e-43 & 4.28e-68 & 4.46e-42 \\
\bottomrule
\end{tabular}
\caption{p-value analysis on the Spearman correlation scores in Figure \ref{fig:classification_performance_llm_annotators}. The p-values were calculated using a two-sided Student's t-test and indicate the statistical significance of the measured correlations (lower is better). Across all models and languages, even the highest p-values are extremely small. This underpins the statistical significance of our results.}
\label{fig:p_values}
\end{table}

\subsection{Classification Based Evaluation}\label{app:llm_classification}

As discussed in Sec.~\ref{sec:metric_lm_selection}, we use the Spearman correlation between the LLMs’ predictions and the human ground truth to evaluate the annotator capabilities of the models. This metric is preferred because it effectively captures the models’ ability to rank document quality, which is central to our task.

\begin{figure*}[t]
\centering
\includegraphics[width=1\textwidth]{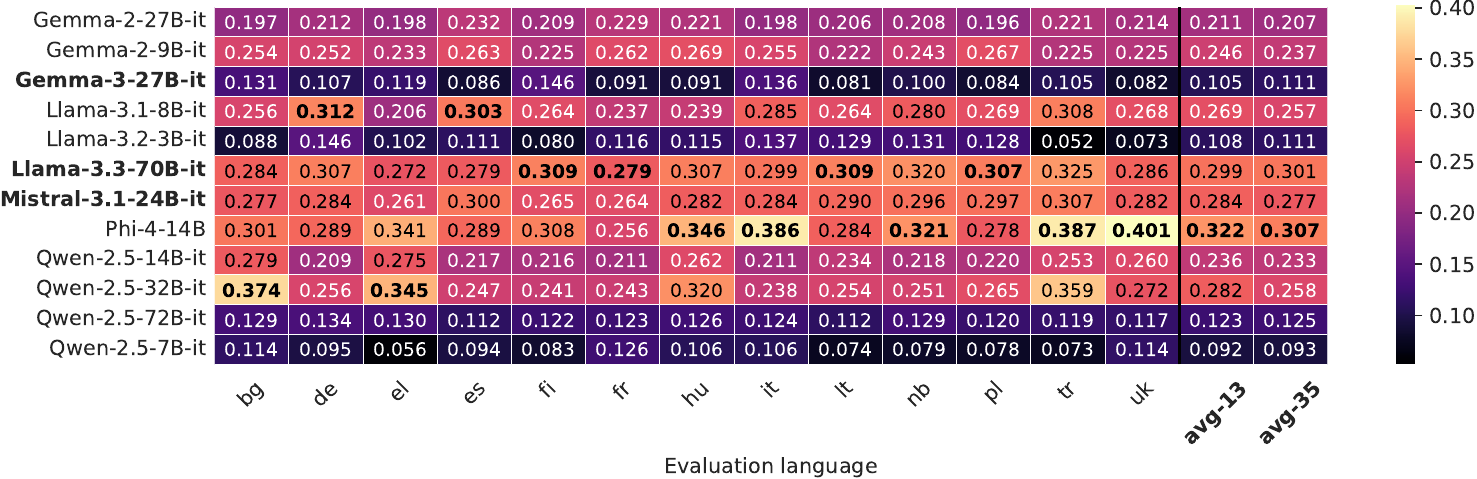}
\caption{Multilingual LLM classification performance (macro F1-score) on human-annotated ground truth. Scores are reported individually for the 13 languages subset, as well as averaged across these 13 languages (avg-13) and across all 35 evaluated languages.}
\label{fig:classification_performance_llm_annotators}
\end{figure*}

\begin{figure*}[h]
\centering
\includegraphics[width=0.7\textwidth]{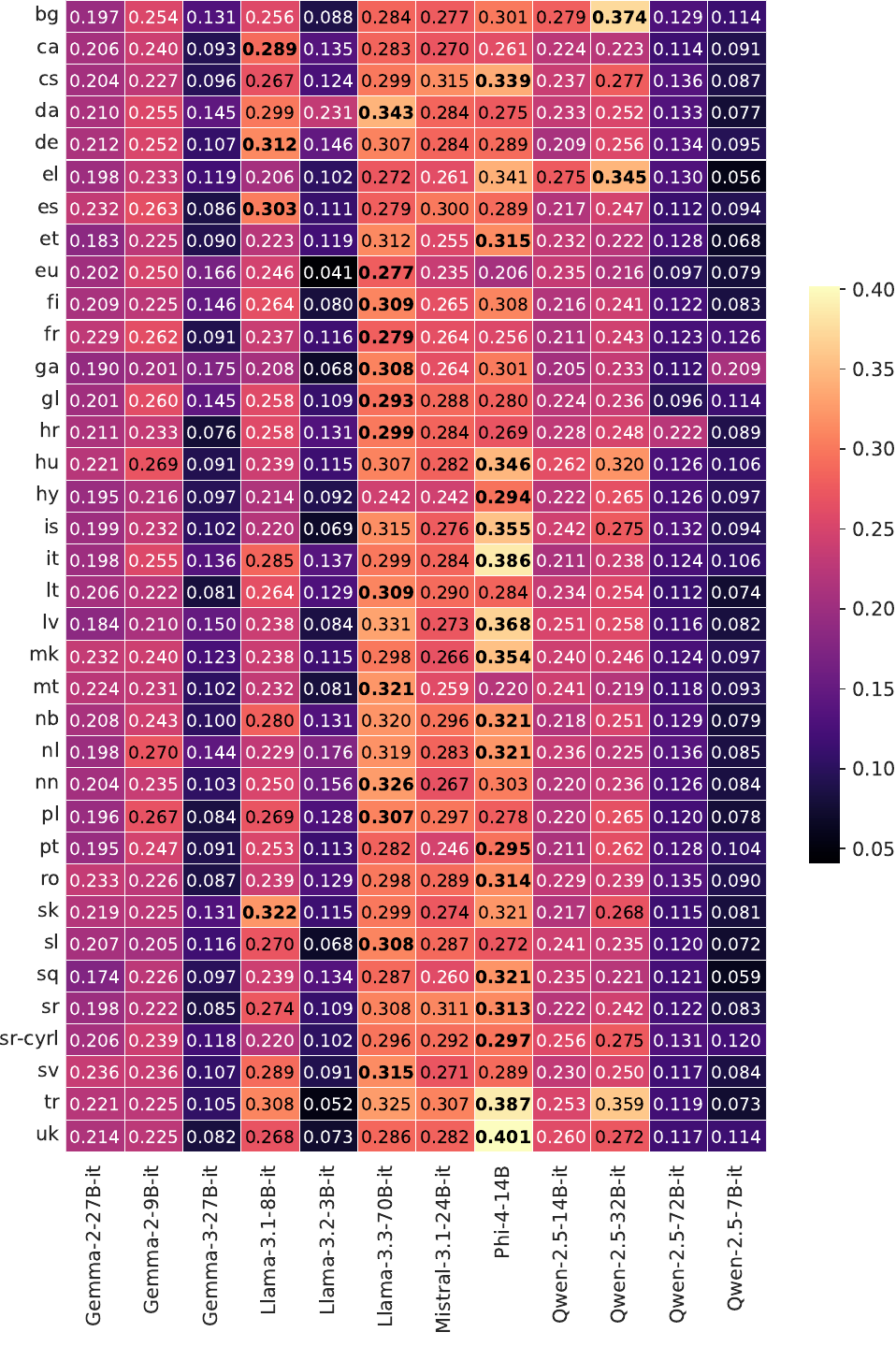}
\caption{Classification performance in terms of macro F1 score for each model across all languages.}
\label{fig:f1_performance_llm_annotators_full}
\end{figure*}

Here, we illustrate the limitations of traditional classification metrics for assessing LLM annotator performance. The figures~\ref{fig:classification_performance_llm_annotators} and~\ref{fig:f1_performance_llm_annotators_full} show the F1 scores of the LLMs when predicting the correct quality classes (0 to 5). Notably, Gemma-3-27B-IT appears among the worst-performing models in terms of F1 score, suggesting a limited ability to classify document quality. This stands in contrast to its relatively strong performance when evaluated using Spearman correlation (see Sec.~\ref{sec:llm_evaluation}).

\begin{figure}[t]
  \centering
  \begin{subfigure}[t]{0.32\textwidth}
    \includegraphics[width=\linewidth]{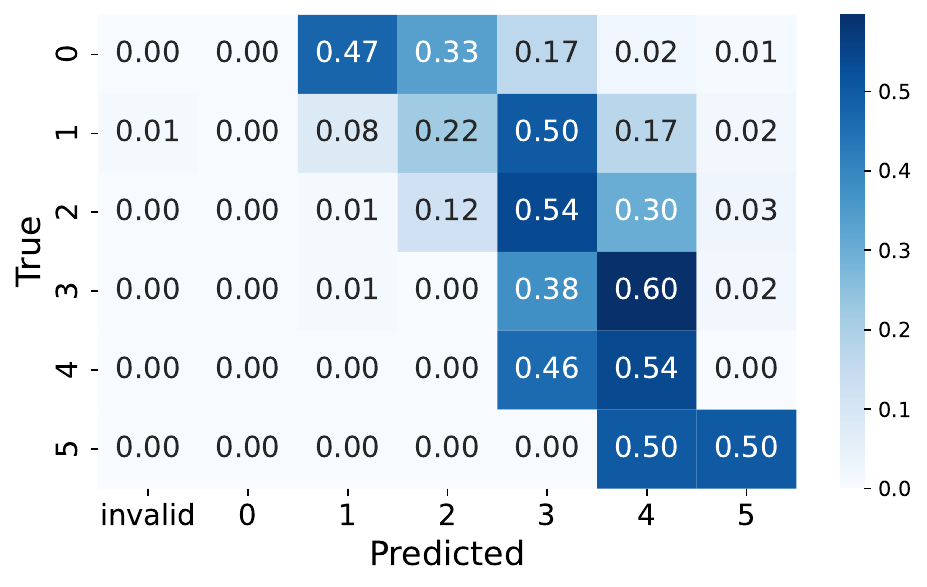}
    \caption{Gemma-3-27B-it}
  \end{subfigure}
  \hfill
  \begin{subfigure}[t]{0.32\textwidth}
    \includegraphics[width=\linewidth]{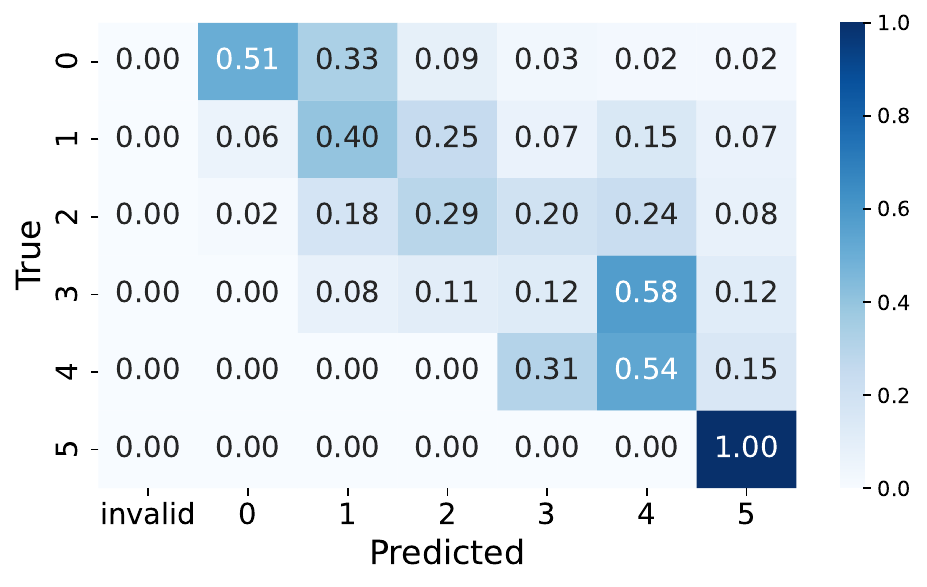}
    \caption{Llama-3.3-70B-it}
  \end{subfigure}
  \hfill
  \begin{subfigure}[t]{0.32\textwidth}
    \includegraphics[width=\linewidth]{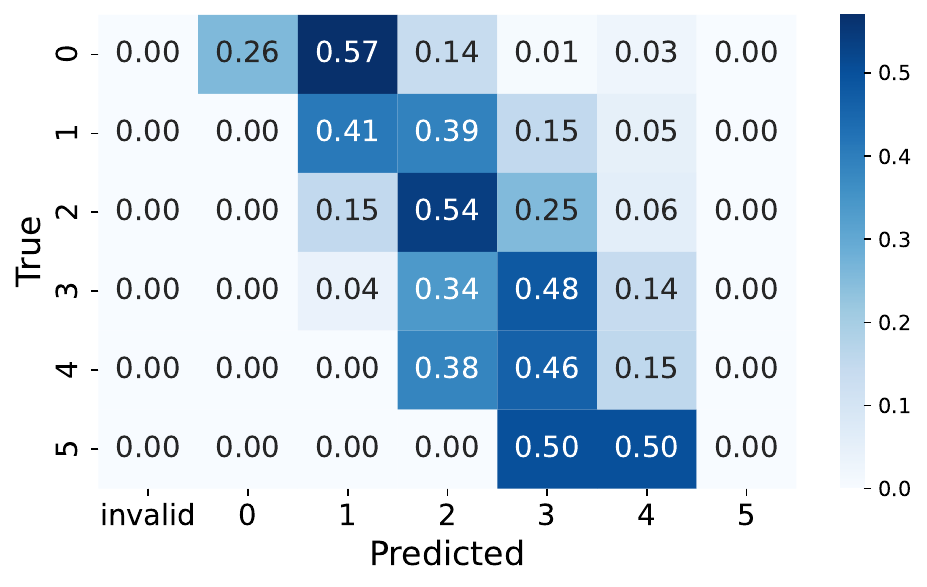}
    \caption{Mistral-3.1-24B-it}
  \end{subfigure}
  \caption{Confusion matrices of the three ablated LLMs on the 511 human annotated ground truth documents in English. Note that Gemma-3-27B-IT predictions tend to be shifted by 1 to the right which degrades the classification accuracy but does not influence the ranking performance. Both LLama-3.3-70B and Mistral-Small-3.1-24B are well aligned with the human annotations, explaining the high classification accuracy.}
  \label{fig:confusion_matrices_llms}
\end{figure}

This discrepancy can be explained by examining the confusion matrices in Fig.~\ref{fig:confusion_matrices_llms}. 
While Mistral-3.1-24B tends to predict more reliably within the central quality classes (1 to 3), Gemma-3-27B-IT shows a tendency to shift predictions across the scale, particularly within these same classes. 
As a result, its F1 scores are low due to class misalignment, but its Spearman correlation remains high because it preserves the relative ranking of document quality.

\begin{figure*}[t]
\centering
\includegraphics[width=0.7\textwidth]{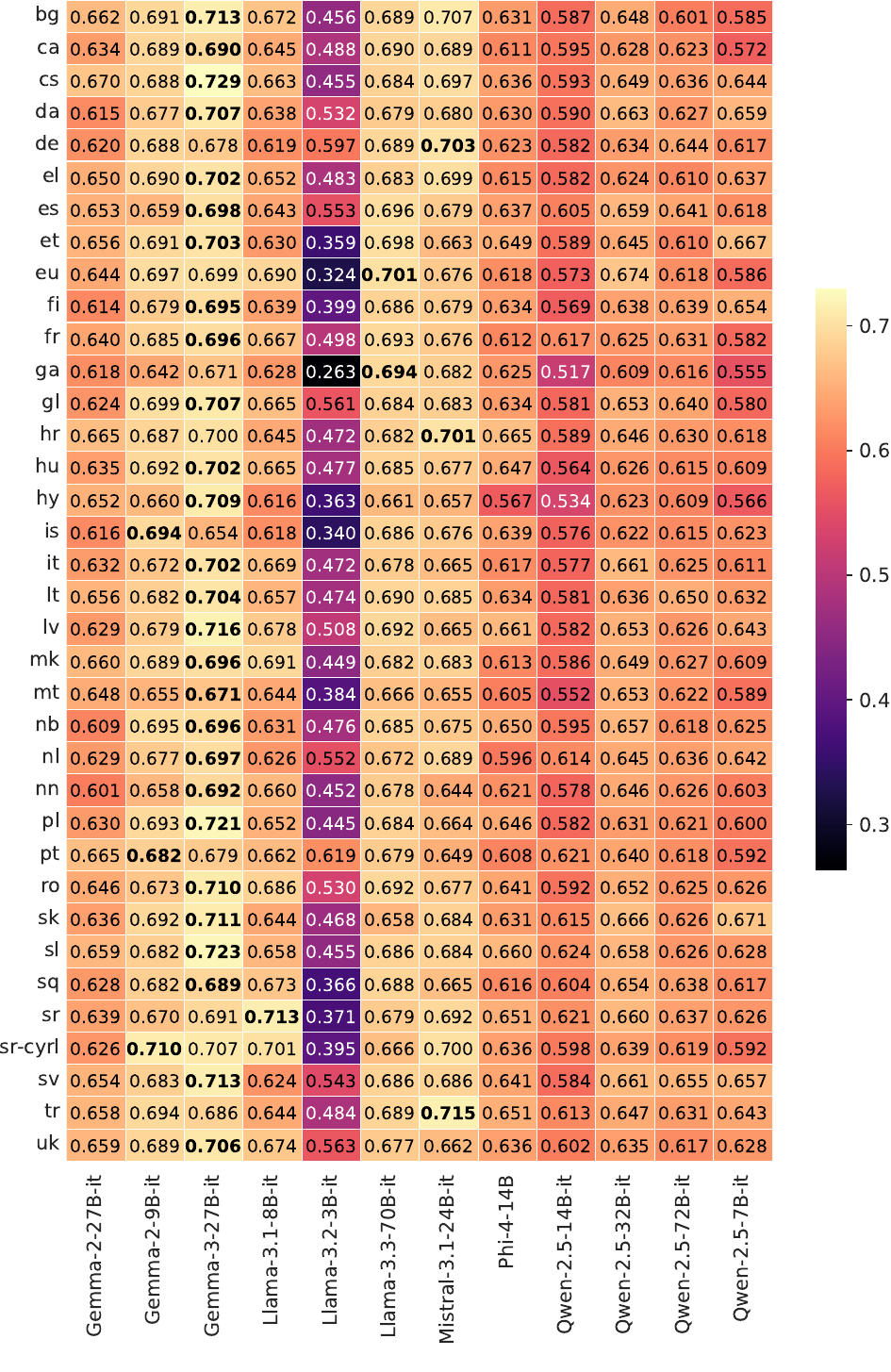}
\caption{Ranking performance in terms of Spearman correlation for each model across all languages.}
\label{fig:ranking_performance_llm_annotators_full}
\end{figure*}

\subsection{Predicted Annotation Distributions Across LLM Based Annotators}

In Sec.~\ref{app:llm_classification}, we showed using predictions from Gemma-3-27B-IT that different models can shift their predictions across the quality scale. This has important implications for selecting thresholds when filtering documents based on predicted quality.

\begin{figure}[t]
  \centering
  \begin{subfigure}[t]{0.32\textwidth}
    \includegraphics[width=\linewidth]{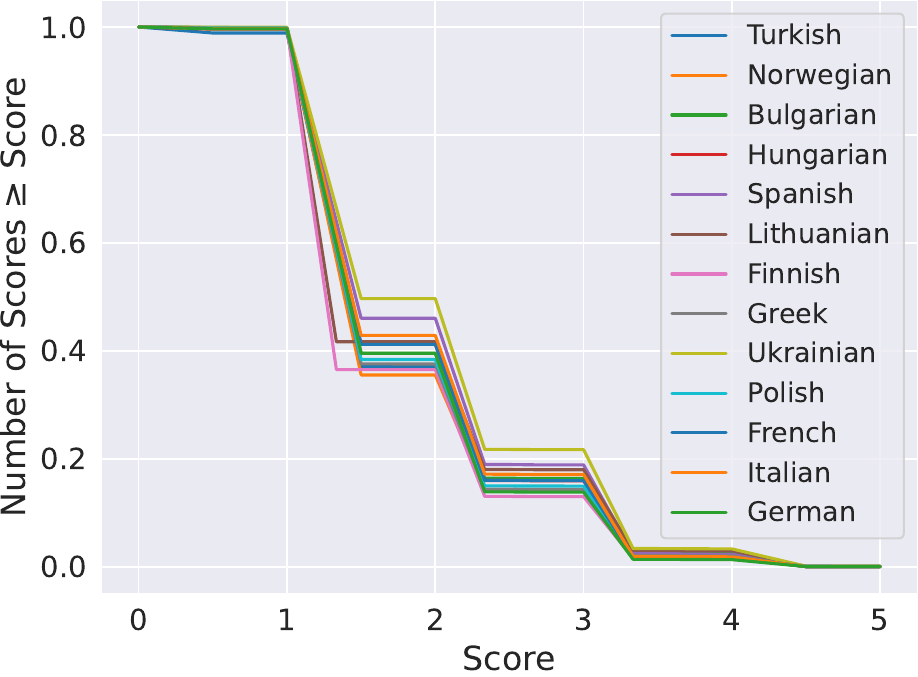}
    \caption{Gemma-3-27B-it}
  \end{subfigure}
  \hfill
  \begin{subfigure}[t]{0.32\textwidth}
    \includegraphics[width=\linewidth]{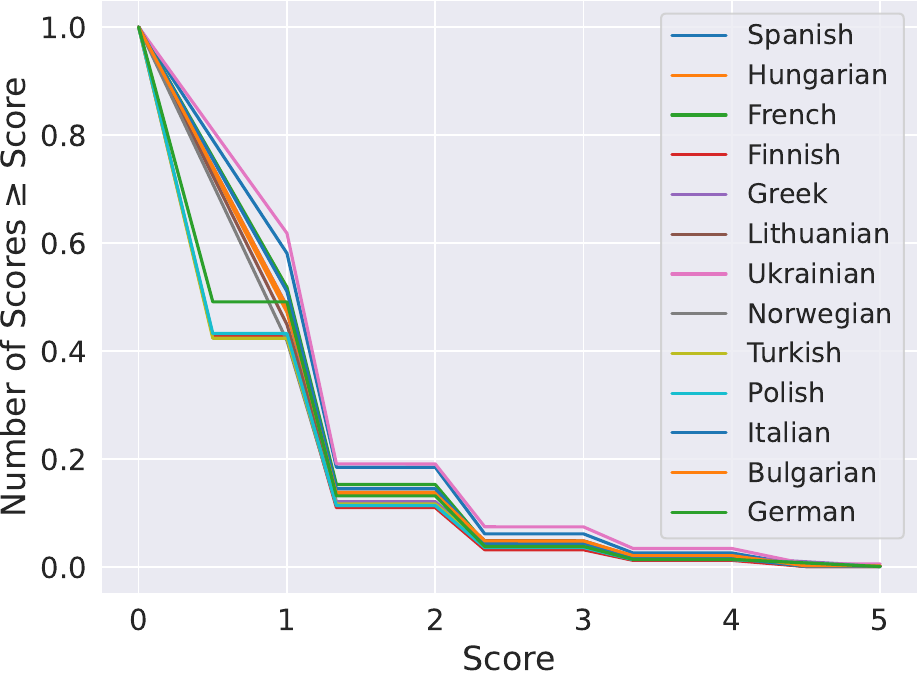}
    \caption{LLama-3.3-70B-it}
  \end{subfigure}
  \hfill
  \begin{subfigure}[t]{0.32\textwidth}
    \includegraphics[width=\linewidth]{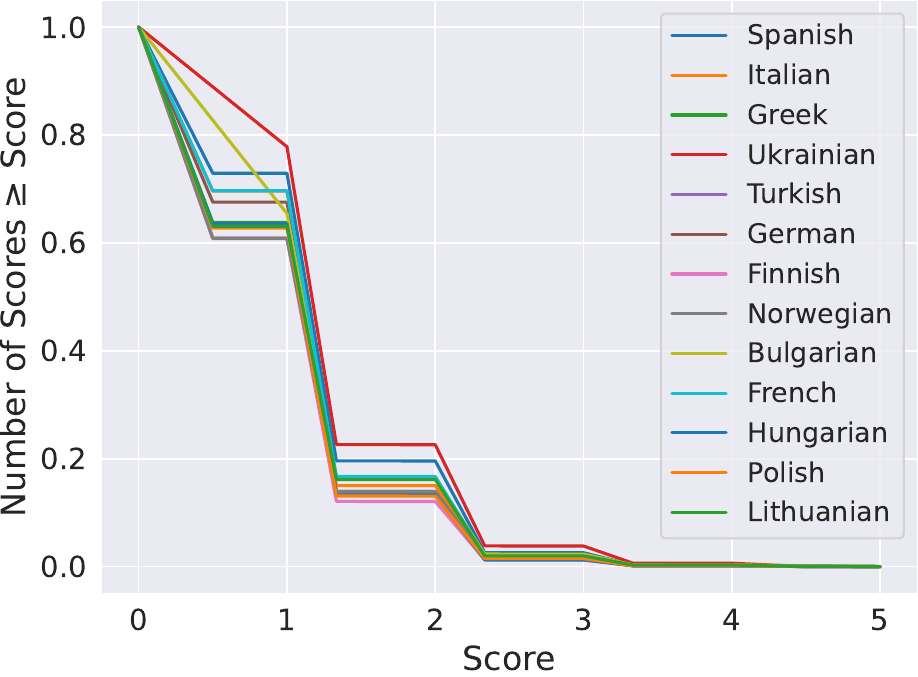}
    \caption{Mistral-3.1-24B-it}
  \end{subfigure}
  \caption{Right cumulative distribution of the scores predicted by the three ablated models. Alternatively, the curves can be interpreted as the number of documents whose scores is greater or equal to the given score. Note that the differences in the monotonously decreasing curves between models, motivates the model-specific threshold for pre-training data sampling. Notably, we found a Spearman correlation of 0.83 between the three models, indicating similar ranking orders despite the scale shifts.}
  \label{fig:llm_annotator_cumulative_score_dist}
\end{figure}

Figure~\ref{fig:llm_annotator_cumulative_score_dist} shows the cumulative distribution of predicted scores for annotated training datasets (approximately 450k documents per language) by Gemma-3-27B-IT, LLaMA-3.3-70B, and Mistral-Small-3.1-24B. We observe that, for a fixed filtering threshold, different models yield varying amounts of retained data. For example, with a threshold of $\geq3$, Gemma-3-27B-IT retains more data than the other two models, while LLaMA-3.3-70B retains more than Mistral-Small-3.1-24B. This highlights that the threshold is model-specific, effectively determining how much data is preserved and raising questions about the quality–quantity trade-off.

To address this, we advocate using the p-quantile rather than a fixed absolute threshold, ensuring consistent data retention across models. The high Spearman correlation (0.83) between the predicted scores of the three models indicates that, despite differences in absolute scoring, all models are capable of ranking documents by quality reliably.

\begin{table}
\centering
\begin{tabular}{lllrr}
\toprule
Language & Code & Translator & \#Testsamples & \#Trainsamples \\
\midrule
Bulgarian & bg & DeepL & 511 & 499.799 \\
Czech & cs & DeepL & 511 & 496.428 \\
Croatian & hr & ChatGPT & 502 & 497.692 \\
Macedonian & mk & ChatGPT & 509 & 499.446 \\
Polish & pl & DeepL & 511 & 487.150 \\
Slovak & sk & DeepL & 511 & 478.122 \\
Slovenian & sl & DeepL & 511 & 475.949 \\
Serbian & sr & ChatGPT & 509 & 496.172  \\
Serbian Cyrillic & sr-cyrl & ChatGPT & 511 & 499.691 \\
Ukrainian & uk & DeepL & 511 & 499.376 \\
Catalan & ca & ChatGPT & 511 & 488.937 \\
Spanish & es & DeepL & 511 & 499.260 \\
French & fr & DeepL & 511 & 499.642 \\
Galician & gl & ChatGPT & 511 & 493.112 \\
Italian & it & DeepL & 511 & 478.998 \\
Portuguese & pt & ChatGPT & 509 & 486.995 \\
Romanian & ro & DeepL & 511 & 499.733  \\
Danish & da & DeepL & 511 & 459.948 \\
German & de & DeepL & 511 & 498.699 \\
Icelandic & is & ChatGPT & 508 & 495.902 \\
Dutch & nl & DeepL & 511 & 495.574 \\
Norwegian (Bokmål) & nb & DeepL & 511 & 493.847 \\
Norwegian (Nynorsk) & nn & ChatGPT & 505 & 304.239 \\
Swedish & sv & DeepL & 511 & 491.974 \\
Lithuanian & lt & DeepL & 511 & 488.415 \\
Latvian & lv & DeepL & 511 & 438.257 \\
Greek & el & DeepL & 511 & 499.270 \\
Irish & ga & ChatGPT & 505 & 390.309 \\
Estonian & et & DeepL & 511 & 458.828 \\
Finnish & fi & DeepL & 511 & 490.227 \\
Hungarian & hu & DeepL & 511 & 496.488 \\
Basque & eu & ChatGPT & 508 & 486.467 \\
Maltese & mt & ChatGPT & 510 & 327.441 \\
Turkish & tr & DeepL & 511 & 495.888 \\
Albanian & sq & ChatGPT & 510 & 499.536 \\
Armenian & hy & ChatGPT & 508 & 498.795 \\
\bottomrule
\end{tabular}
\caption{Number of samples for each language contained in the test set and the regressor training set, including their language codes.}
\label{tab:datasets}
\end{table}

\clearpage
\section{Lightweight Annotators}\label{app:annotators}
\subsection{Experimental Setup and Parameter Choice}\label{app:annotator-setup}

To reduce computational overhead and accelerate development, we precomputed and cached all document embeddings prior to training. Since the embedding models remain frozen throughout training and account for over 99\% of the total parameter count, this approach significantly reduces iteration time. 

The regression head is implemented as a lightweight neural network: a single-layer multilayer perceptron (MLP) with ReLU activation and a final linear output layer producing a scalar prediction score. We performed a hyperparameter sweep over the hidden dimension of the MLP, exploring values from 10 to 10k. Based on this search, we selected a hidden size of 1k as a robust default. Depending on the input embedding dimension, the regression head comprises approximately 770k to 1.03M trainable parameters. 

We trained the regression heads using the AdamW optimizer with a cosine annealing learning rate schedule, which consistently outperformed constant and linearly decaying alternatives in our experiments. The initial learning rate was set to $5\times10^{-4}$, based on a sweep over values from $10^{-2}$ to $10^{-6}$. We also tested batch sizes from 16 to 4096 (in powers of two) and found a batch size of 1024 to offer the best balance between convergence speed and computational efficiency.

We trained annotators for up to 20 epochs. To monitor generalization performance, 10\% of the training data is held out for validation. We applied early stopping if the validation Spearman rank correlation fails to improve by at least $10^{-3}$ over five consecutive epochs.

\subsection{Backbone Selection}
\label{app:annotators_backbone}

We conducted an ablation study comparing three multilingual embedding models as potential backbones for our lightweight JQL annotators: \texttt{gte-multilingual-base}~\cite{zhang2024mgte}, \texttt{jina-embeddings-v3}~\cite{sturua2024jinaembeddingsv3multilingualembeddingstask}, and \texttt{snowflake-arctic-embed-m-v2.0}~\cite{yu2024arcticembed20multilingualretrieval}.

We trained a total of 18 regression heads, covering all combinations of the three embedding models and three annotation models used to generate the ground truth scores. Each combination is trained twice: once on a randomly sampled training set, and once on a class-balanced variant to mitigate the skewed distribution of education scores. Training data is sampled uniformly across all 35 languages 
The training setup—including hyperparameters and early stopping criteria—follows the procedure described in the previous section.


\begin{table}[t!]
\centering

\begin{tabular}{lccc}
\toprule
embedder & gte-multilingual-base & jina-embeddings-v3 & snowflake-arctic-embed-m-v2 \\
annotator + balancing &  &  &  \\
\midrule
Gemma-3-27B-it bal. & 0.697 ± 0.013 & \textbf{0.722} ± 0.018 & 0.720 ± 0.021 \\
Gemma-3-27B-it & 0.708 ± 0.014 & 0.734 ± 0.020 & \textbf{0.737} ± 0.028 \\
Llama-3.3-70B-it bal. & 0.693 ± 0.012 & 0.712 ± 0.010 & \textbf{0.716} ± 0.014 \\
Llama-3.3-70B-it & 0.695 ± 0.011 & 0.716 ± 0.009 & \textbf{0.724} ± 0.016 \\
Mistral-3.1-24B-it bal. & 0.707 ± 0.011 & 0.735 ± 0.011 & \textbf{\underline{0.744}} ± 0.016 \\
Mistral-3.1-24B-it & 0.687 ± 0.011 & 0.722 ± 0.017 & \textbf{0.736} ± 0.024 \\
\bottomrule
\end{tabular}

\caption{Mean and standard deviation of the Spearman correlation on all 35 testing languages. Each cell corresponds to a training setup combining an annotating model (with either raw or class-balanced annotations) and an embedding model. The best result per row is highlighted in bold. Overall best result underlined.}
\label{tab:backbone-selection}
\end{table}

Results are presented in Tab.~\ref{tab:backbone-selection}. The Snowflake embedding model consistently outperforms the other backbones across annotators and training set variants. Its best configuration—combined with the Mistral-3.1 annotation model and class-balanced training—yields the highest overall correlation (\textit{0.744} ± 0.016).

\subsection{End-to-End Training: Embedder and Regression Head}\label{app:annotators_full_training}
While the regression head alone already yields strong performance when trained on frozen embeddings, we further investigate whether end-to-end training of the full model — including both the embedding model and the regression head — can lead to improved results. To this end, we integrate the embedding model into the training loop.

This end-to-end setup comes with substantially increased memory and computational requirements. First, the embedding model accounts for over 99\% of the total parameter count. Second, the model input now consists of full-text documents instead of precomputed embeddings, resulting in significantly larger input data. These factors necessitate a reduction in batch size, which, in combination with the increased parameter count, further increases overall training time.

To conduct the end‑to‑end experiment, we adopted the learning‑rate schedule and effective batch size (via gradient accumulation) recommended in the Snowflake technical report \cite{yu2024arcticembed20multilingualretrieval}. With these settings, a single epoch on an NVIDIA A100‑SXM4‑80GB GPU takes multiple hours, whereas updating only the regression head completes an epoch in about a minute. This stark contrast quantifies the computational advantage of training only the regression head while keeping the embedding model frozen.

Due to these substantially higher runtime and memory demands, we restricted end-to-end training to the best-performing combination of Mistral annotations and Snowflake embeddings. Additionally, we observed that the model could only be trained reliably using float32 precision, as attempts with brainfloat16 led to numerical instability. This further increased the memory footprint compared to our default setup.

\begin{figure}[t!]
    \centering
    \includegraphics[width=0.9\linewidth]{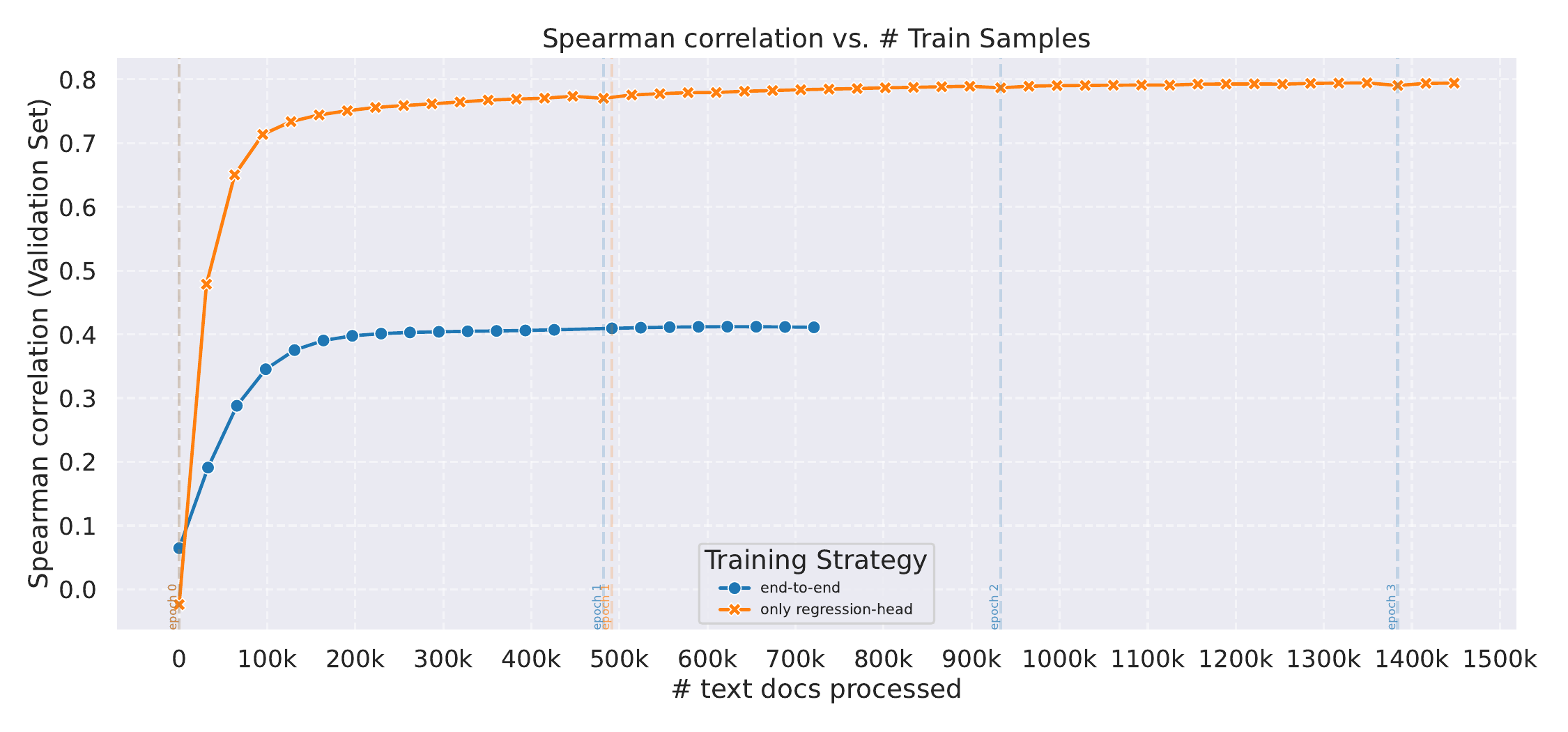}
    \caption{Validation performance (Spearman correlation) as a function of the number of processed training samples, comparing two training strategies. The end-to-end model (blue) jointly trains both the embedding backbone and the regression head, while the regression-head model (orange) fine-tunes only the regression layer on top of a frozen embedder. Performance is evaluated on a held-out validation set, and both models are trained with early stopping. Epoch boundaries are marked with dashed lines. While both models show rapid initial gains, especially during the first 100k samples, the full end-to-end model converges to a significantly lower final correlation, suggesting limited benefit from updating the embedding backbone under the given supervision signal.}
    \label{fig:regression_head_vs_end_to_end_training_convergence}
\end{figure}

Figure~\ref{fig:regression_head_vs_end_to_end_training_convergence} illustrates the training progress of both setups: the end-to-end strategy, where the embedding model is fine-tuned alongside the regression head, and the regression-head-only setup, which keeps the embedding model fixed. The figure plots the Spearman correlation on the validation set against the number of processed training samples.

While both models quickly begin to converge, the performance plateau of the end-to-end model is substantially lower than that of the regression-head-only variant. Despite the additional degrees of freedom introduced by updating the full model. This suggests that fine-tuning the embedding model does not offer any additional benefit in our setup and may even hinder performance—likely due to overfitting or insufficient optimization stability under the increased complexity.

\subsection{Training Data Amount}\label{app:annotators_data}
We conduct an ablation study to determine the minimum amount of training data required for our lightweight JQL annotators. To this end, we perform multiple training runs using varying amounts of data, randomly sampled from all 35 languages. The remainder of the experimental setup, including all hyperparameters, remains unchanged and is as described in \ref{app:annotators}. 

\begin{figure}[t!]
    \centering
    \small
    \includegraphics[width=1.1\linewidth]{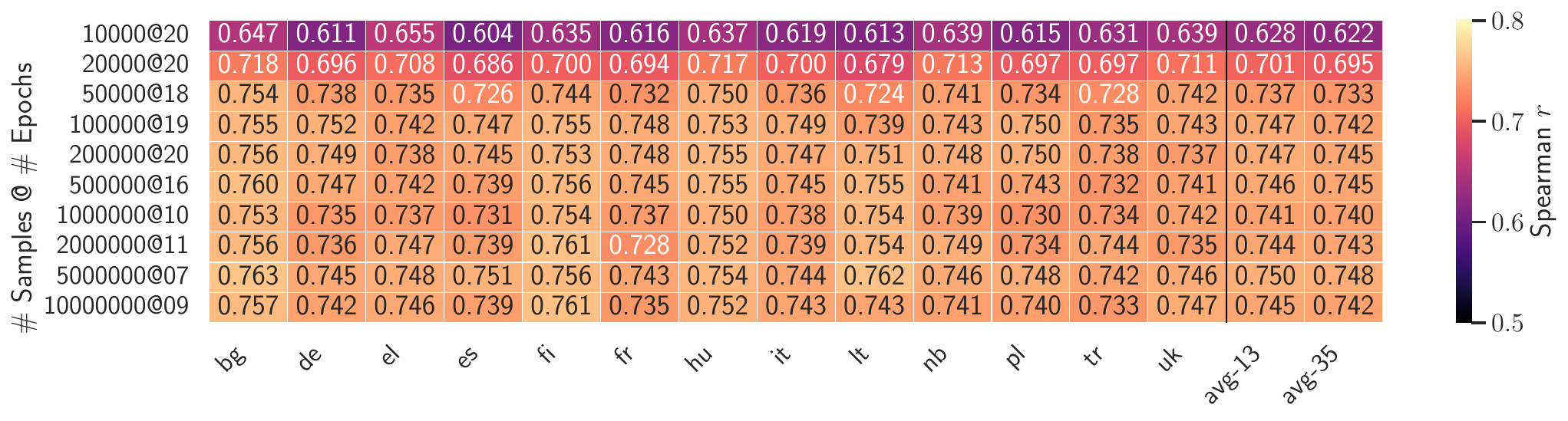}
    \caption{Ten training runs (one per row), utilizing between 10k and 10M training samples (text documents). The number of samples and corresponding training epochs are shown on the y-axis. Training is capped at 20 epochs, with early stopping based on Spearman correlation monitored on a held-out validation set. Each resulting model is evaluated in terms of Spearman correlation across all 35 test languages.}
\label{fig:training_data_evaluation}
\end{figure}

As shown in Figure~\ref{fig:training_data_evaluation}, using fewer than 50k training samples results in noticeably lower Spearman correlations. Performance continues to improve modestly up to approximately 500k samples. Beyond this point, adding more data does not yield significant gains, suggesting that training progress begins to converge. As expected, the number of training epochs required until early stopping decreases with larger training volumes.

One advantage of using smaller training set sizes is improved class balance. Since our dataset exhibits a highly imbalanced distribution of education scores—with high and very high scores being strongly underrepresented—we do not sample randomly but instead enforce approximate class balance during data selection. Achieving this balance becomes increasingly difficult as the total number of training samples increases.

Based on these considerations, we select a training set size of 500k samples.

\subsection{Detailed results.}\label{app:annotators_results}


\begin{figure}[ht]
    \centering
    \begin{subfigure}[t]{0.989\textwidth}
        \includegraphics[width=\textwidth]{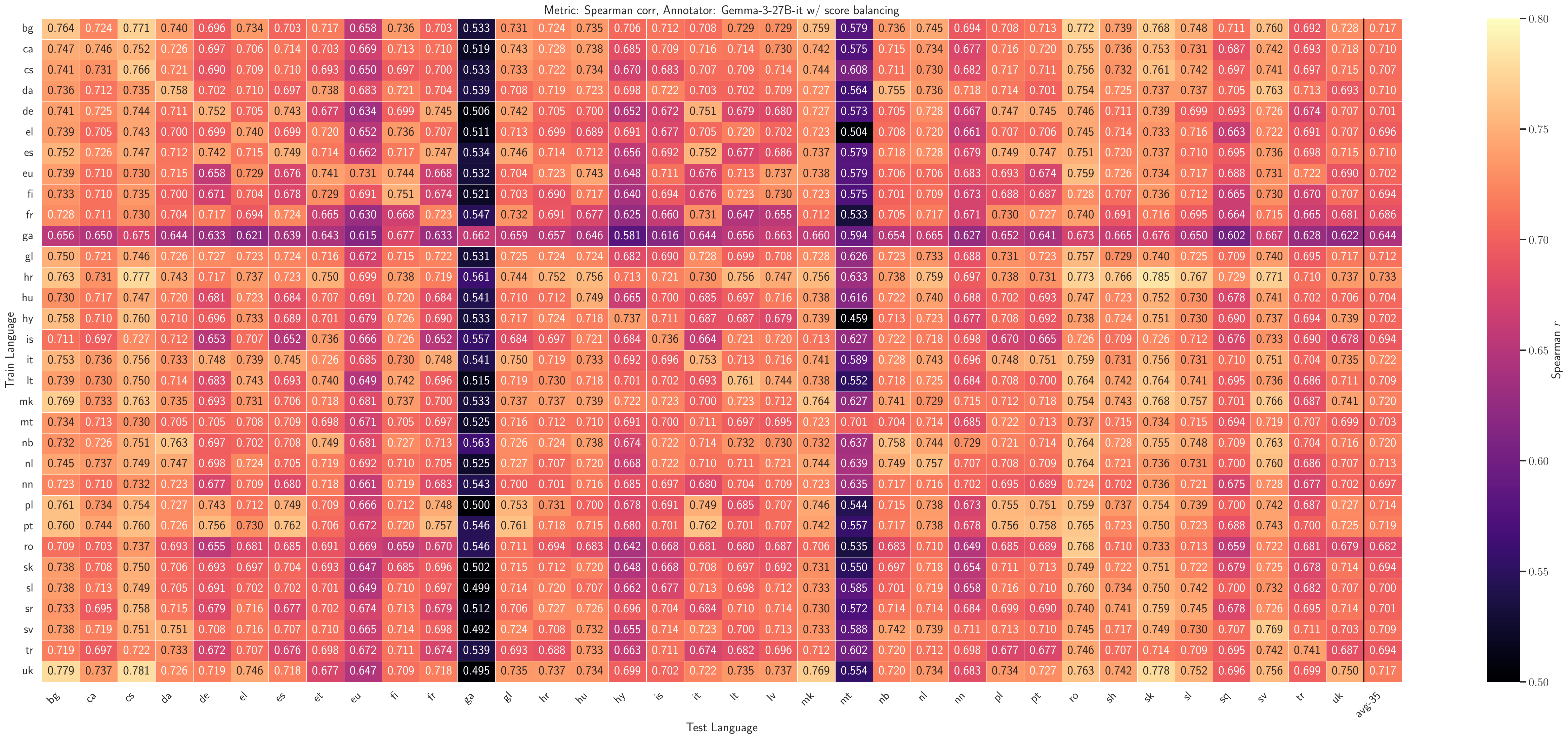}
        \label{fig:trained_individually_tested_individually_gemma_snowflake}
    \end{subfigure}
    
    \begin{subfigure}[t]{0.989\textwidth}
        \includegraphics[width=\textwidth]{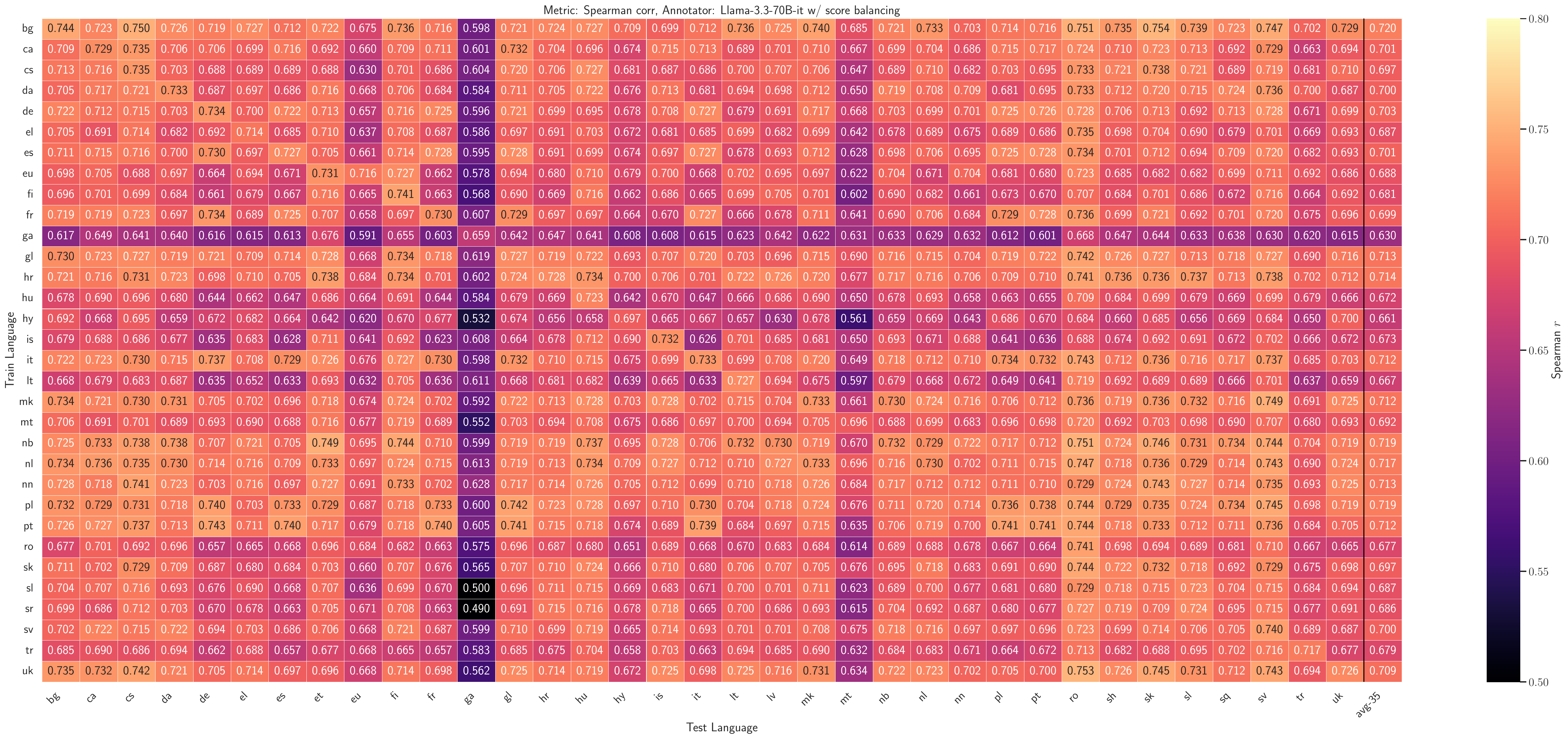}
        \label{fig:trained_individually_tested_individually_llama_snowflake}
     \end{subfigure}

    \begin{subfigure}[t]{0.989\textwidth}
        \includegraphics[width=\textwidth]{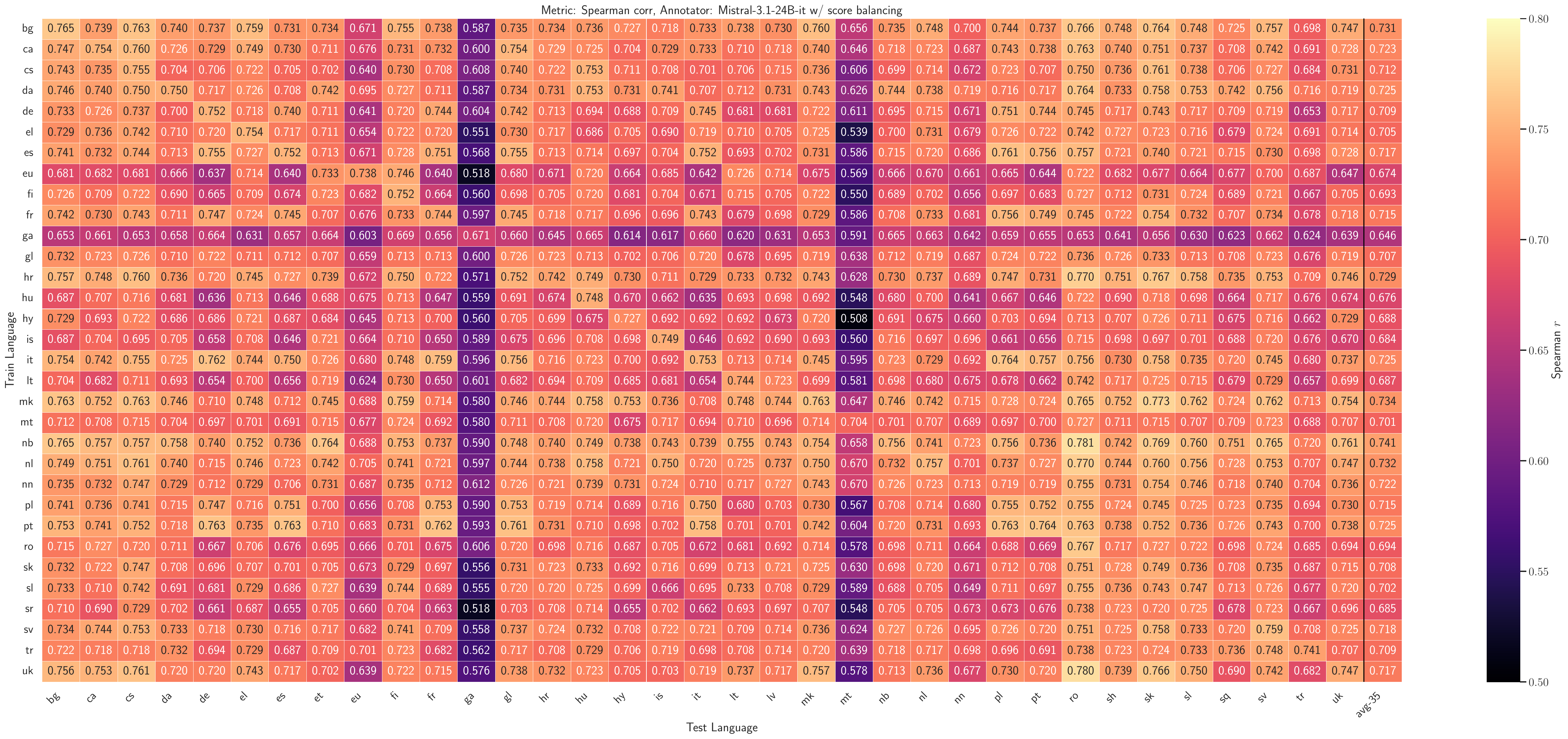}
        \label{fig:trained_individually_tested_individually_mistral_snowflake}
     \end{subfigure}
    \caption{Full cross-lingual transfer; One plot per Annotation model (balanced); training/evaluation setup is otherwise identical to the best performing setup. Rows represent the only training language of a regression head, while columns indicate the testing language. Each cell reports the Spearman correlation between predicted and human-annotated scores.}
    \label{fig:trained_individually_tested_individually_snowflake}
\end{figure}

We here provide additional details complementing the main results. Specifically, Fig~\ref{fig:trained_individually_tested_individually_snowflake} shows the full matrix of cross-lingual transfer performance across all languages considered in our study. Each row corresponds to a regression head trained solely on one specific language, while each column represents the test language.

The values in each cell indicate the Spearman correlation between the model’s predictions and human-annotated scores. This exhaustive view highlights the generalization capability of the model across language boundaries.

\clearpage
\section{Assessing Training Data Quality}\label{app:downstream_analysis}
In this Section, we provide further details and ablations on our lightweight annotators discussed in Section~\ref{sec:downtream_analysis}.

\subsection{Experimental Setup}\label{app:downstream_analysis_setup}
We here provide further details on experimental setup and hyperparameter for our LLM training ablations.

\paragraph{Architecture.}
\begin{itemize}
    \item 262144 vocab size SentencePiece {tokenizer} from Gemma-3 \cite{gemmateam2025gemma3technicalreport}.
    \item Dense {Llama} architecture
    \item 2048 {hidden dimension}
    \item 24 {hidden layers}
    \item 32 {attention heads}
    \item Silu {activation} 
    \item Root Mean Square Layer Normalization ({RMSNorm}) with $\epsilon=1.0e-05$
    \item Rotary Position Embeddings ({RoPE}) with $\theta=130000$
    \item \textbf{Weight tying} for embedding and LM head is customary for small LLMs \cite{allal2025smollm2smolgoesbig}
\end{itemize}

\paragraph{Training.}
\begin{itemize}
    \item \textbf{Nanotron}\footnote{\url{https://github.com/huggingface/nanotron}} as training framework with tokenization using \textbf{Datatrove}\footnote{\url{https://github.com/huggingface/datatrove}}
    \item 2048 {sequence length}
    \item Simple {document concatenation} as Datatrove does not support advanced packing algorithms
    \item AdamW {optimizer} with $\beta_1 = 0.9$, $\beta_2 = 0.95$, $\epsilon=1.0e-8$
    \item cosine learning rate {decay}, peak $lr=1.5e-4$, decay to $lr=1.5e-5$
    \item linear {warmup} for 150 steps
    \item global {batch size} 960 with micro-batch size $3$ and gradient accumulation $5$. 
    \item 1,966,080 tokens per step
    \item Training on 64 NVIDIA A100-SXM4-80GB with full data parallelism and no tensor or pipeline parallelism
\end{itemize}

\paragraph{Data Curation.}
Our custom data curation data pipeline for annotation, filtering and tokenization builds on Datatrove. We use the transformers implementation with a batch size of 1000 documents per GPU for embedding calculation. Surprisingly, we observed no speedup when using torch compile. 

\paragraph{Benchmarks.}
In order to conduct our benchmarks, we utilize custom \textbf{Lighteval}\footnote{\url{https://github.com/huggingface/lighteval}} tasks. To provide a unified interface, we reformatted ArcX and MMMLU sources and repacked them to maintain a coherent structure. For MMMLU, we used off-the-shelf HF-datasets. 
In all our selected sources, we considered the highest-quality translations available, such as human translations from openai/mmmlu, and only resorted to automatic translations if necessary.
The mapping of the different languages to sources is provided in Tab.~\ref{tab:lang_codes_sources}.
\begin{table}[h]
    \centering
    \renewcommand{\arraystretch}{1.5}
    \begin{tabular}{l c p{4cm} p{4cm} p{4cm}}
    \toprule
    \textbf{Language} & \textbf{Code} & \textbf{ArcX Source} & \textbf{MMMLU Source} & \textbf{HellaSwag Source} \\
    \midrule
    Bulgarian & bg & openGPT-X/arcx & openGPT-X/mmlux  & openGPT-X/hellaswagX  \\
    German & de & openGPT-X/arcx & openai/MMMLU  & openGPT-X/hellaswagX  \\
    Greek & el & openGPT-X/arcx & CohereLabs/Global-MMLU  & openGPT-X/hellaswagX \\
    Spanish & es & openGPT-X/arcx & openai/MMMLU & openGPT-X/hellaswagX \\
    Finnish & fi & openGPT-X/arcx & openGPT-X/mmlux  & openGPT-X/hellaswagX \\
    French & fr & openGPT-X/arcx & openai/MMMLU  & openGPT-X/hellaswagX \\
    Hungarian & hu & openGPT-X/arcx & openGPT-X/mmlux  & openGPT-X/hellaswagX \\
    Italian & it & openGPT-X/arcx & openai/MMMLU  & openGPT-X/hellaswagX \\
    Lithuanian & lt & openGPT-X/arcx & CohereLabs/Global-MMLU  & openGPT-X/hellaswagX \\
    Norwegian & nb & alexandrainst/m\_arc  & NbAiLab/nb-global-mmlu  & alexandrainst/m\_hellaswag  \\
    Polish & pl & openGPT-X/arcx & CohereLabs/Global-MMLU  & openGPT-X/hellaswagX \\
    Turkish & tr & malhajar/arc-tr  & CohereLabs/Global-MMLU  & malhajar/hellaswag-tr \\
    Ukrainian & uk & alexandrainst/m\_arc & CohereLabs/Global-MMLU  & alexandrainst/m\_hellaswag  \\
    \bottomrule
    \end{tabular}
    \caption{Mapping of language to corresponding ArcX, MMMLU, and HellaSwag sources.}
    \label{tab:lang_codes_sources}
\end{table}

\subsection{Details on Annotation Distribution}\label{app:downstream_analysis_distribution}
Subsequently, we provide a more detailed insights beyond the annotation distribution analyzed in Sec.~\ref{sec:annotation_distribution}.

\begin{figure}[b]
    \centering
    \includegraphics[width=0.6\linewidth]{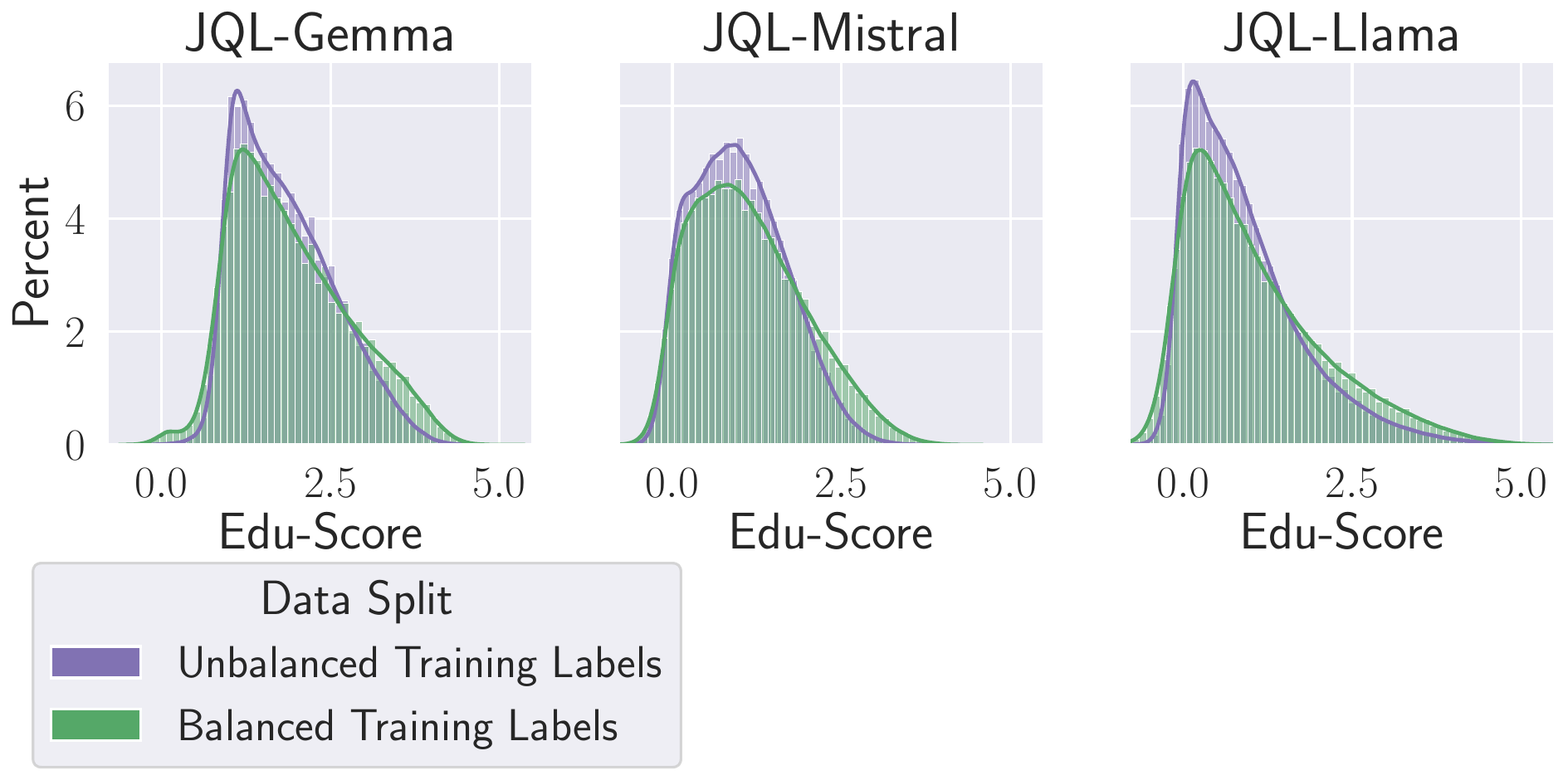}
    \caption{Distribution of different lightweight annotation heads on CC release 2024-14 over 13 languages. Training heads on balanced labels produces slightly smoother distributions.}
    \label{fig:annotation_distribution_label_balance}
\end{figure}

In Fig.~\ref{fig:annotation_distribution_label_balance}, we visualize the downstream impact of balancing the training data of lightweight annotation heads. Training heads on balanced labels produces slightly smoother distributions, which makes dynamic thresholding less volatile.

\begin{figure}[b]
    \centering
    \includegraphics[width=0.8\linewidth]{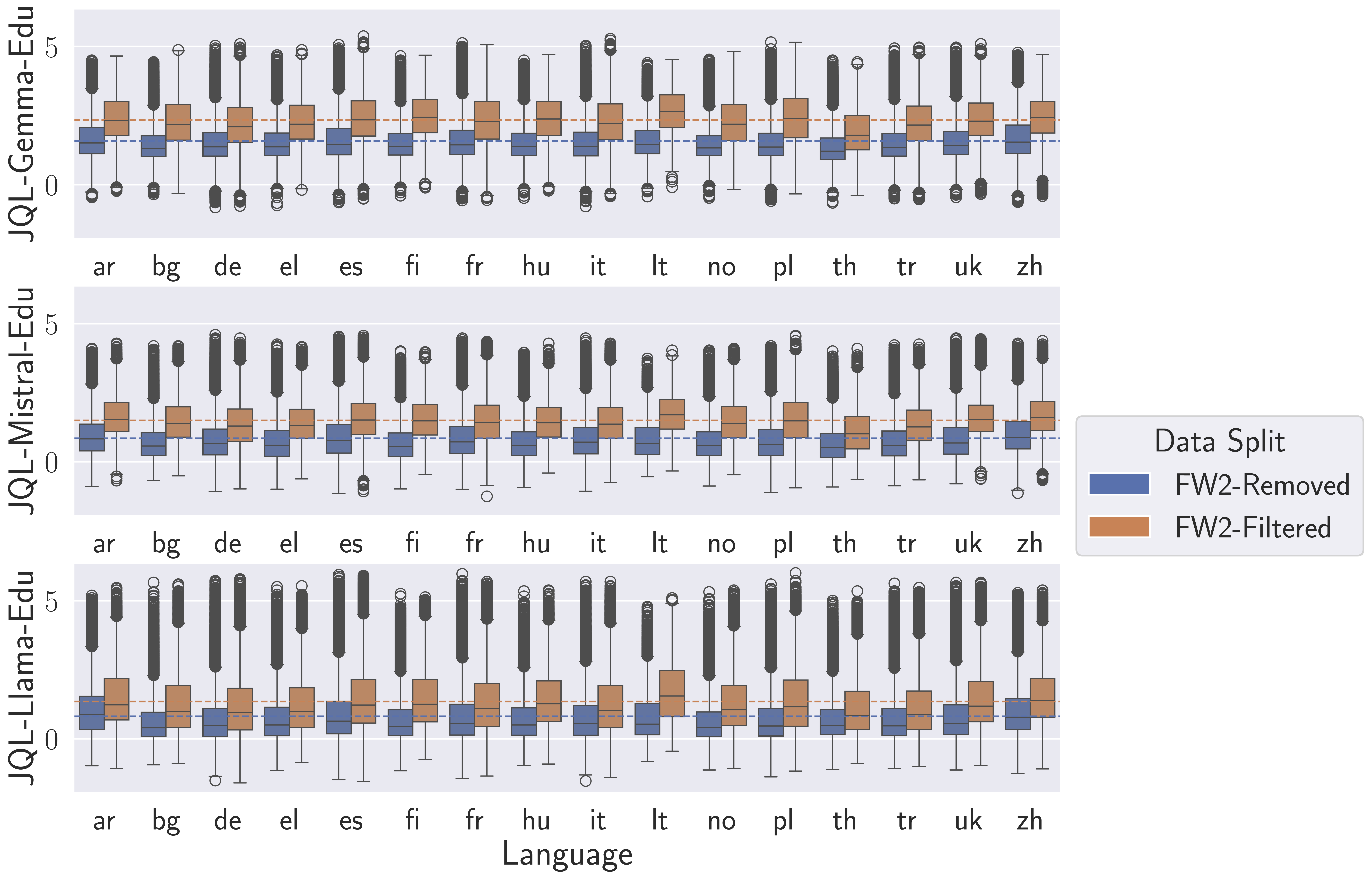}
    \caption{Distribution of edu score annotations by language. Dotted lines represent the respective mean.}
    \label{fig:annotation_distribution_languages}
\end{figure}

Additionally, we show the difference in label distributions per language in Fig,~\ref{fig:annotation_distribution_languages}. The results demonstrate that the heuristic FW-2 filters doe not uniformly produce similar document quality levels. For example, the average educational value of retained documents in Lithuanian is significantly higher than in other languages. Further, we can see a significant overlap in scores within the filtered and removed subsets. These results further highlight the difficulty of constructing heuristic filters that generalize well to different languages. Instead, approaches like JQL that use document semantics extracted from cross-lingually aligned embeddings tend to generalize better. 
\subsection{Further Results.}
\begin{figure}[t]
    \centering
    \begin{minipage}{.48\textwidth}
        \centering
        \includegraphics[width=\linewidth]{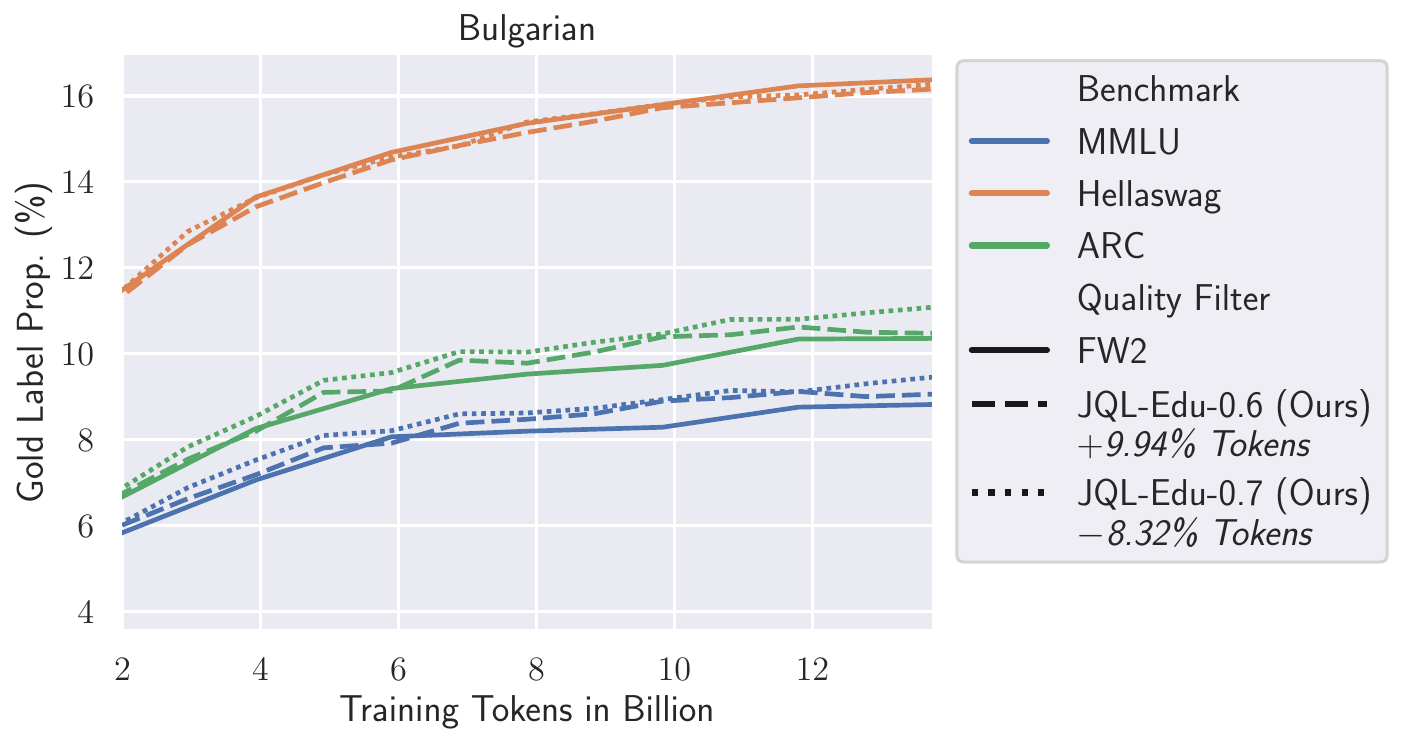}
        \caption{Dataset training performance for Bulgarian.}
        \label{fig:train_progress_bg}
    \end{minipage}%
    \hfill
    \begin{minipage}{0.48\textwidth}
        \centering
        \includegraphics[width=\linewidth, height=0.15\textheight]{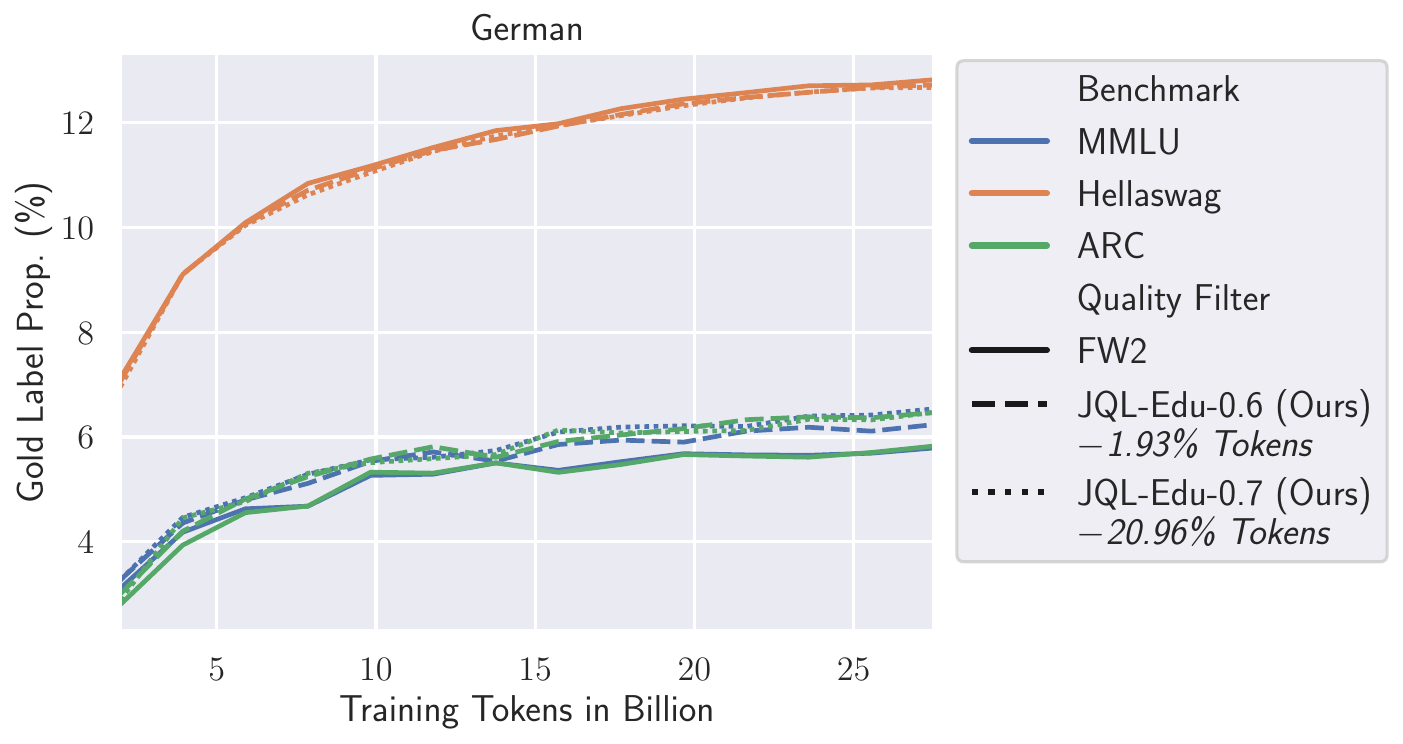}
        \caption{Dataset training performance for German.}
       
    \end{minipage}
\end{figure}

\begin{figure}[]
    \centering
    \begin{minipage}{.48\textwidth}
        \centering
        \includegraphics[width=\linewidth]{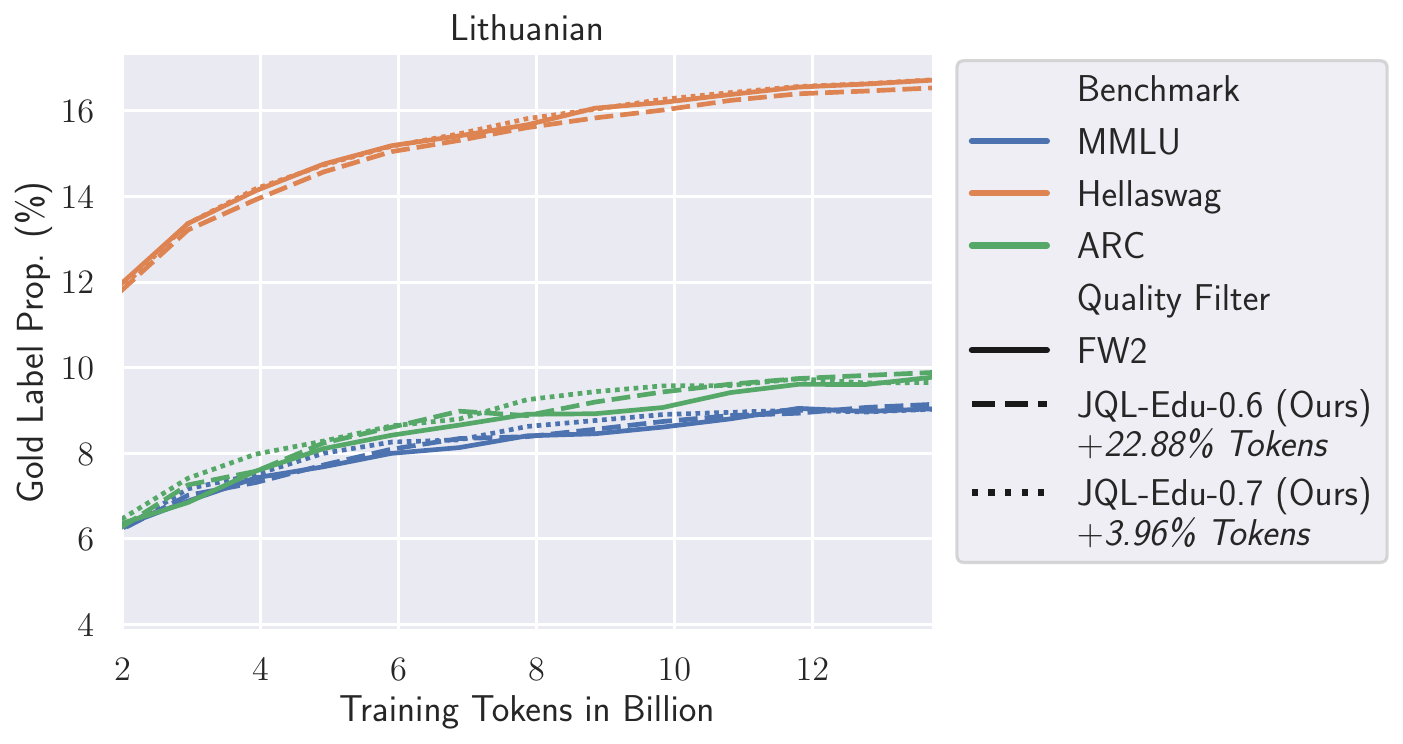}
        \caption{Dataset training performance for Lithuanian.}
        \label{fig:train_progress_lt}
    \end{minipage}%
    \hfill
    \begin{minipage}{.48\textwidth}
        \centering
        \includegraphics[width=\linewidth]{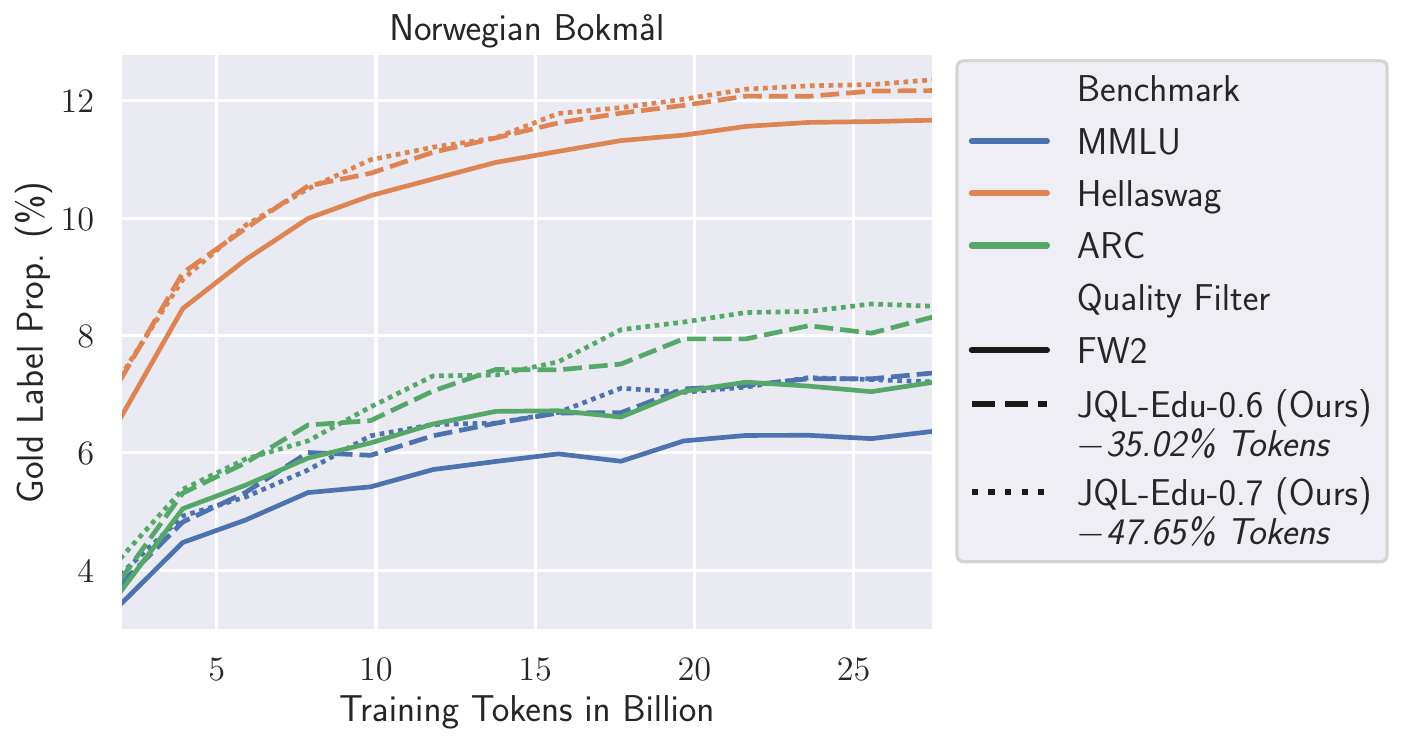}
        \caption{Dataset training performance for Norwegian (Bokmål).}
    \end{minipage}
\end{figure}

\begin{figure}[]
    \centering
    \begin{minipage}{0.48\textwidth}
        \centering
        \includegraphics[width=\linewidth, height=0.15\textheight]{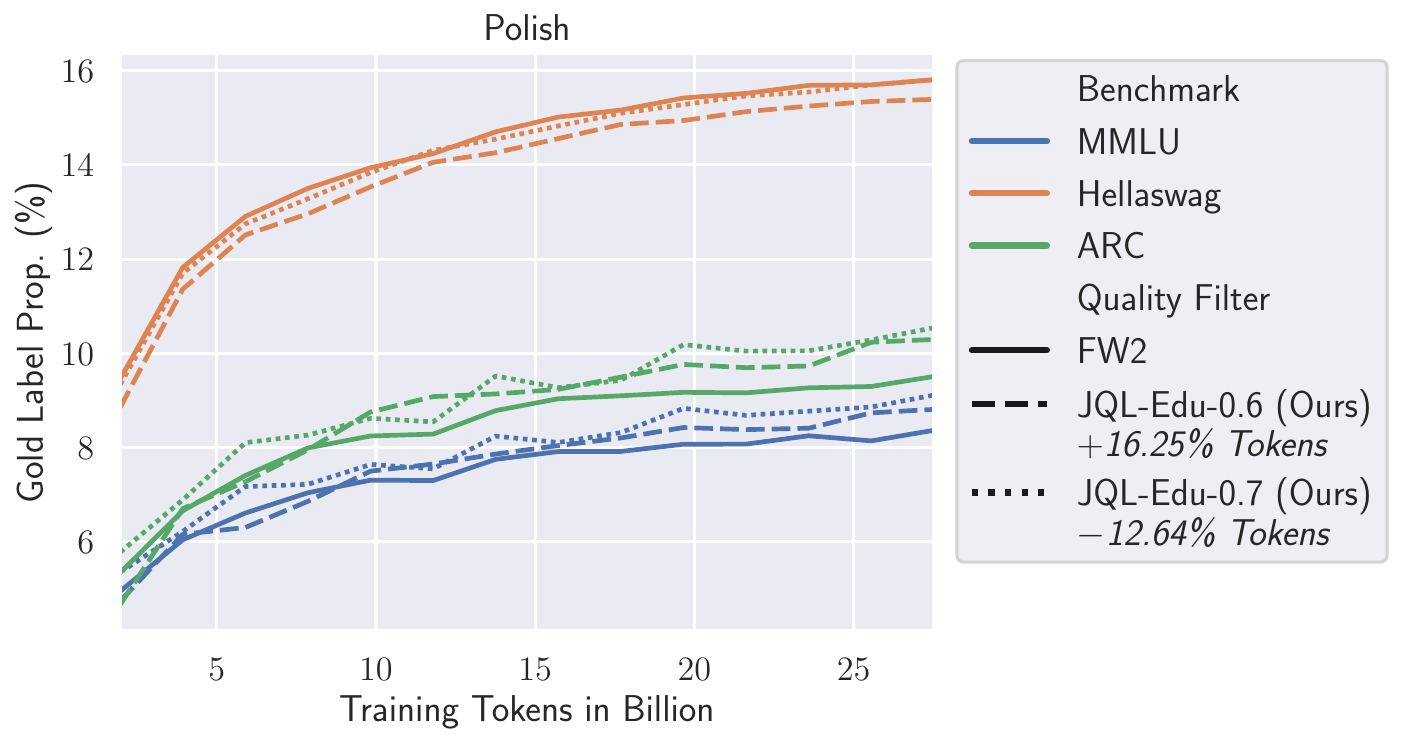}
        \caption{Dataset training performance for Polish.}
        \label{fig:train_progress_pl}
    \end{minipage}
    \hfill
    \begin{minipage}{.48\textwidth}
        \centering
        \includegraphics[width=\linewidth]{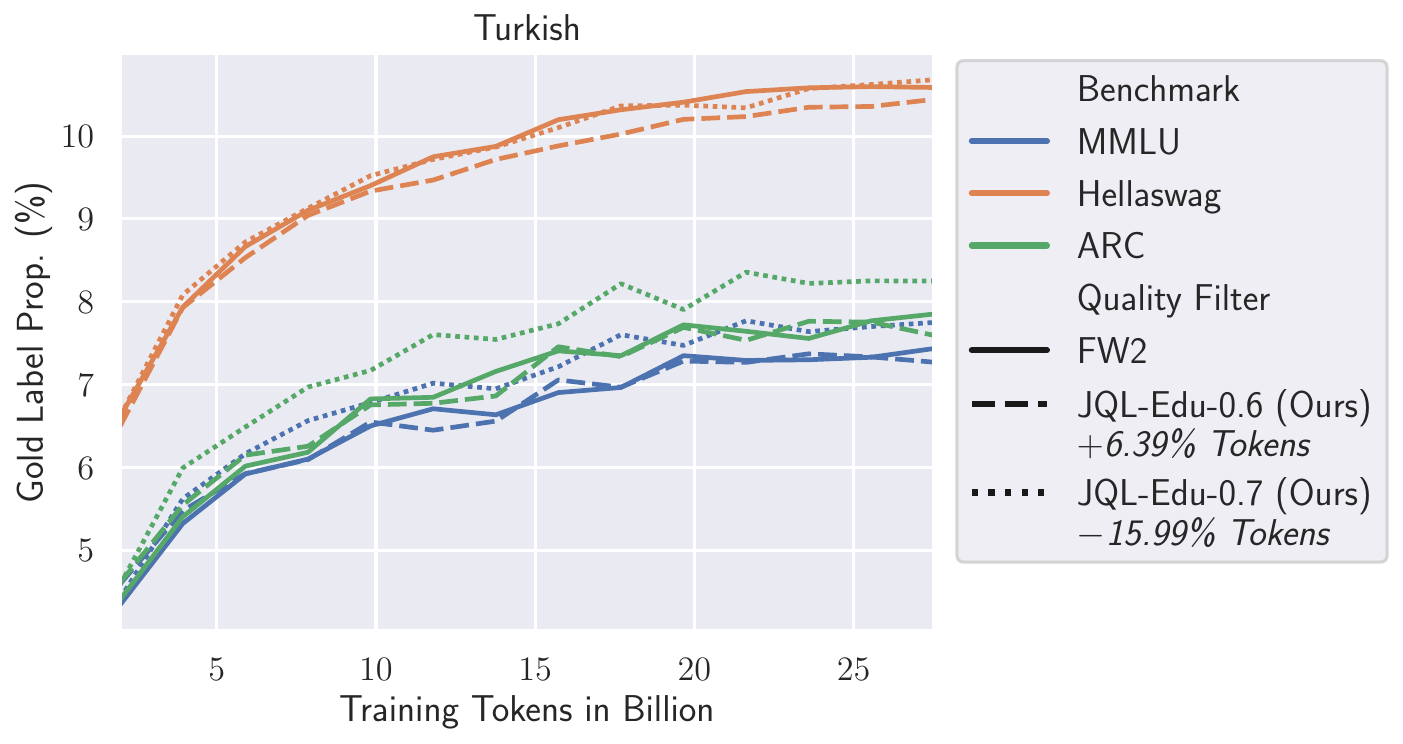}
        \caption{Dataset training performance for Turkish.}
    \end{minipage}%
\end{figure}

\begin{figure}[t]
    \centering
     \begin{minipage}{0.48\textwidth}
        \centering
        \includegraphics[width=\linewidth, height=0.15\textheight]{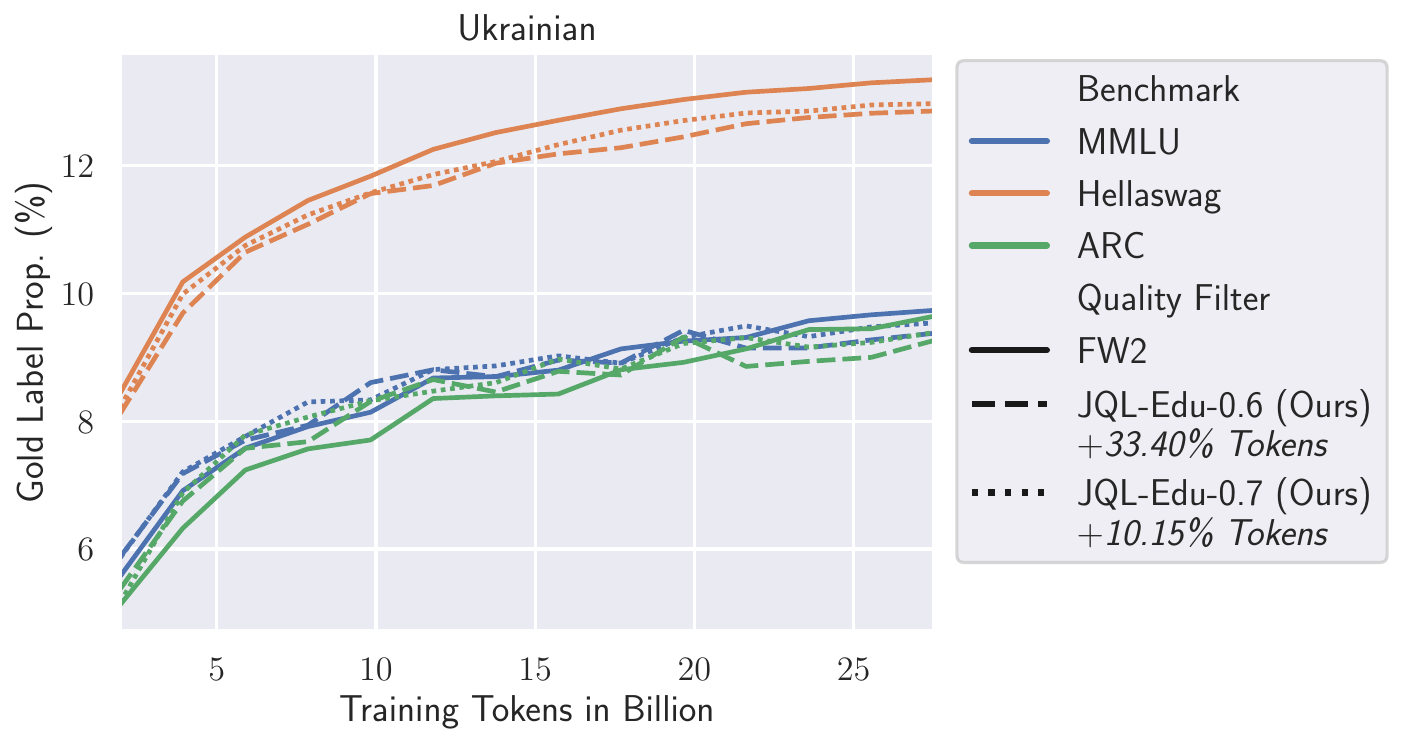}
        \caption{Dataset training performance for Ukrainian.}
       \label{fig:train_progress_uk}
    \end{minipage}
    \hfill
    \begin{minipage}{.48\textwidth}
        \centering
        \includegraphics[width=\linewidth]{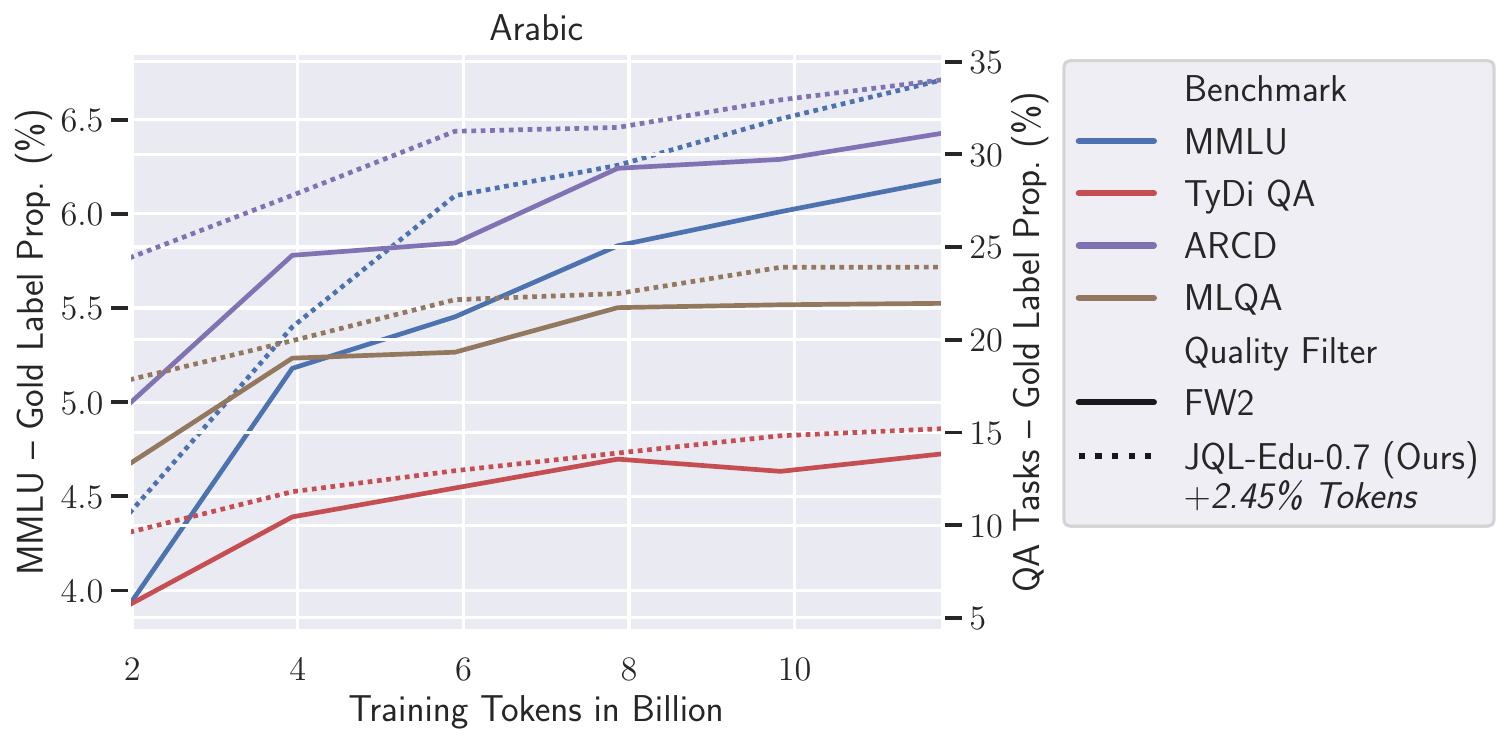}
        \caption{Dataset training performance for Arabic.}
       \label{fig:train_progress_ar}
    \end{minipage}%
    
\end{figure}

We provide more details of the results shown in the main body. Specifically, we depict the results for all languages under consideration in Fig.~\ref{fig:train_progress_bg}-Fig.~\ref{fig:train_progress_uk}.
For almost all languages, we observe significant improvements over the FW2 baseline, especially on MMLU and Hellaswag. Additionally, we see higher retention rates for many languages. For example, in Polish (see Fig.~\ref{fig:train_progress_pl}), our lightweight edu annotation model with a dynamic threshold of 0.6 outperforms FW2 while retaining 16\% more tokens.
The only two languages with no clear improvements are Lithuanian (Fig.~\ref{fig:train_progress_lt}) and Ukranian (Fig.~\ref{fig:train_progress_uk}). However, in these cases, we maintain comparable performance while retaining up to 23\% and 33\% more tokens, respectively. 


\begin{figure}[t]
    \centering
    \begin{minipage}{.48\textwidth}
        \centering
        \includegraphics[width=\linewidth]{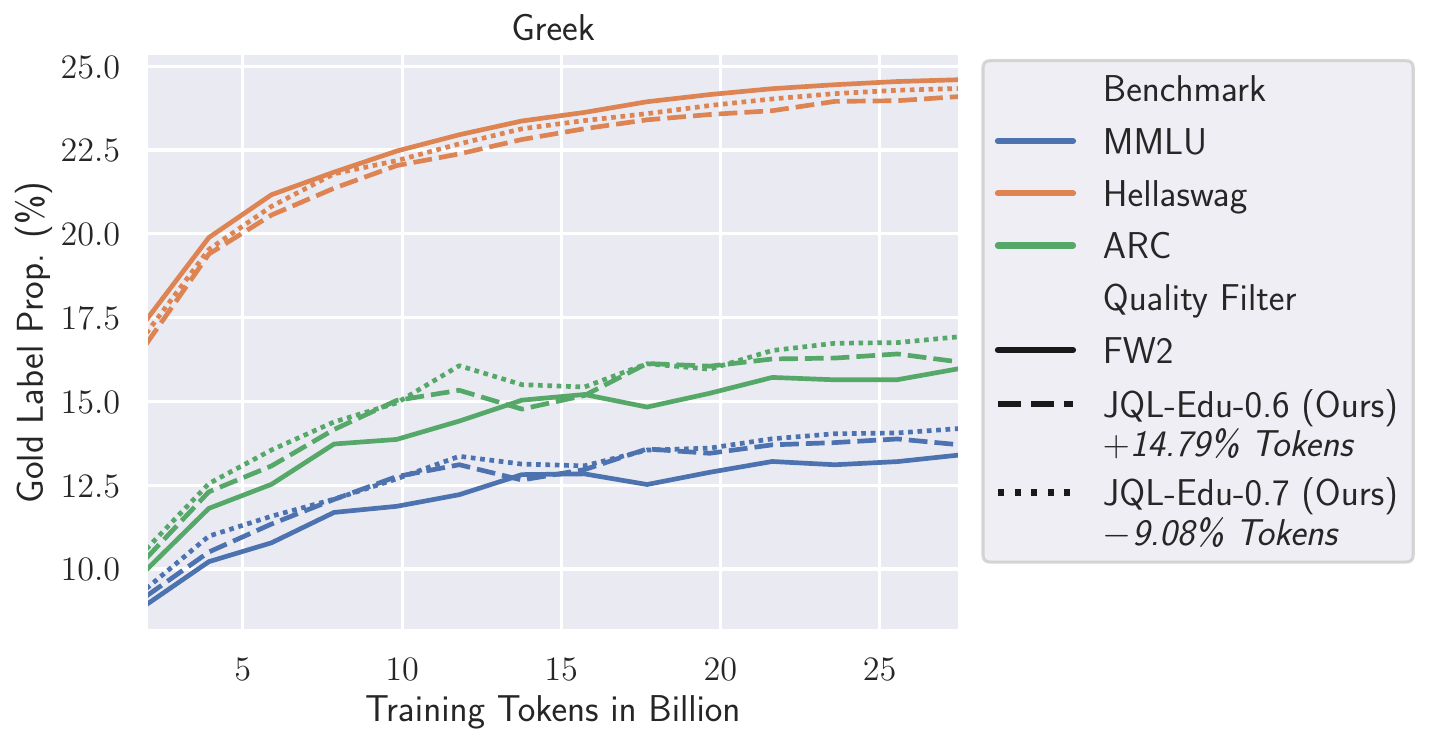}
        \caption{Dataset training performance for Greek.}
       
    \end{minipage}%
    \hfill
    \begin{minipage}{0.48\textwidth}
        \centering
        \includegraphics[width=\linewidth, height=0.15\textheight]{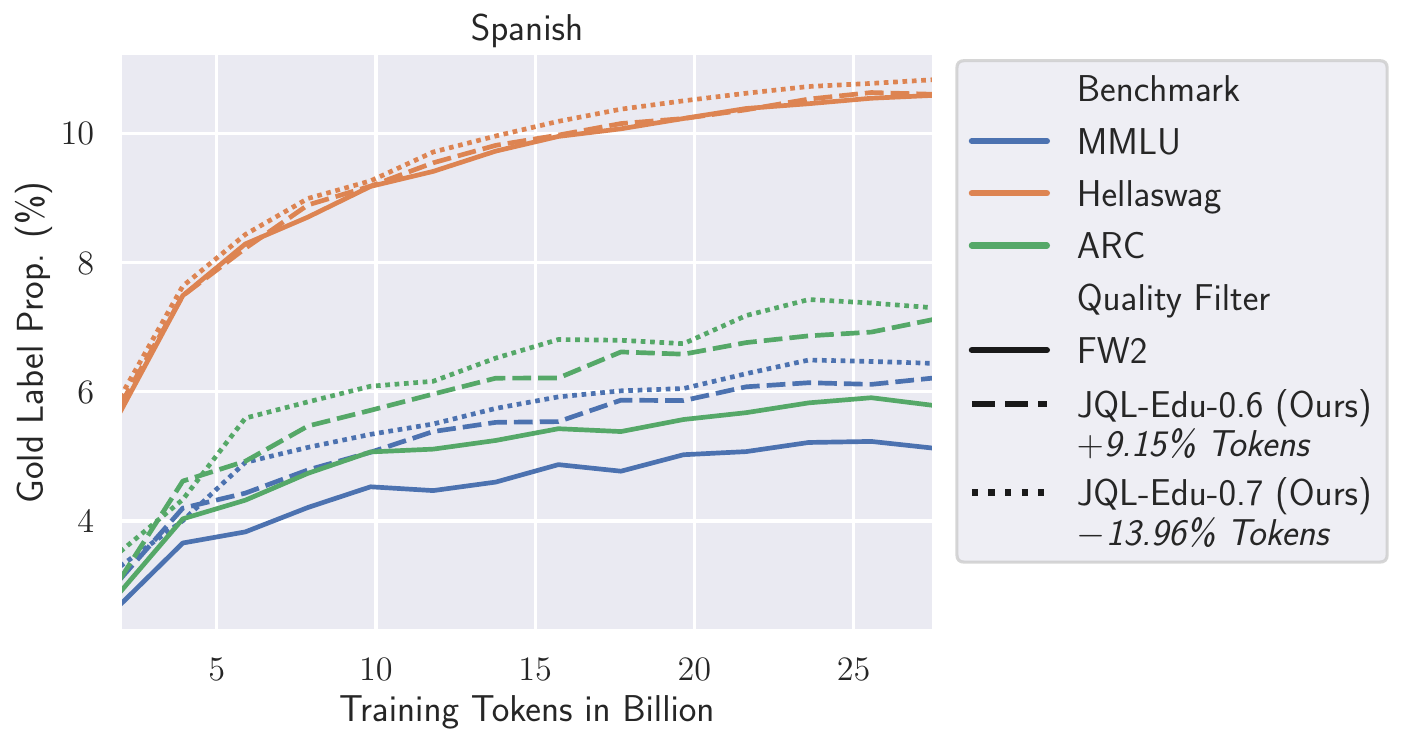}
        \caption{Dataset training performance for Spanish.}
       
    \end{minipage}
\end{figure}

\begin{figure}[]
    \centering
    \begin{minipage}{.48\textwidth}
        \centering
        \includegraphics[width=\linewidth]{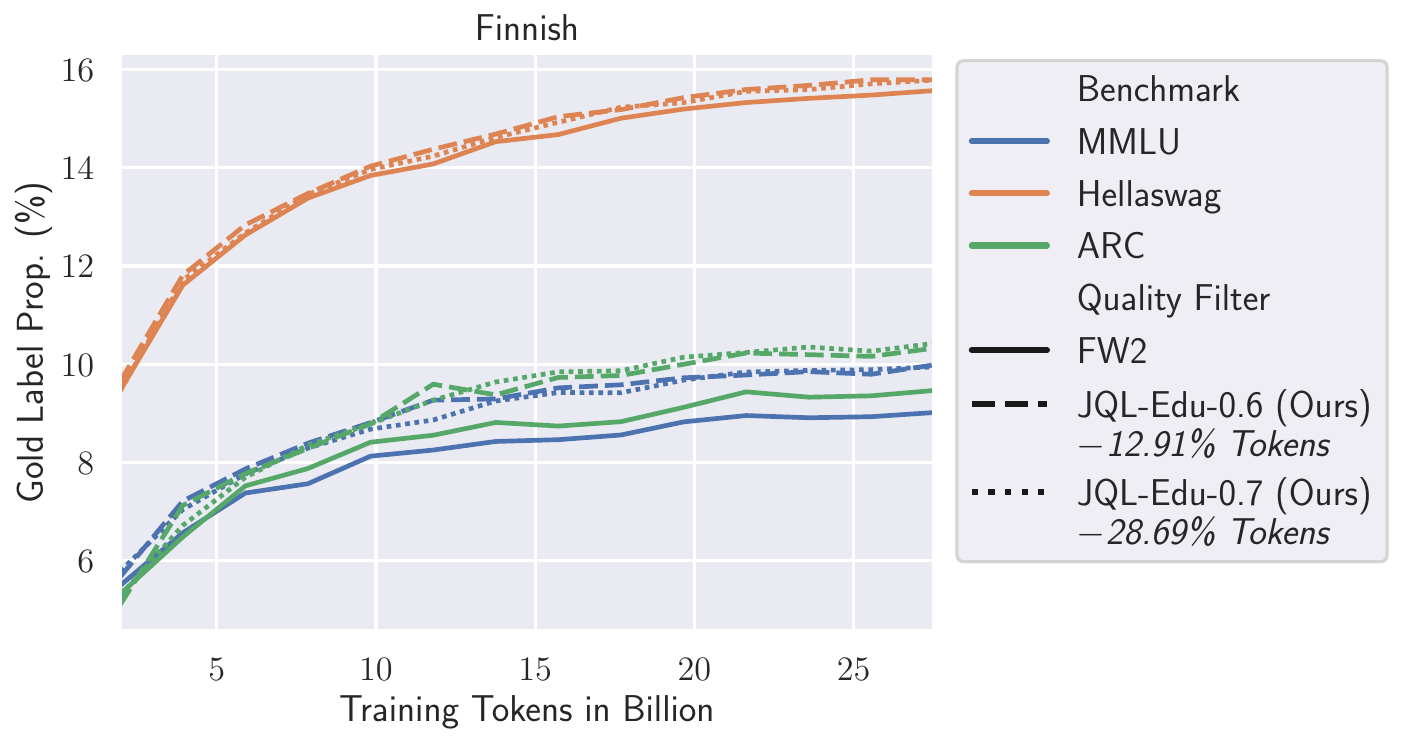}
        \caption{Dataset training performance for Finnish.}
       
    \end{minipage}%
    \hfill
    \begin{minipage}{0.48\textwidth}
        \centering
        \includegraphics[width=\linewidth, height=0.15\textheight]{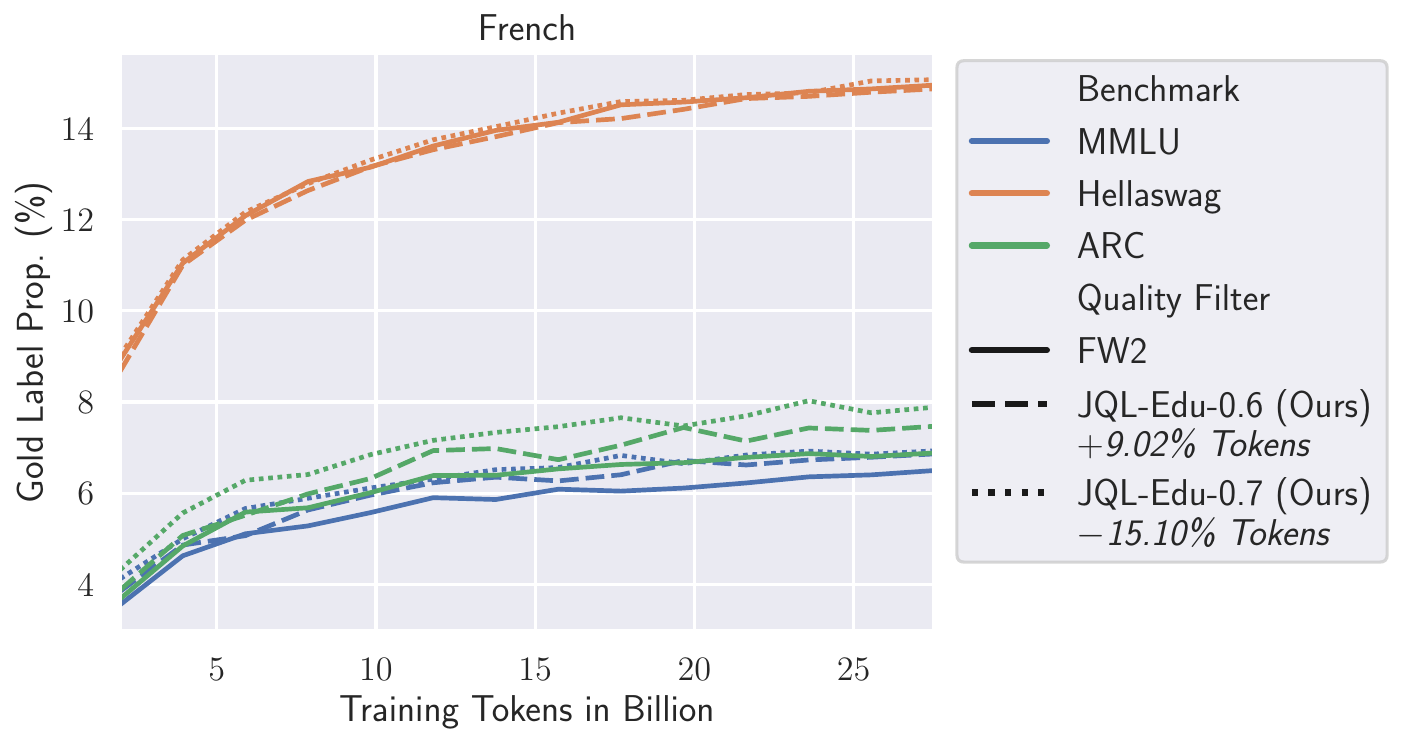}
        \caption{Dataset training performance for French.}
       
    \end{minipage}
\end{figure}

\begin{figure}[]
    \centering
    \begin{minipage}{.48\textwidth}
        \centering
        \includegraphics[width=\linewidth]{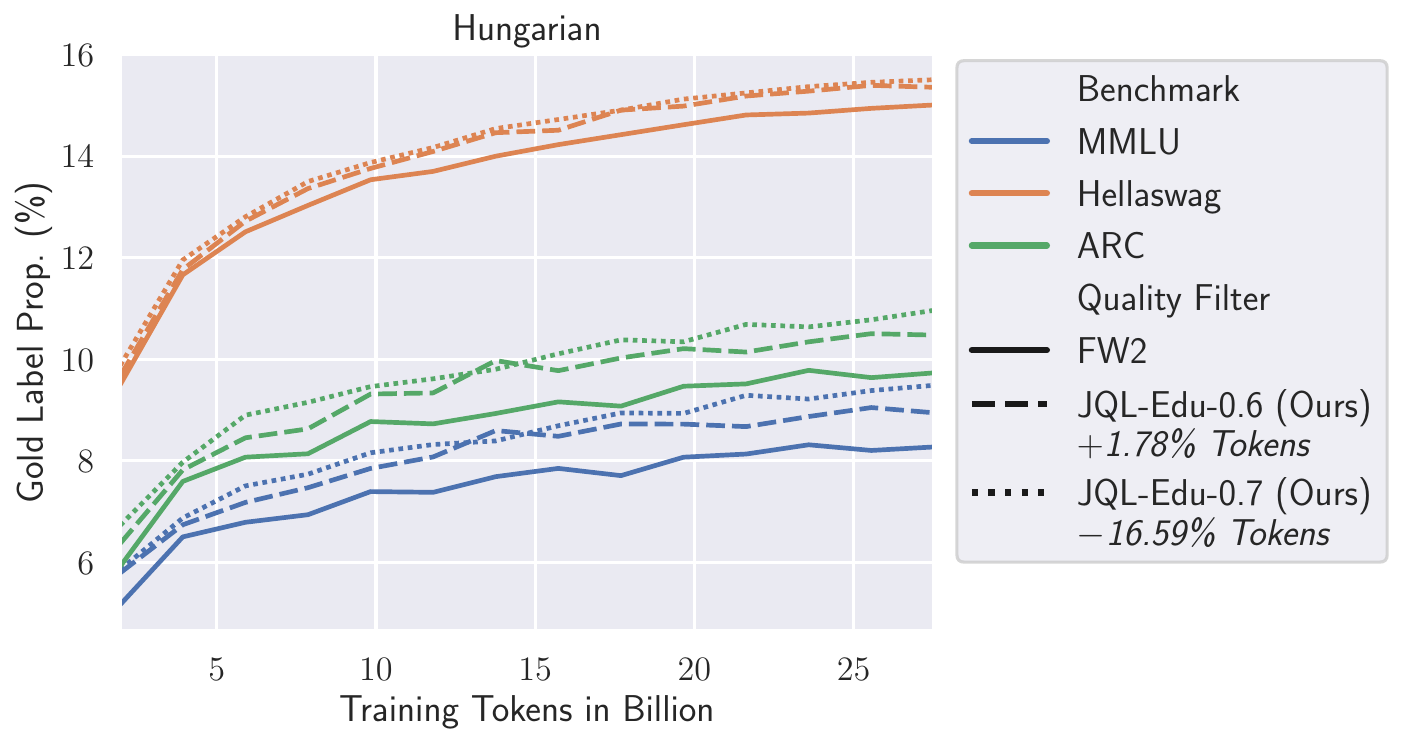}
        \caption{Dataset training performance for Hungarian.}
       
    \end{minipage}%
    \hfill
    \begin{minipage}{0.48\textwidth}
        \centering
        \includegraphics[width=\linewidth, height=0.15\textheight]{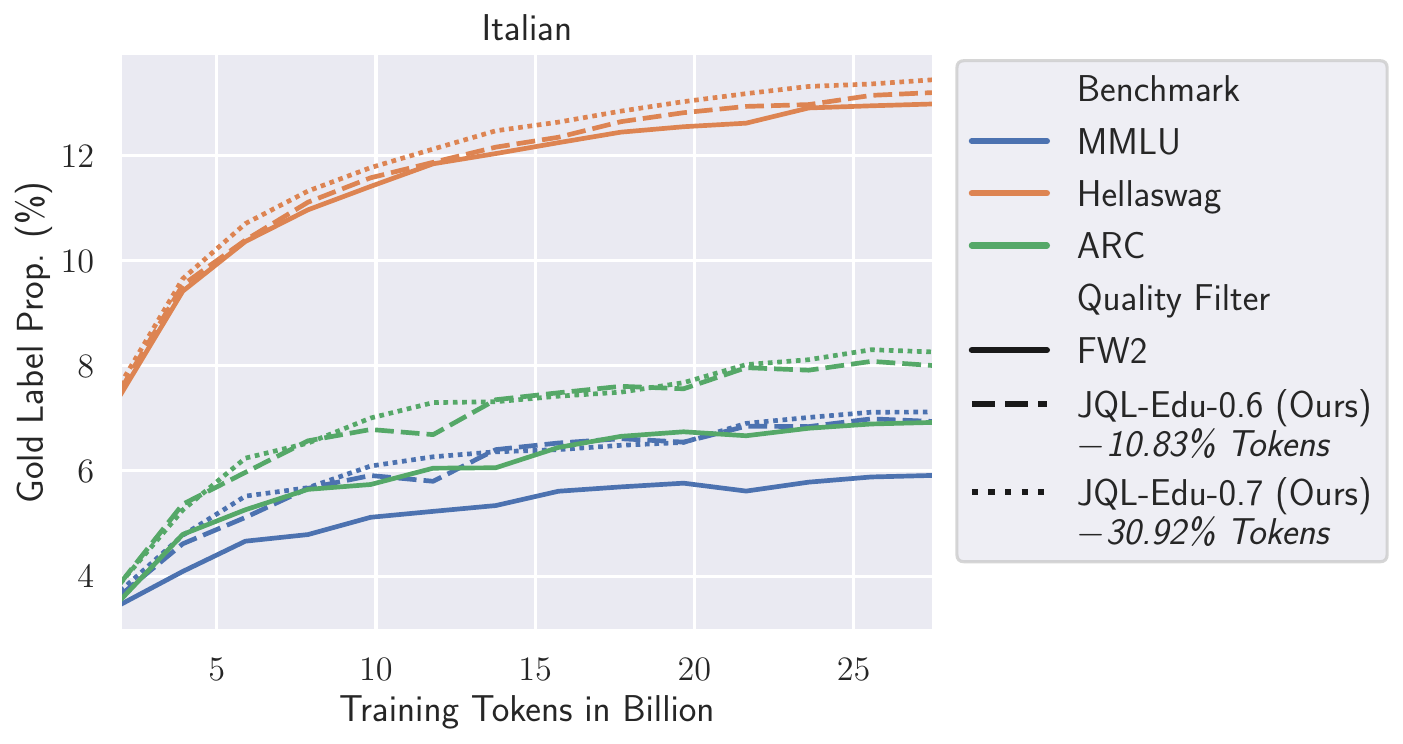}
        \caption{Dataset training performance for Italian.}
       
    \end{minipage}
\end{figure}

\clearpage

\begin{figure}[t]
    \centering
     \begin{minipage}{0.48\textwidth}
        \centering
        \includegraphics[width=\linewidth, height=0.15\textheight]{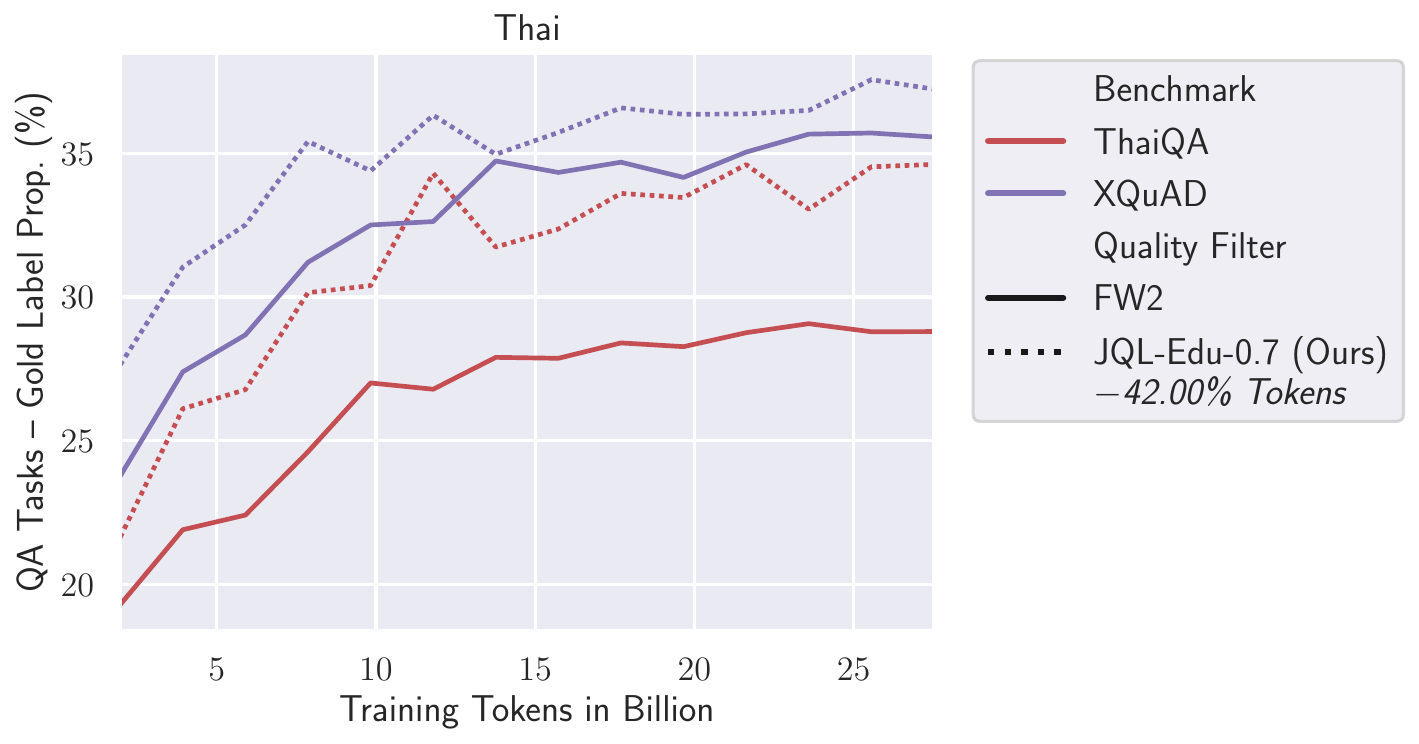}
        \caption{Dataset training performance for Thai.}
       \label{fig:train_progress_th}
    \end{minipage}
    \hfill
    \begin{minipage}{.48\textwidth}
        \centering
        \includegraphics[width=\linewidth]{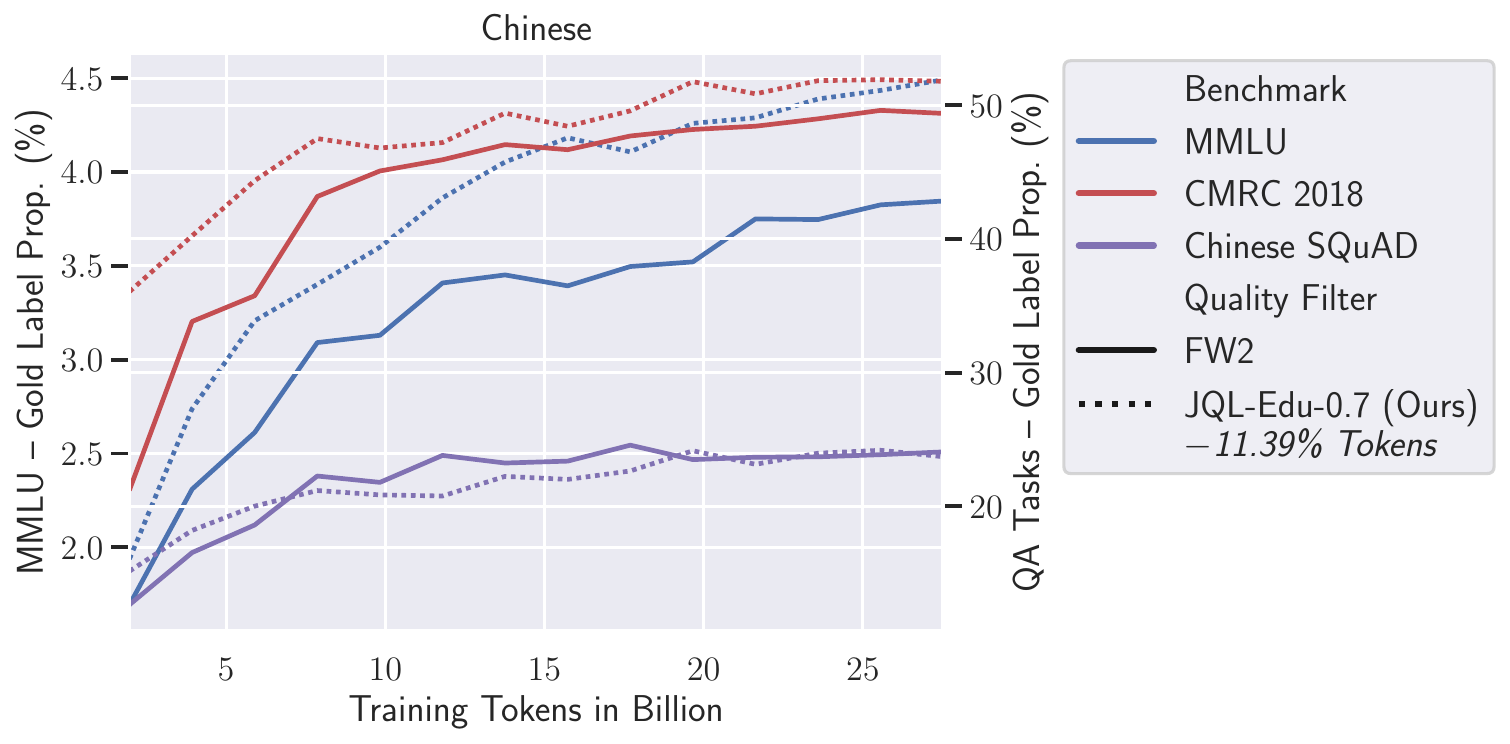}
        \caption{Dataset training performance for Chinese.}
        \label{fig:train_progress_zh}
    \end{minipage}%
    
\end{figure}
\section{Generalization to Unseen languages}\label{app:generalization}
In this Section, we provide further details and ablations on our generalization experiment on Arabic, Thai, and Chinese in Section~\ref{sec:generalization}.

\begin{figure}
    \centering
    \includegraphics[width=0.5\linewidth]{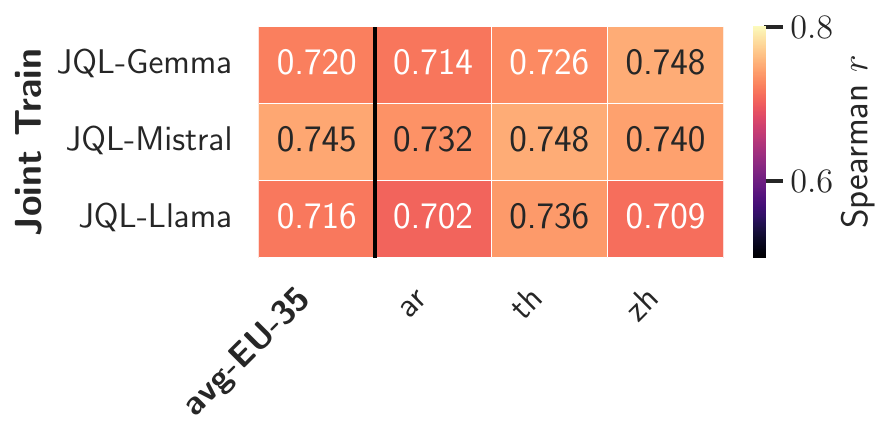}
    \caption{Strong cross-lingual performance of our lightweight JQL annotators on unseen languages (Arabic, Thai, and Chinese). Compared to the average performance of the European languages on which the annotators are trained, we observe an even better correlation with human GT for some languages. }
    \label{fig:cross_lingual_generalization}
\end{figure}

\subsection{Evaluation of Lightweight PQL-Annotator}\label{app:generalization_annotator}
We first translated our ground truth documents in the 3 new target languages. The zero-shot performance of our previously trained lightweight annotators is depicted in Fig.~\ref{fig:cross_lingual_generalization}.
For these three topologically new languages, we can see the same level of performance as for the European languages. For Thai, we even observed better performance than the European language average across all annotators. Consequently, JQL generalizes well to new languages (families). 

\subsection{Further Results}\label{app:generalization_results}
In Figs.~\ref{fig:train_progress_ar},~\ref{fig:train_progress_th} and~\ref{fig:train_progress_zh}, we compare the training curves of Arabic, Thai, and Chinese, respectively. Since we only found high-quality MMLU versions for Arabic and Chinese, we additionally evaluated the benchmarks proposed by the Fineweb team \cite{penedo2024fineweb-2}. Specifically, we extend our evaluation with the following QA benchmarks:
\begin{itemize}
    \item \textbf{XQuAD} (\texttt{google/xquad}) – 1.190 English QA pairs professionally translated into 10 languages \cite{Artetxe:etal:2019}. We report results for Thai.
    \item \textbf{MLQA} (\texttt{facebook/mlqa}) – 5.000\,+ extractive QA instances across seven languages \cite{lewis2019mlqa}. We report results for Arabic.
    \item \textbf{TyDi\,QA} (\texttt{google-research-datasets/tydiqa}) – 204\,k questions covering 11 languages \cite{tydiqa}. We include Arabic.
    \item \textbf{ARCD} – The Arabic Reading Comprehension Dataset. 1.395 crowd-sourced Arabic questions on Wikipedia articles \cite{mozannar-etal-2019-neural}.
    \item \textbf{CMRC 2018} – Chinese machine reading comprehension task \cite{cui-etal-2019-span}. \(\sim\)20.000 Chinese span-extraction QA pairs from Wikipedia.
    \item \textbf{Chinese-SQuAD} – a machine-translated and manually corrected Chinese version of SQuAD v1.1/2.0.
    \item \textbf{ThaiQA-SQuAD} – 4.074 Thai questions released in SQuAD format.
\end{itemize}
The results show strong improvements using the JQL filters instead of FW2 across all languages. Interestingly, though, we can see heavily diverging impacts on document retention. While our JQL-Edu filters (at the 0.7 percentile threshold) retain 2\% more tokens for Arabic, we see a drop in retained tokens of 40\% for Thai.

\begin{figure}
    \centering
    \includegraphics[width=0.6\linewidth]{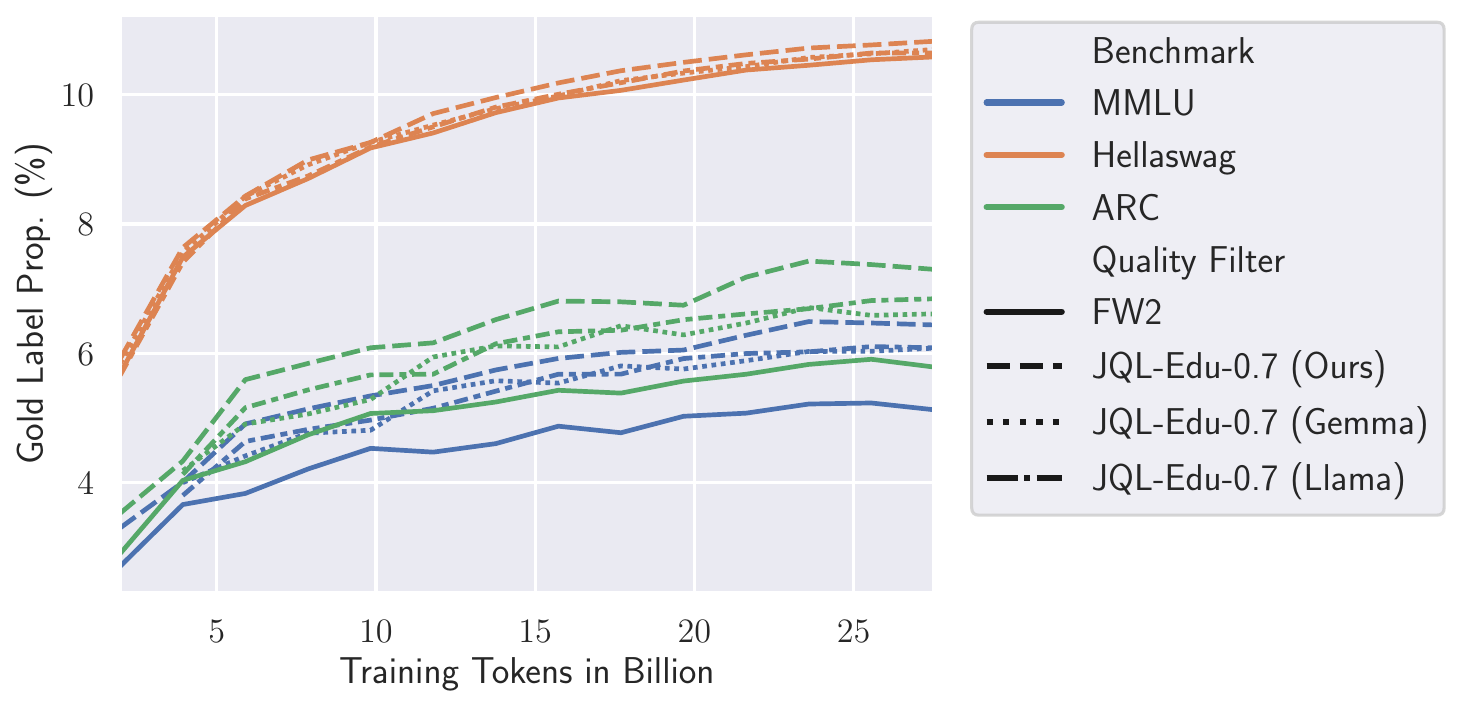}
    \caption{Direct comparison of Gemma and Llama as annotators.}
    \label{fig:gemma_vs_lamma}
\end{figure}
\clearpage
\section{Additional Ablations}

\subsection{Ablation on Long Context Documents}
\begin{figure}[ht]
    \centering
    \begin{subfigure}[t]{\textwidth}
        \includegraphics[width=\textwidth]{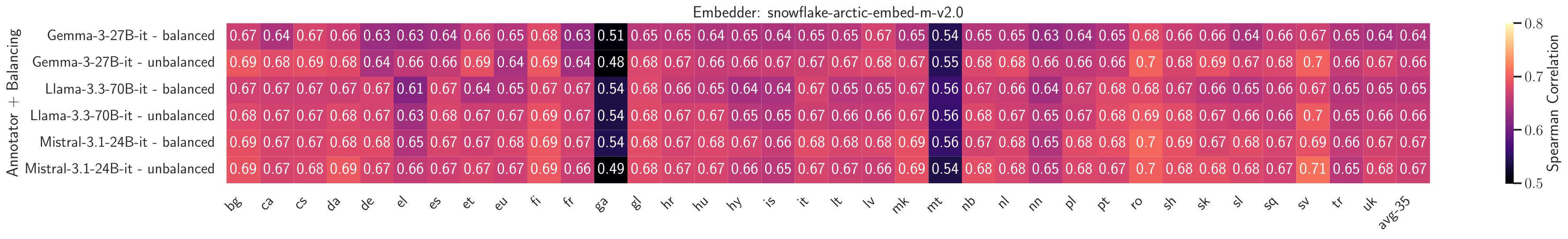}
        \caption{Spearman correlation on test set with \textbf{512} tokens context length.}
        \label{fig:heatmap_less-or-equal-512-tokens_snowflake}
    \end{subfigure}
    
    \begin{subfigure}[t]{\textwidth}
        \includegraphics[width=\textwidth]{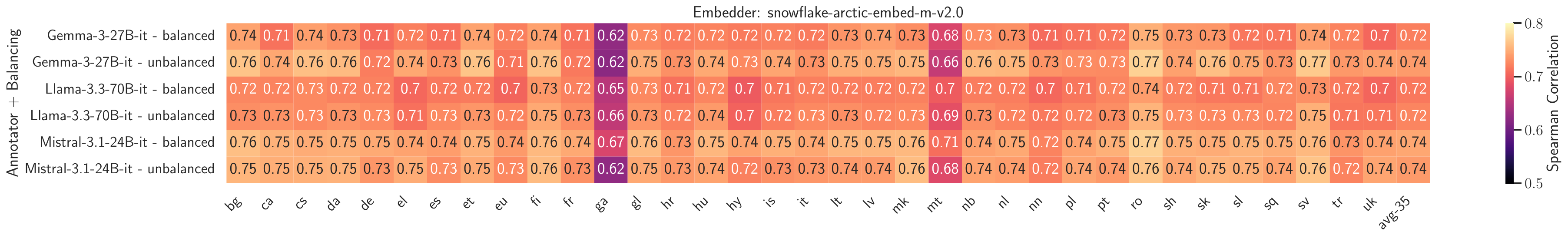}
        \caption{Spearman correlation improves when using full \textbf{8192} tokens context length.}
        \label{fig:heatmap_greater-512-tokens_snowflake}
     \end{subfigure}
    \begin{subfigure}[t]{\textwidth}
    \centering
    \includegraphics[width=0.8\linewidth]{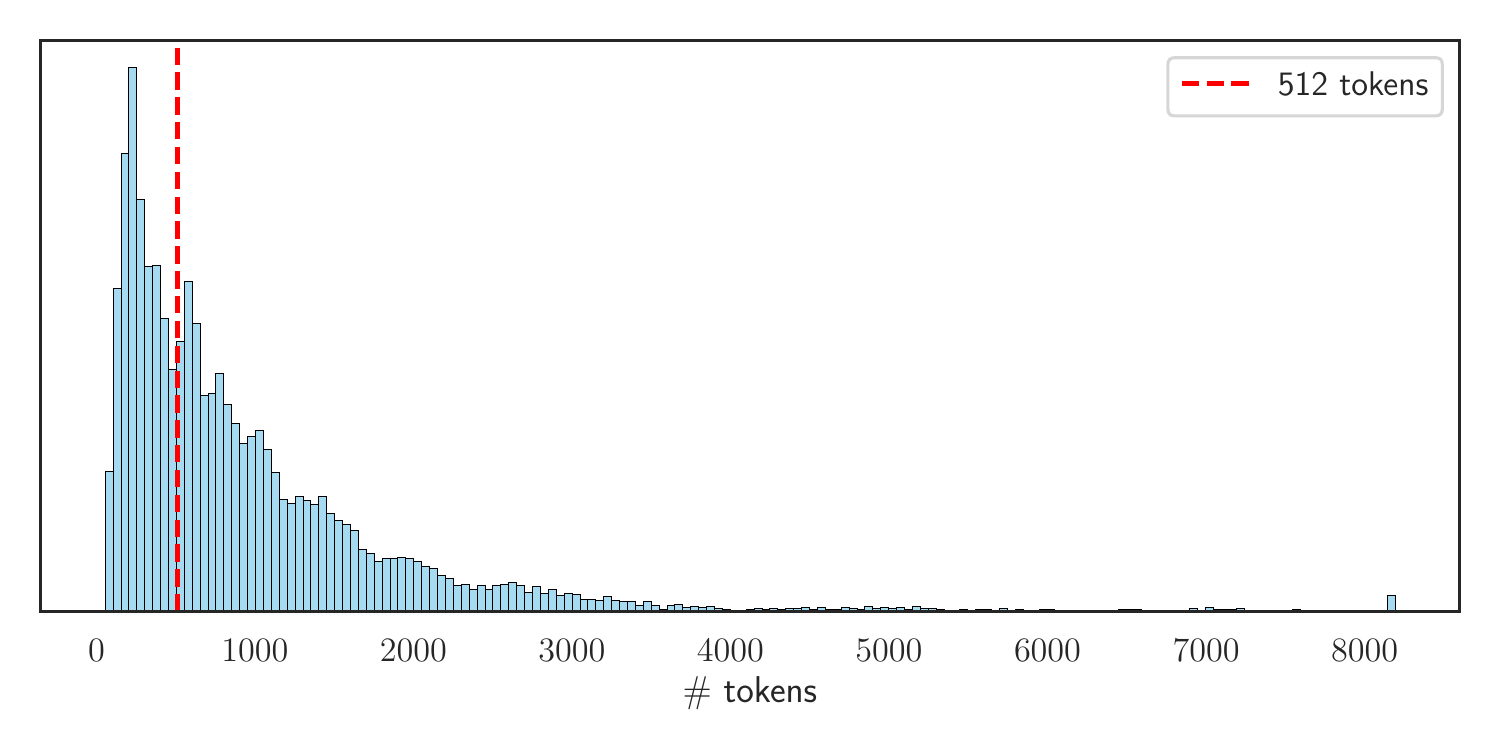}
    \caption{Token Counts across all Test Languages. We observe a meaningful percentage of documents longer then 512 tokens. }
    \label{fig:token-counts}
    \end{subfigure}
    \caption{Increased context length of lightweight JQL-annotators improved performance. }
    \label{fig:long_context_ablation}
\end{figure}

Contrary to previous works \cite{DBLP:conf/nips/PenedoKALMRW024}, JQL leverages embedding models with long context windows (i.e., 8k). \citet{DBLP:conf/nips/PenedoKALMRW024}, for example, only considered the initial 512 tokens of any document when assigning educational scores. 
Fig.~\ref{fig:long_context_ablation} highlights that a meaningful portion of documents is indeed longer then 512 tokens. Consequently, we observe a significant performance improvement of about 7 percentage points on average when using the lightweight annotator at 8192 tokens context length. For low-resource languages like Irish or Maltese, improvement increases up to 12 percentage points. 

\subsection{Influence of Ranking Performance and Ensembles on Data Quality}\label{app:classification_ablation}
In Sec.~\ref{sec:llms}, we observed that Mistral achieves higher classification accuracy against human ground truth compared to Gemma, while both models exhibit similarly strong ranking capabilities. To systematically evaluate the impact of this distinction, we conducted a controlled ablation study using the Spanish subset. Specifically, we compared data filtering outcomes using single annotator models—from Gemma and Llama labels—each applying their respective 0.7 percentile thresholds independently. Additionally, this setup simultaneously allows us to assess the value of ensemble-based annotations.

The results in Fig.~\ref{fig:gemma_vs_lamma} clearly indicate that the datasets filtered individually by Gemma and Llama yield very similar downstream training performance.
Consequently, we can conclude that strong ranking performance is substantially more relevant than classification accuracy for the task of selecting high-quality training data. Furthermore, we observed that both single-model-filtered datasets performed worse than the dataset selected through ensemble-based annotation, thereby underscoring the robustness provided by ensemble consensus filtering. These findings emphasize the limited practical importance of absolute classification accuracy when compared to our design pipeline, which focuses on ranking capabilities and uses an ensemble to enhance annotation robustness.

\section{Datasets}

Tab.~\ref{tab:datasets} presents the dataset statistics for our training and human-annotated test sets across all 35 languages included in our study.

For all languages except Norwegian (Nynorsk; 304.2k), Irish (390.3k), Latvian (438.3k), and Maltese (327.4k), we have at least 450k training annotations. In some cases, the test set contains fewer than 511 samples due to the removal of incorrectly translated documents.

\section{License of Used Artifacts}

Table~\ref{tab:artifacts_licenses} summarizes the licenses of the artifacts used in the context of our work. 
The majority of artifacts are shared under permissive license (e.g., CC, MIT, or Apache). The custom license agreements of the two LLMs we used\footnote{Note that Mistral is shared under Apache License} (Llama-3.3-70B-it and Gemma-2-27b-it) specifically allow for the use of generated outputs as conducted in our work. The only non-commercial licenses occurred for some of the benchmark datasets, which we solely used for academic evaluation.
Consequently, our usage aligns with the terms and intended scope of all respective licenses.

\begin{table}[h]
    \centering
    \begin{tabular}{ll}
        \toprule
        \textbf{Artifacts} & \textbf{License} \\
        \midrule
        \textbf{Pre-trained Models:} \\
        Gemma-2-27B-it & gemma \\
        Gemma-2-9B-it & gemma \\
        Gemma-3-27B-it & gemma \\
        Llama-3.1-8B-it & Llama 3.1 Community License Agreement \\
        Llama-3.2-3B-it & Llama 3.2 Community License Agreement \\
        Llama-3.3-70B-it & Llama 3.3 Community License Agreement\\
        Mistral-3.1-24B-it & Apache 2.0 License \\
        Phi-4-14B & MIT License \\
        Qwen-2.5-14B-it & Apache 2.0 License \\
        Qwen-2.5-32B-it & Apache 2.0 License \\
        Qwen-2.5-72B-it & Qwen License Agreement \\
        Qwen-2.5-7B-it & Apache 2.0 License \\
        Snowflake-arctic-embed-v2.0 & Apache-2.0 License \\ \hline
        \textbf{Libraries:} \\
        Nanotron & Apache-2.0 License\\
        Datatrove & Apache-2.0 License \\
        Lighteval & MIT License \\
        Transformers & Apache-2.0 License \\\hline
        \textbf{Pre-training Artifacts:} \\
        Fineweb-Edu & ODC-BY \\
        Fineweb-2 & ODC-BY \\ \hline
        
        \textbf{Benchmarks:} \\
        Open-AI-MMMLU & MIT License \\
        Cohere-GLobal-MMLU & Apache-2.0 License\\
        openGPT-X-arcx & Creative Commons Attribution Share Alike 4.0 \\
        openGPT-X-hellaswag-x & MIT License \\
        alexandrainst-m\_arc & Creative Commons Attribution Non Commercial 4.0 \\
        NbAiLab-nb-global-mmlu & Apache-2.0 License\\
        alexandrainst-m\_hellaswag & Creative Commons Attribution Non Commercial 4.0 \\
        malhajar-arc-tr & MIT License \\
        malhajar-hellaswag-tr & MIT License \\
        google-xQuAD & Creative Commons Attribution Share Alike 4.0 \\
        facebook-mlqa & Creative Commons Attribution Share Alike 3.0 \\
        google-tydiqa & Apache-2.0 License\\
        arcd &MIT License \\
        cmrc-2028 & Creative Commons Attribution Share Alike 4.0 \\
        chinese-squad & No license information available \\
        thaiQA-squad & Creative Commons Attribution Non Commercial Share Alike 3.0\\
        \bottomrule
    \end{tabular}
    \caption{Overview of used artifacts and their licenses.}
    \label{tab:artifacts_licenses}
\end{table}
\section{Data Containing Personally Identifiable Information or Offensive Content}

In this work, we introduce JQL, a method designed to enhance the quality of raw pre-training data by filtering out low-quality content. 
As part of this effort, we necessarily engage with data that may contain personally identifiable information (PII) or offensive material, as such content is commonly found in large-scale web corpora. 
While we do not explicitly quantify JQL’s effectiveness in isolating PII or offensive content, we assume that its JQL in general is capable in identifying such content.

\section{Infrastructure \& Compute Requirements}

In Table~\ref{tab:llm_gpu_hours}, we provide a summary of our compute requirements. 
To generate the LLM training annotations, we leveraged a large-scale compute cluster equipped with thousands of H100 GPUs, enabling efficient processing at scale.
ll tasks involving the lightweight annotators and downstream model training were performed on a cluster equipped with several hundreds of A100 GPUs.

\begin{table}[h]
\centering
\begin{tabular}{lccr}
\toprule
\textbf{Model} & \textbf{Task} & \textbf{GPU Type} & \textbf{GPU Hours}\\
\midrule
Gemma-3-27B-IT & Annotation Generation & H100 & 9072  \\
Mistral-3.1-24B-IT & Annotation Generation & H100 & 4464   \\
Llama-3.3-70B-IT & Annotation Generation & H100 & 10944   \\
\midrule
Lightweight Annotators & Embedding Training Data & A100 & 200 \\
Lightweight Annotators & Ablations & A100 & 300 \\
\midrule
Lightweight Annotators & Web Corpus Annotation & A100 & 23000 \\
\midrule
Custom LLMs (2B) & Downstream Training & A100 & 52000 \\
Custom LLMs (2B) & Evaluation & A100 & 720 \\
\bottomrule
\end{tabular}
\caption{Estimate of total compute requirements (in GPU hours) across different stages of the pipeline, including annotation generation and model training.}
\label{tab:llm_gpu_hours}
\end{table}


\section{Usage of AI Tools}

We made use of AI-assisted tools such as ChatGPT and GitHub Copilot to support writing and coding tasks. 
All AI-generated outputs were thoroughly validated to ensure their correctness.




\end{document}